\PassOptionsToPackage{unicode}{hyperref}
\PassOptionsToPackage{hyphens}{url}
\PassOptionsToPackage{dvipsnames,svgnames,x11names}{xcolor}
\documentclass[12pt]{article}

\usepackage{algorithm}
\usepackage{algpseudocode}

\usepackage{booktabs}
\usepackage{multirow}
\usepackage{caption}
\usepackage{graphicx}

\usepackage{amssymb}
\usepackage{amsmath}

\usepackage{ulem}
\usepackage{xcolor} % 

\usepackage[round]{natbib} % \citet \citep
\usepackage[colorlinks=true, citecolor=blue]{hyperref}
\usepackage{hyperref}
\hypersetup{
    colorlinks=true,
    linkcolor=blue,
    filecolor=black,
    urlcolor=blue,
    pdftitle={Overleaf Example},
    }
\usepackage{booktabs}% \toprule、\midrule、\bottomrule
\usepackage{multirow}% \multicolumn{}{}{}
\usepackage{graphicx}
\usepackage{epstopdf} % 
\usepackage{subcaption} % 
\usepackage{float}%H
\numberwithin{equation}{section}

\usepackage{amsmath,amssymb}
\usepackage{bbm}
\usepackage{iftex}
\ifPDFTeX
  \usepackage[T1]{fontenc}
  \usepackage[utf8]{inputenc}
  \usepackage{textcomp} % provide euro and other symbols
\else % if luatex or xetex
\fi
\usepackage{lmodern}
\ifPDFTeX\else  
\fi
\IfFileExists{upquote.sty}{\usepackage{upquote}}{}
\IfFileExists{microtype.sty}{% use microtype if available
  \usepackage[]{microtype}
  \UseMicrotypeSet[protrusion]{basicmath} % disable protrusion for tt fonts
}{}
\makeatletter
\@ifundefined{KOMAClassName}{% if non-KOMA class
  \IfFileExists{parskip.sty}{%
    \usepackage{parskip}
  }{% else
    \setlength{\parindent}{0pt}
    \setlength{\parskip}{6pt plus 2pt minus 1pt}}
}{% if KOMA class
  \KOMAoptions{parskip=half}}
\makeatother
\usepackage{xcolor}
\newcounter{condcounter}

\makeatletter
\ifx\paragraph\undefined\else
  \let\oldparagraph\paragraph
  \renewcommand{\paragraph}{
    \@ifstar
      \xxxParagraphStar
      \xxxParagraphNoStar
  }
  \newcommand{\xxxParagraphStar}[1]{\oldparagraph*{#1}\mbox{}}
  \newcommand{\xxxParagraphNoStar}[1]{\oldparagraph{#1}\mbox{}}
\fi
\ifx\subparagraph\undefined\else
  \let\oldsubparagraph\subparagraph
  \renewcommand{\subparagraph}{
    \@ifstar
      \xxxSubParagraphStar
      \xxxSubParagraphNoStar
  }
  \newcommand{\xxxSubParagraphStar}[1]{\oldsubparagraph*{#1}\mbox{}}
  \newcommand{\xxxSubParagraphNoStar}[1]{\oldsubparagraph{#1}\mbox{}}
\fi
\makeatother

\usepackage{longtable,booktabs,array}
\usepackage{calc} % for calculating minipage widths
\usepackage{etoolbox}
\makeatletter
\patchcmd\longtable{\par}{\if@noskipsec\mbox{}\fi\par}{}{}
\makeatother
\IfFileExists{footnotehyper.sty}{\usepackage{footnotehyper}}{\usepackage{footnote}}
\makesavenoteenv{longtable}
\usepackage{graphicx}
\makeatletter
\def\maxwidth{\ifdim\Gin@nat@width>\linewidth\linewidth\else\Gin@nat@width\fi}
\def\maxheight{\ifdim\Gin@nat@height>\textheight\textheight\else\Gin@nat@height\fi}
\makeatother
\setkeys{Gin}{width=\maxwidth,height=\maxheight,keepaspectratio}
\makeatletter
\def\fps@figure{htbp}
\makeatother

\makeatletter
\@ifpackageloaded{caption}{}{\usepackage{caption}}
\AtBeginDocument{%
\ifdefined\contentsname
  \renewcommand*\contentsname{Table of contents}
\else
  \newcommand\contentsname{Table of contents}
\fi
\ifdefined\listfigurename
  \renewcommand*\listfigurename{List of Figures}
\else
  \newcommand\listfigurename{List of Figures}
\fi
\ifdefined\listtablename
  \renewcommand*\listtablename{List of Tables}
\else
  \newcommand\listtablename{List of Tables}
\fi
\ifdefined\figurename
  \renewcommand*\figurename{Figure}
\else
  \newcommand\figurename{Figure}
\fi
\ifdefined\tablename
  \renewcommand*\tablename{Table}
\else
  \newcommand\tablename{Table}
\fi
}
\@ifpackageloaded{float}{}{\usepackage{float}}
\floatstyle{ruled}
\@ifundefined{c@chapter}{\newfloat{codelisting}{h}{lop}}{\newfloat{codelisting}{h}{lop}[chapter]}
\floatname{codelisting}{Listing}

\makeatother
\makeatletter
\@ifpackageloaded{caption}{}{\usepackage{caption}}
\@ifpackageloaded{subcaption}{}{\usepackage{subcaption}}
\makeatother

\ifLuaTeX
  \usepackage{selnolig}  % disable illegal ligatures
\fi
\usepackage{bookmark}

\IfFileExists{xurl.sty}{\usepackage{xurl}}{} % add URL line breaks if available
\hypersetup{
  pdftitle={Title},
  pdfauthor={Author 1; Author 2},
  pdfkeywords={3 to 6 keywords, that do not appear in the title},
  colorlinks=true,
  linkcolor={blue},
  filecolor={Maroon},
  citecolor={Blue},
  urlcolor={Blue},
  pdfcreator={LaTeX via pandoc}}

\newcommand{\bzero}{{\mathbf{0}}}
\newcommand{\htheta}{{\hat\theta}}
\newcommand{\hepsilon}{{\hat{\epsilon}}}

\newcommand{\blambda}{{\mbox{\boldmath$\lambda$}}}
\newcommand{\bs}{{\mbox{\boldmath$s$}}}
\newcommand{\bt}{{\mbox{\boldmath$t$}}}
\newcommand{\bg}{{\mbox{\boldmath$g$}}}

\newcommand{\bmu}{{\mbox{\boldmath$\mu$}}}

\newcommand{\bX}{{\mbox{\boldmath$X$}}}
\newcommand{\bepsilon}{{\mbox{\boldmath$\epsilon$}}}
\newtheorem{defi}{\sc Definition}
\newtheorem{theo}{\sc Theorem}
\newtheorem{lemm}{\sc Lemma}

\newtheorem{coll}{\sc Corollary}
\newtheorem{rema}{\sc Remark}

\newcommand{\tod}{\stackrel{d}{\longrightarrow}}

\newcommand{\dto}{\stackrel{d}{\rightarrow}}
\newcommand{\pto}{\stackrel{p}{\rightarrow}}
\newcommand{\toiid}{\stackrel{iid}{\sim}}

\renewcommand{\baselinestretch}{1.25}

\def\be{\begin{equation}}
\def\ee{\end{equation}}
\def\nn{\nonumber}
\def\bea{\begin{eqnarray}}
\def\eea{\end{eqnarray}}
\newcommand{\anon}{1}

\begin{document}

\def\spacingset#1{\renewcommand{\baselinestretch}%
{#1}\small\normalsize} \spacingset{1}

%%%%%%%%%%%%%%%%%%%%%%%%%%%%%%%%%%%%%%%%%%%%%%%%%%%%%%%%%%%%%%%%%%%%%%%%%%%%%%

\if1\anon
{
  \title{\bf  Empirical Likelihood-Based Fairness Auditing: Distribution-Free Certification and Flagging}
  \author{
  Jie Tang$^1$,  
  Chuanlong Xie$^{1}$\thanks{Author list is in alphabetical order by last names. Correspondence to Chuanlong Xie. E-mail: clxie@bnu.edu.cn; ORCID: https://orcid.org/0000-0003-4292-8782.} \hspace{.1cm},
Xianli Zeng$^2$, 
and Lixing Zhu$^1$ \\
    (\small $^1$Department of Statistics, Beijing Normal University, Zhuhai, Guangdong 519087, China\\
   \small $^2$School of Economics, Xiamen University, Xiamen, Fujian 361005, China)}
  \maketitle
} \fi

\if0\anon
{
  \bigskip
  \bigskip
  \bigskip
  \begin{center}
    {\LARGE\bf Empirical Likelihood-Based Fairness Auditing: Distribution-Free Certification and Flagging}
\end{center}
  \medskip
} \fi

\bigskip
\begin{abstract}
%\tecr{The text of your unstructured abstract. 200 or fewer words.} 
Machine learning models in high-stakes applications, such as recidivism prediction and personnel selection, often exhibit performance disparities across sensitive subpopulations, raising critical concerns regarding algorithmic bias. Fairness auditing addresses these risks through two primary functions: certification, which verifies adherence to fairness constraints; and flagging, which isolates specific demographic groups experiencing disparate treatment. However, existing auditing techniques are frequently limited by restrictive distributional assumptions or prohibitive computational overhead.
We propose a novel empirical likelihood-based framework that constructs robust statistical measures for model performance disparities. Unlike traditional methods, our approach is non-parametric;  the proposed disparity statistics follow asymptotically chi-square or mixed chi-square distributions, ensuring valid inference without assuming underlying data distributions. This framework uses a constrained optimization profile that admits stable numerical solutions, facilitating both large-scale certification and efficient subpopulation discovery.
Empirically, our method outperforms bootstrap-based approaches, yielding coverage rates closer to nominal levels while reducing computational latency by several orders of magnitude. We demonstrate the practical utility of this framework on the COMPAS dataset, where it successfully flags intersectional biases—specifically identifying a significantly higher positive prediction rate for African-American males under 25 and a systemic under-prediction for Caucasian females relative to the population mean.
\end{abstract}

\noindent%
{\it Keywords:} 
%\tecr{3 to 6 keywords, that do not appear in the title.} 
Fairness auditing;  Empirical likelihood; Confidence intervals
\vfill

\newpage
\spacingset{1.8} % DON'T change the spacing!

\section{Introduction}\label{sec-intro}

% Background
Highly accurate Artificial Intelligence (AI) models are now deeply integrated into the fabric of society, profoundly influencing critical life outcomes in sectors such as hiring, credit approval, risk assessment, medical resource allocation, and criminal justice. 
However, high predictive accuracy does not inherently guarantee fairness. When trained on biased historical data, even state-of-the-art models often perpetuate and amplify embedded systemic biases, thereby exacerbating inequities in decision-making.
A prominent example is found in the United States criminal justice system, where recidivism prediction algorithms have demonstrated significantly higher false positive rates for African American individuals compared to Caucasian parolees \citep{AngwinEtAl2016}. 
Similar disparities have been documented in targeted advertising \citep{Dastin2018} and automated hiring systems \citep{Simonite2015}. % remove DattaEtAl2015
%These documented cases of bias motivate the critical task of fairness certification, which aims to verify whether a model adheres to specified fairness requirements across subpopulations. Furthermore, models that appear equitable in aggregate may still exhibit biases within smaller, intersectional subgroups. For instance, facial recognition systems have shown significantly higher misclassification rates for dark-skinned women than for either dark-skinned individuals or women as seperate group \citep{BuolamwiniGebru2018}. This observation drives the need for unfairness identification, a task focused on localizing and flagging performance disparities across fine-grained subpopulations. Both certification and identification are essential for ensuring the ethical deployment of AI and protecting vulnerable populations. 

In response to these concerns, diverse stakeholders have formulated ethical principles to guide AI development, emphasizing third-party auditing as a vital mechanism for verifying fairness claims and fostering trustworthiness \citep{BrundageEtAl2020, LandersBehrend2022, LaineEtAl2024, LacmanovićŠkare2025}. 
Fairness audits are designed to determine whether a machine learning algorithm exhibits discrimination and to identify the specific subpopulations adversely affected 
\citep{SchaakeClark2022, CherianCandes2024}. %remove PaganoEtAl2023, 
Specifically, a certification audit identifies groups for which disparities are statistically insignificant \citep{TramerEtAl2017, MorinaEtAl2019, XueEtAl2020, TaskesenEtAl2021,  RoyMohapatra2023}, while a flagging audit uncovers groups where disparities exceed a tolerable threshold \citep{YanZhang2022, vonZahnEtAl2023, CherianCandes2024}.  

%Current auditing methodologies generally focus on either group fairness or individual fairness \citep{MitchellEtAl2021, PessachShmueli2022, CatonHaas2024, HeLi2025}. 
From a statistical perspective, the audits rely on the statistical test that rejects the null hypothesis if any disparity is detected across a predefined set of sensitive groups \citep{TramerEtAl2017, DiCiccioEtAl2020, SiEtAl2021}. 
Research in fairness auditing has followed two primary trajectories: the development of interval estimates with statistical guarantees for quantifying disparities \citep{XueEtAl2020, MaityEtAl2021, RoyMohapatra2023, CherianCandes2024}, and the use of multiple hypothesis testing to localize unfair subpopulations among a large collection \citep{YanZhang2022, SchaakeClark2022, vonZahnEtAl2023}. Despite their utility, existing approaches face significant practical limitations: Permutation-based methods require the strong assumption of identical data distributions across groups and suffer from high computational costs due to repeated variance re-estimation \citep{DiCiccioEtAl2020}. Bootstrap-based methods are highly sensitive to resampling schemes and become computationally prohibitive when applied to large-scale datasets \citep{XueEtAl2020, RoyMohapatra2023, CherianCandes2024}. Optimal transport distance-based methods offer geometric robustness but are currently restricted to specific metrics and require knowledge of model internals. Moreover, they are struggle with the optimization complexity of large-scale subpopulation discovery \citep{MorinaEtAl2019, XueEtAl2020, TaskesenEtAl2021, SiEtAl2021}. 

To address these challenges, we propose an Empirical Likelihood-based Fairness Auditing (ELFA) framework. The empirical likelihood (EL) method, introduced by \cite{Owen1988}, offers several distinct advantages for fairness auditing. First, it is non-parametric and requires no specific distributional assumptions, making it more robust to real-world data. Second, through internal %studentization, 
standardization, it avoids explicit variance estimation and the need for pivot quantities. Third, EL confidence regions are data-adaptive, with shapes and orientations purely determined by the sample itself. Unlike the bootstrap, which uses equal weights, EL profiles a multinomial likelihood supported on the sample \citep{Chen1996}. Crucially, EL is Bartlett-correctable, reducing coverage error to $O(n^{-2})$, an improvement over the $O(n^{-1})$ error rates typical of bootstrap methods \citep{DiCiccioEtAl1991}. While the EL method has been successfully applied to linear models \citep{Owen1991}, general estimating equations \citep{QinLawless1994}, and high-dimensional settings \citep{LengTang2012}, its application to fairness remains sparse. A recent exception includes using EL to build fair prediction models via covariance proxies \citep{LiuZhao2024}, but its potential for fairness auditing has not been fully explored. 

In this paper, we leverage the disparity definitions proposed by \citep{CherianCandes2024} to create a framework that covers general group disparity scenarios. Since our disparities are based on model outputs, the ELFA framework remains model-agnostic. We construct empirical likelihood ratio statistics whose limiting distributions are chi-squared, enabling the construction of rigorous confidence intervals. We also extend the framework to accommodate inequality-constrained null hypotheses and provide theoretical guarantees for Type I and II errors. Our framework performs two core functions:
(1) certification and (2) flagging. The former verifies model fairness by checking whether zero falls within the confidence interval. The latter  discover discriminated groups through the Benjamini-Hochberg (BH) procedure. 
Simulation studies demonstrate that ELFA more closely approximates nominal confidence levels and significantly reduces computational time compared to bootstrap alternatives. 
It is also more robust to skewed data than the standard $T$-test.
Finally, an empirical analysis of the COMPAS dataset reveals that the algorithm exhibits higher positive prediction rates for African-American males (specifically those under 25) compared to Caucasians, while Caucasian females receive a rate lower than the average.

The primary contributions of this paper are summarized as follows:

\begin{itemize} \item \textbf{Methodological innovation:} We introduce the ELFA framework, a novel approach based on empirical likelihood for fairness auditing. This framework operates without requiring model-specific or distributional assumptions, ensuring robustness and broad applicability to complex, real-world scenarios. Notably, ELFA provides a unified treatment for diverse group fairness disparity metrics and offers flexible auditing procedures tailored to specific application contexts.

\item \textbf{Theoretical guarantees:} We establish the large-sample theory for the empirical likelihood ratio statistics, proving that their limiting distributions follow chi-squared distributions. For the certification task, we leverage these limiting distributions to construct confidence regions that provide rigorous Type I error guarantees. 
Furthermore, we derive theoretical guarantee for Type I and II errors across various hypothesis testing scenarios, ensuring the statistical validity of both certifying and flagging tasks. 
%To support the flagging task, we integrate our testing procedure with the BH procedure to effectively control false flagging rates during multiple hypothesis testing.

\item \textbf{Computational advantages:} Our methodology eliminates the need for computationally intensive bootstrap resampling and avoids the requirement for explicit pivotal quantities. Due to its data-adaptive nature, the ELFA framework demonstrates high efficiency and robustness. Simulation studies confirm that our method significantly reduces computational latency compared to bootstrap-based approaches.
%achieves coverage probabilities closer to nominal levels while significantly reducing computational latency compared to traditional bootstrap-based approaches.

\item \textbf{Dual-task framework:} We develop a cohesive statistical framework that simultaneously addresses two complementary objectives: the statistically-guaranteed certification of fair subpopulations and the systematic flagging of unfair subpopulations. This dual-task capability provides a comprehensive toolkit for auditors to verify compliance and identify risks within a single, rigorous pipeline. \end{itemize}

% \item \textbf{Empirical validation:} We validate the proposed method through comprehensive simulation studies and empirical analysis on the COMPAS dataset, demonstrating its practical effectiveness in identifying performance disparities across sensitive subpopulations.

The article is organized as follows. 
Section~\ref{sec-setup} introduces the setup and notations. 
Sections~\ref{certifying} and~\ref{sec-flagging} present the main results for certifying and flagging performance disparities, respectively. 
Section~\ref{sec-simu} reports 
%results from 
simulation studies, and Section~\ref{sec-real-data} presents %the analysis of 
a real data example. Section~\ref{sec-concluding} includes concluding. %and discussion. 
%All technical details and proofs are presented in the Supplemental Material~ A, B and C.
Technical details are postponed into the Supplemental Material.

\section{The setup}
\label{sec-setup}

We assume that only a holdout dataset $D = \{X_i, Y_i\}^n_{i=1} \toiid P$ is accessible. This is sometimes referred to audit trails \citep{BrundageEtAl2020}. We say that some prediction rule $h$ exhibits a performance disparity on a subpopulation $G$ if the conditional expectation of some metric $M(h (X), Y)$ 
given $(X, Y)\in G$
differs substantially from a target $\theta_P \in \mathbb{R}$. 
%Our statistical audits are conducted by testing and constructing confidence intervals for the group-wise performance disparity, a metric formalised by \citep{CherianCandes2024} as follows. 
The group-wise performance disparity is given by \cite{CherianCandes2024} as follows. 

\begin{defi}\label{defi1} A group-wise performance disparity is defined as \be \epsilon_G=\mathbb{E}_P[M(h (X), Y ) | (X, Y) \in G] - \theta_P,\ee where $G \in \mathcal{G}$,  $\mathcal{G} = \{ G_1, G_2, \dots, G_m\}$ and $m  < \infty$.
%$m =|\mathcal{G}| < \infty$ where $|\mathcal{G}|$ is defined as the cardinality of $\mathcal{G}$.
\end{defi} 

While subgroup membership is often determined by a subset of covariates (such as sensitive attributes) that may not be explicitly used by the prediction rule, we use the same covariate vector $X$ for both the model input and the group definitions to maintain notational simplicity. Definition \ref{defi1} serves as a flexible, unified framework capable of subsuming nearly any standard group fairness definition through the specific choice of $\epsilon_G$. To illustrate this versatility, consider Statistical Parity (also referred to as Demographic Parity) \citep{DworkEtAl2012, MehrabiEtAl2021, GargEtAl2020, VermaRubin2018a}. 
This criterion demands that the probability of a positive prediction remains invariant across different groups: $\mathbb{P}(h(X)=1\mid X \in G)=\mathbb{P}(h(X)=1).$ 
The disparity $\epsilon_G$ is
$$\epsilon_G = \mathbb{P}(h(X)=1\mid X \in G)- \mathbb{P}(h(X)=1).$$ 
By mapping this to Definition \ref{defi1}, we simply assign: 
\bea M(h (X), Y ) = \mathbbm{1} \{h (X) = 1\} \quad \mathrm{and} \quad  \theta_P = \mathbb{P}(h(X)=1). \nn\eea
For the certification task, we evaluate whether the model $f$ satisfies the fairness criterion by testing the null hypothesis $H_0: \epsilon_G = 0, \forall G\in \mathcal{G}$. %For the auditing task, 
We construct confidence sets using 
a significance level $\alpha$.
Our fairness certification is then established by determining whether zero is contained within these confidence intervals. 
As for the flagging task, where $\mathcal{G} = \{ G_1, G_2, \dots, G_m\}$ comprises multiple subgroups, we identify unfair groups by testing the null hypothesis %$H_0(G_i): \epsilon_{G_i} \leq \epsilon$ 
$H_0: \epsilon_{G_i} \leq \epsilon_0$
for each group $G_i$, where $\epsilon_0$ represents a pre-specified tolerance threshold. %Utilizing the methodology detailed in Section 4, we compute $p$-values for each individual null hypothesis. The Benjamini-Hochberg (BH) procedure is then applied to the collection of $p$-values to control for multiple testing. 
Ultimately, a group $G_i$ is flagged as unfair if its corresponding null hypothesis $H_0(G_i)$ is rejected.
%Equal Opportunity \citep{HardtEtAl2016}: Compares model's false negative rates, i.e., prediction probabilities for the negative class ($\hat{Y} = 0$) for the known positive class ($Y = 1$): 
%$$P(\hat{Y} = 0|Y = 1, G = 0) = P(\hat{Y} = 0|Y = 1, G = 1)$$
%Equalised Odds \citep{HardtEtAl2016}: Compares model's prediction probabilities for the positive class ($\hat{Y} = 1$) for different ground truth classes ($Y = 1$ and $Y = 0$): $P(\hat{Y} = 1|Y = y, G = 0) = P(\hat{Y} = 1|Y = y, G = 1)$, where $y \in \{0, 1\}$.

% To evaluate these disparities, our proposed audit  requires only access to a holdout dataset $D = \{X_i, Y_i\}^n_{i=1} \toiid  P$; this is sometimes referred to audit trails \citep{BrundageEtAl2020}.

{\bf Notation.}
Let $\mathbb{P}(G) := \mathbb{P}((X, Y ) \in G)$ denote the  probability of $(X, Y )$ belonging to subpopulation $G$.
To simplify our notation,
we replace$ (X, Y ) \in G$ with $G $ whenever the meaning is clear. 
We also replace $M(h (X), Y )$ by the abbreviation $M$.
For a function $f : X \times Y \to \mathbb{R}^k$, $P [f]$ is shorthand for $\mathbb{E}_P [f (X, Y )]$, $\mathbb{P}_n[f ]$ is shorthand  for $n^{-1} \sum_{i=1}^n f (X_i, Y_i)$. 
We also write $(\mathbb{P}_n -  P )[f ]$ in place of $\mathbb{P}_n[f ] - P [f ]$. Given a class of functions $\mathcal{F}$ , we regard $f \mapsto \sqrt{n}(\mathbb{P}_n - P )[f ]$ as a mapping belonging to $\ell_{\infty}(\mathcal{F})$. We also denote  $\mathbb{P}_n(G) = \frac{1}{n}  \sum_{i = 1}^n \mathbbm{1} \{(x_i, y_i) \in G\}$ and $|G| = n  \cdot \mathbb{P}_n(G)$. 
%$\btheta = \{ \theta_P \}_{i=1}^m$, and $\hat{\btheta} = \left\{ \htheta \right\}_{i=1}^m$

\section{ Model fairness certification} \label{certifying}

This section addresses the primary objective of this paper: certifying model fairness across multiple subpopulations. 
%Let $\hat{\bepsilon} = \{ \hat{\epsilon}_{G_i}\}_{i=1}^m$ represent the vector of observed disparities and $\bepsilon^* = \{ \epsilon_{G_i}\}_{i=1}^m$ denote the vector of true disparities. 
Let $ \bepsilon = \{ \epsilon_{G_i}\}_{i=1}^m$ represent the vector of disparities.
We formulate the certification task as a hypothesis testing problem: %where the null hypothesis is defined as:
$$H_0: \bepsilon = \mathbf{0} \quad \text{versus} \quad H_1: \bepsilon \neq \mathbf{0}.$$
A rejection of $H_0$ indicates statistically significant performance disparities, thereby flagging potential fairness concerns.

To conduct this test, we employ Empirical Likelihood (EL) to construct confidence intervals for the true disparity of each group. For a pre-specified significance level $\alpha$, the resulting $1-\alpha$ confidence interval is guaranteed to cover the true disparity with a probability of at least $1-\alpha$.
Certification is determined by the position of $\mathbf{0}$ relative to this interval:\\
{\it Reject $H_0$:} If the interval does not contain $\mathbf{0}$ (equivalently, if $p < \alpha$), the model fails certification for some subpopulation.\\
{\it Fail to Reject $H_0$:} If the interval contains $\mathbf{0}$ (equivalently, if $p \geq \alpha$), we issue a certificate of fairness, indicating the model satisfies the fairness requirements.

%{\color{red}Another way for this problem is to use $p$-values that are derived from the limiting distribution of the empirical likelihood ratio. This way is also useful in the cases where  the number of subpopulations $m$ is large. Note that in these cases,  computing the standard EL ratio for every group becomes computationally prohibitive. To address this bottleneck and enhance scalability, we propose the Euclidean Empirical Likelihood (EEL) method. This approach streamlines the process, enabling efficient, simultaneous testing across all groups without the heavy computational overhead of traditional EL. Then, we can use multiple hypothesis testing method to handle this task.}

\subsection{EL certification}

Unlike the empirical likelihood method for mean models, our estimation equation involves mean functions multiplied by indicator functions, which poses additional challenges in asymptotic analysis.
To address this, we employ empirical process theory to prove that, under mild conditions, the function class satisfies the P-Glivenko-Cantelli property and ensures uniform convergence over the entire function class.
Based on this key theoretical result, we are able to derive the asymptotic distribution of the empirical likelihood ratio statistic. See Proof of Theorem 1 in the Supplemental Material for more details.

First, consider the case where $\theta_P$ is known a priori.
By definition (\ref{defi1}) and $\mathbb{E}_P  [ M \mid G ] = \frac{1}{\mathbb{P}(G)} \mathbb{E}_P [ M \cdot \mathbbm{1}_G ]$, we obtain the moment equations
\bea  \label{moment_constraint}
\mathbb{E}_P \left\{ \frac{1}{\mathbb{P}(G)} [ M - \theta_P - \epsilon_G ] \cdot \mathbbm{1}_G  \right\} = 0. 
\eea
%Then the estimating equations for the method of moments are as follows
%\bea  
%g(G, x, y; \theta_P, \epsilon_G) 
%=    \left[ M - \theta_P - \epsilon_G \right]\cdot \mathbbm{1}_G . \nn
%\eea
%We denote
%$g(\epsilon_G) := g(G, x, y; \theta_P, \epsilon_G) $.
The we denote $p_i=P(x_i, y_i)$ and $\bg_i(\bepsilon;\theta_P) = \bg(x_i, y_i; \bepsilon,\theta_P)$, where
\bea \label{eq:3.2}
\bg(x, y; \bepsilon, \theta_P)  := 
\left(
	\begin{array}{c}
		  \left[ M - \theta_P - \epsilon_{G_1} \right]\cdot \mathbbm{1}_{G_1}   \\
		  \vdots \\
		  \left[ M - \theta_P - \epsilon_{G_m} \right]\cdot \mathbbm{1}_{G_m}  \\
	\end{array} 
\right). 
\eea
% This can be easily extended to multiple groups by making 
%  $\bg(x, y;\bepsilon) = \{g(\epsilon_{G_i})\}_{i=1}^m $ and then
% \bea 
% \bg(x, y; \bepsilon)  = 
% \left(
% 	\begin{array}{c}
% 		  \left[ M - \theta_P - \epsilon_{G_1} \right]\cdot \mathbbm{1}_{G_1}   \\
% 		  \vdots \\
% 		  \left[ M - \theta_P - \epsilon_{G_m} \right]\cdot \mathbbm{1}_{G_m}  \\
% 	\end{array} 
% \right)_{m \times 1} . \nn
% \eea
%
% Using the above function, we have 
% \bea \label{gEL}
% \bg(x_i, y_i; \bepsilon) =  
% \left(
% 	\begin{array}{c}
% 		   \left[ M_i - \theta_P - \epsilon_{G_1} \right]\cdot \mathbbm{1}_{G_1}   \\
% 		  \vdots \\
% 		   \left[ M_i - \theta_P - \epsilon_{G_m} \right]\cdot \mathbbm{1}_{G_m}  \\
% 	\end{array} 
% \right)_{m \times 1} . 
% \eea
Thus, the fairness constraint (\ref{moment_constraint}) is defined to be
$ 
\sum_{i=1}^{n} p_i \bg_i(\bepsilon;\theta_P) = \bzero 
$,
% Thus, let $\bg_i(\bepsilon) := \bg(x_i, y_i; \bepsilon)$, and the proposed constraint is defined to be
% \bea \label{RELM}
% \sum_{i=1}^{n} p_i \bg_i(\bepsilon) = \bzero,   \nn
% \eea
and then the EL function at $\bepsilon$ is defined to be
\bea 
L_{EL}(\bepsilon;\theta_P) =\sup_{p_i,1\le i\le n}  \left\{ \prod^n_{i=1} p_i \ \Big| \ p_i\geq 0, \,\, \sum_{i=1}^np_i = 1, \,\, \sum_{i=1}^n p_i  \bg_i(\bepsilon;\theta_P) = \bzero \right\}. \nn
\eea
The resulting empirical log-likelihood ratio function is defined to be
% \bea \label{REL}
% R_{EL}(\bepsilon;\theta_P)=\sup_{p_i,1\le i\le n}  \left\{ \prod^n_{i=1}n p_i \ \Big| \ p_i\geq 0, \quad \sum_{i=1}^np_i = 1, \quad \sum_{i=1}^n p_i  \bg_i(\bepsilon;\theta_P) =\bzero \right\}.  \nn
% \eea
\bea \label{log-EL}
\tilde{\ell}_{EL}(\bepsilon;\theta_P)=\sup_{p_i,1\le i\le n}  \left\{ \log(\prod^n_{i=1}n p_i) \ \Big| \ p_i\geq 0, \,\, \sum_{i=1}^np_i = 1, \,\, \sum_{i=1}^n p_i  \bg_i(\bepsilon;\theta_P) =\bzero \right\}.  \nn
\eea
Following \cite{Owen1990}, for a given $\bepsilon$, a unique maximum exists. Provided that $\bzero$ is inside the  convex hull of the points $\{\bg_1(\bepsilon;\theta_P) , \cdots , \bg_n(\bepsilon;\theta_P)\}$, one can show that
\bea 
\widehat{p}_i = \widehat{p}_i(\bepsilon) = \frac{1}{n\left(1 + \blambda^{\tau} \bg_i(\bepsilon;\theta_P)\right)}, \quad 1 \leq i \leq n,  \nn
\eea
and 
$\blambda \in \mathbb{R}^m$ is the solution of the following equation:
\bea \label{lambda}
{ 1 \over n} \sum_{i=1}^n{ \bg_i(\bepsilon;\theta_P) \over 1+\blambda^{\tau}  \bg_i(\bepsilon;\theta_P) }=\bzero.  
\eea
%, which establish the asymptotic normality of  $\sqrt{n}(\theta_P-\theta)$
%and $\sqrt{n}\{(\theta_P_{\gamma_2}-\theta_P_{\gamma_1})-(\theta_{\gamma_2}-\theta_{\gamma_1})\}$.
The detailed derivation is given in the supplemental material, Section~\ref{appsec-lagrange-multiplier}.
Therefore, the empirical log-likelihood ratio statistic at $\bepsilon$ is $\ell_{EL}(\bepsilon;\theta_P) = -2 \tilde{\ell}_{EL}(\bepsilon;\theta_P)$. To obtain the limiting distribution of $\ell_{EL}(\bepsilon;\theta_P)$,  we need the following assumptions.

\begin{itemize}
\setcounter{condcounter}{0}

\refstepcounter{condcounter}\item[C1.]\label{C1} ${ (X_i, Y_i), 1 \le i \le n}$ are independent and identically distributed random  variables.

\refstepcounter{condcounter}\item[C2.]\label{C2}  $\mathbb{P}(G) $ and $Var( M \mid G )$ is bounded away from 0 for all $G \in \mathcal{G}$. 

\refstepcounter{condcounter}\item[C3.]\label{C3} $\mathbb{E}_P[M^2] < \infty$, $0 < Var(M) < \infty$, $\|M\|_{\infty}$ are finite, where $\|\cdot \|_{\infty}$ denotes the supremum norm. 

\refstepcounter{condcounter}\item[C4.]\label{C4} ${\epsilon_G}$ are  uniformly bounded in absolute value for all $G \in \mathcal{G}$.

%\item[C5.] $m = |\mathcal{G}|$ is finite.

\refstepcounter{condcounter}\item[C5.]\label{C5} There is a constants $c_j > 0, j = 1, 2,$ such that $0 < c_1 \le  \lambda_{min}  (\Sigma_{EL}  ) \le  \lambda_{max}  (\Sigma_{EL}  ) \le c_2 < \infty$, where $\lambda_{min}(A)$ and $\lambda_{max}(A)$  denote the minimum and maximum eigenvalues of a matrix $A$, respectively 
\bea \label{SigmaEL}
\Sigma_{EL} = \Sigma_{EL}^{\tau} = Cov \left(  {1 \over \sqrt{n}}{ \sum_{i=1}^n  \bg_i(\bepsilon;\theta_P) } \right) =\left( \sigma^2_{kj}\right)_{m \times m}, 
\eea
where 
\bea
\sigma^2_{kj} = { \mathbb{E}\left[(M - \theta_P - \epsilon_{G_k})(M - \theta_P - \epsilon_{G_j}) \mathbbm{1}_{G_k \cap  G_j }\right]  }, \nn
\eea
and
\bea
\sigma^2_{kk} = { \mathbb{E} \left[(M - \theta_P - \epsilon_{G_k})^2 \mathbbm{1}_{G_k} \right] } = { \mathbb{P}(G_k) }Var(M \mid G_k). \nn
\eea

\refstepcounter{condcounter}\item[C6.]\label{C6} The estimator of the target %in Definition \ref{defi1} 
is asymptotically linear, i.e.,
% \be \label{htheta}
% \sqrt{n}(\htheta\tecr{(D)} - \theta_P) = {1 \over \sqrt{n}}\sum_{i=1}^{n}\psi(X_i, Y_i) + o_P(1),
% \ee
\be \label{htheta}
\sqrt{n}(\htheta - \theta_P) = {1 \over \sqrt{n}}\sum_{i=1}^{n}\psi(X_i, Y_i) + o_P(1),
\ee
where $\psi$ is an influence function with mean zero and finite variance. 
\end{itemize}

\begin{rema}
Condition~(\ref{C2}) %and 5 jointly 
ensures the well-posedness of the covariance matrix $\Sigma_{EL}$. 
It prevents numerical instabilities caused by vanishing denominators or irregular behavior in the numerator terms.
%Condition~(\ref{C5}) guarantees that  $\Sigma_{EL}$ is finite-dimensional, thereby satisfying the regularity conditions of the central limit theorem (CLT).
\end{rema}

\begin{rema}
If $\mathcal{G}$ is a Vapnik-Chervonenkis (VC) class, then the conditions (\ref{C3})--(\ref{C4}) ensures %equivalently
that function classes of indicators indexed by subpopulations in  $\mathcal{G}$ satisfy the Donsker property. 
Specifically, the function class $\mathcal{F} = \{ \mathbbm{1}_{G}  \mid  G \in \mathcal{G} \}$ is a $P$-Donsker class.
This means that the empirical process indexed by $f \in \mathcal{F}$ converges in distribution to a tight Gaussian limit in  $\ell_{\infty}(F)$. Formally, $\sqrt{n}(\mathbb{P}_n - P )[\cdot] \dto \mathbb{G}[\cdot]$ where the limiting process $f \mapsto \mathbb{G}[f ]$ is a Gaussian process that is also a tight Borel-measurable element of  $\ell_{\infty}(F )$. 
If $\mathcal{F}$ is a $P$-Donsker class,  then it is also $P$-Glivenko-Cantelli \citep{vanderVaart2000}, i.e., $\sup_{f \in \mathcal{F} } |(\mathbb{P}_n - P)[f]| \pto 0$. 
\end{rema}

\begin{rema}\label{rema-candes2}
Condition~(\ref{C6}) is satisfied by any estimator expressible as a differentiable function of averages, i.e., $\hat{\theta}:= f(\sum_i \upsilon(x_i, y_i)/n)$ for some differentiable function $f$ and known features $\upsilon: \mathcal{X} \times \mathcal{Y} \to \mathbb{R}^k$. In this paper, we most often compare group-wise performance to the population average $\theta_P = E_P [M(h(X), Y)]$, and consequently, $\htheta = {1 \over n}  \sum_{i=1}^n M(h (x_i), y_i)$ trivially satisfies this assumption. 
% \tecr{When $\theta_P = E_P [M(f(X), Y)  \mid G_2]$, and consequently, $\htheta(D) = {1 \over |G_2|} \sum_{i \in G_2}M_i$. }
Asymptotic linearity is a canonical assumption for Gaussian approximation methods in statistics.
It is also satisfied by any $M$ or $Z$-estimator, e.g., if $\htheta$ is the empirical risk minimizer for a smooth convex loss \citep{vanderVaart2000, LehmannRomano2005}. 
\end{rema}

%We now state the first main result.
Now we are ready to state the main results.
%\noindent
% \begin{theo}\label{theoEL}
% Suppose that conditions (\ref{C1}) to (\ref{C6}) hold. % and the null hypothesis $H_0 : \bepsilon = \bepsilon_0$.
% Let $\bepsilon_0$ be the true disparity. Then as $n\to \infty$,
% \[
% \ell_{EL}(\bepsilon_0)\tod \chi^2_{m},
% \]
% where $\chi^2_{m}$ is a chi-squared distributed random variable with $m$ degrees of freedom.
% \end{theo}
\begin{theo}\label{theoEL}
Suppose that conditions (\ref{C1}) to (\ref{C5}) hold. 
Let $\bepsilon^*$ be the true disparity.
Then, $$\ell_{EL}(\bepsilon^*;\theta_P)\tod \chi^2_{m}\quad\text{as}\quad n\to \infty,$$ 
where %$\bepsilon^*$ is the true disparity and 
$\chi^2_{m}$ is a chi-squared distributed random variable with $m$ degrees of freedom.
\end{theo}
Let $z_{\alpha}(m)$ denote the upper $\alpha$-quantile, i.e. $\mathbb{P} (\chi^2_{m} \ge z_{\alpha}(m)) = \alpha$ for $0 < \alpha < 1$. 
%Following from Theorem \ref{theoEL}, an asymptotically valid ($1-\alpha$)-level confidence region for $\bepsilon_0$ is  
By Theorem \ref{theoEL}, 
an asymptotically confidence region for $\bepsilon$ is
\bea\label{ci-EL}
\left\{\bepsilon : \ell_{EL}(\bepsilon;\theta_P) \le z_{\alpha}(m)\right\}. 
\eea
% By the duality, this confidence region yields a corresponding hypothesis testing for the true disparity $\bepsilon_0$. 
% Specifically, to test the hypothesis \( H_0: \bepsilon = \bzero \), we first construct a confidence region for \( \bepsilon_0 \) and then examine whether \( \bzero\) falls within this region. If \( \bzero\) is contained within the region, we fail to reject the null hypothesis; otherwise, we reject it.
% On the other hand, we can use $p$-values to determine whether to reject the null hypothesis. The $p$-value is calculated as:
This confidence region yields a corresponding hypothesis testing for the true disparity. %$\bepsilon_0$, i.e. 
%the hypothesis \( H_0: \bepsilon = \bzero \,\, \text{v.s.}\,\, H_1: \bepsilon \neq \bzero\). 
If \( \bzero\) is contained within the region (\ref{ci-EL}), we fail to reject $H_0: \bepsilon = \bzero$; otherwise, we reject it.
The corresponding $p$-value is calculated as:
\bea \label{pEL}
p^{EL} = \mathbb{P} \{ \chi^2_m \geq \ell_{EL}(\bzero;\theta_P)\} = 1 - F_m(\ell_{EL}(\bzero;\theta_P))
\eea
where \( F_m(\cdot) \) denotes the cumulative distribution function of $\chi^2_m$. %If $ p^{EL} \geq \alpha $, we fail to reject the null hypothesis; otherwise, we reject it.

\begin{algorithm}
\caption{The EL Certification under the null hypothesis $H_0: \bepsilon = \bzero$ }\label{algorithm1}
\label{algo-EL_certification}
\begin{algorithmic}[1]
\Require Subpopulation set $\mathcal{G}$, holdout dataset $\mathcal{D}$, target $\theta_P$, confidence level $\alpha$.
\State Compute: $\bg_i(\bzero;\theta_P)$ according to \eqref{eq:3.2};
\State Compute: $\blambda$ by solving equation \eqref{lambda};
\State Compute: $\ell_{EL}(\bzero;\theta_P)$;
\State Compute: $p^{EL}$ as in \eqref{pEL};
\If{$p^{EL} \geq \alpha$}
    \State \Return Model fairness certified
\Else
    \State \Return Model unfairness certified
\EndIf
\end{algorithmic}
\end{algorithm}

Further, if $\theta_P$ is not known a priori, we are only interested in the parameter $\epsilon_G$, and $\theta_P$ is a nuisance parameter. 
For instance, the target $\theta_P$ we compare against may be the population average or the occurrence probability of a sensitive subpopulation, which is typically unknown in practice.
One approach is to replace the nuisance parameter in the estimating function by its consistent estimator and then construct the constraint in the empirical likelihood function according to the form of the estimating function. This is the so-called plugged-in empirical likelihood method. 
We assume that it is possible to use the holdout dataset to obtain a consistent estimator $\htheta$. 
We omit the argument specifying the data set used to estimate $\htheta$ when it is clear from context. 
The estimator $\htheta$ is assumed to satisfy Condition~(\ref{C6}), which requires asymptotic linearity as specified in (\ref{htheta}). 
%It can be shown that Condition~(\ref{C6}) holds for any $\htheta$ specified in the main text.
%Our generalization 
This condition enables auditing even if $\htheta$ is more complicated, e.g., it is the solution to some maximum likelihood estimation problem.
In addition, Condition~(\ref{C6}) implies that $\htheta - \theta_P = O_P({1 \over \sqrt{n}})$. Thus, we replace $\theta_P$ with $\htheta$, and have the following result.
\begin{theo}\label{theoPIEL}
Suppose that conditions (\ref{C1}) to (\ref{C6}) hold. Let $\bepsilon^*$ be the true disparity. Then,
\[
\ell_{EL} (\bepsilon^*;\hat\theta )\tod \chi^2_{m} \quad \text{as} \quad n\to\infty,
\]
where %$\bepsilon^*$ is the true disparity and 
$\chi^2_{m}$ is a chi-squared distributed random variable with $m$ degrees of freedom.
\end{theo}

The above theorem demonstrates that when the $\theta_P$ is unknown, the plug-in EL still achieves the same limiting distribution as stated in Theorem \ref{theoEL}, thereby providing a theoretical guarantee for practical applications.
Therefore, in the subsequent content, we will no longer emphasize that certain quantities or functions depend on $\theta_P$ and $\hat \theta$. For simplicity, we will remove $\theta_P$ and $\hat \theta$ from the notations.

\subsection{EEL certification}

%In practical applications, the empirical likelihood method 
Algorithm~\ref{algorithm1} involves solving an implicit equation for the Lagrange multiplier $\lambda$ in \eqref{lambda}, for which no closed-form solution exists in general. This results in substantial growth in computation when the number of groups $m$ increases, as demonstrated in Table \ref{tab-Time-model1} of the simulation section.
To circumvent this computational difficulty, we adopt the empirical Euclidean likelihood method proposed by \cite{Luo1994}. 
This approach not only provides a closed-form expression for the likelihood ratio statistic but also preserves the same asymptotic properties of the conventional empirical likelihood method. 
In the following part, we construct an empirical Euclidean likelihood (EEL) procedure for testing group-wise performance disparities. 

Actually, the quantity $\log \prod_{i=1}^n n p_i = \log \prod_{i=1}^n p_i  - \log \prod_{i=1}^n \frac{1}{n} $ can be interpreted as a likelihood distance between the probability vector $(p_1, p_2, \dots, p_n)^\top$ and the uniform vector $(\frac{1}{n}, \frac{1}{n}, \dots, \frac{1}{n})^\top$. This interpretation motivated \cite{Owen1991} to replace $\log \prod_{i=1}^n n p_i $  with an euclidean distance $ -\frac{1}{2} \sum_{i=1}^n\left( p_i - \frac{1}{n} \right)^2 $. This idea was subsequently suggested by \cite{QinLawless1994} to semi-parametric models. For further details on the strong consistency and asymptotic normality of the parameter estimators derived under the EEL framework, one can refer to \cite{Luo1994}.

Following \cite{Luo1994}, for a given $\bepsilon$, the empirical Euclidean log-likelihood ratio function is 
\bea \label{EEL}
\tilde{\ell}_{EEL}(\bepsilon) = \sup_{p_i,1\le i\le n} \left\{ - \frac{1}{2} \sum_{i=1}^n\left( p_i - \frac{1}{n} \right)^2 \ \middle| \   p_i\geq 0, \,\, \sum_{i=1}^np_i = 1, \,\, \sum_{i=1}^n p_i  \bg_i(\bepsilon) = 0 \right\}.  \nn
\eea
% \bea \label{logREEL}
% \tilde{\ell}_{EEL}(\bepsilon) = \sup_{p_i,1\le i\le n} \left\{ - \frac{1}{2} \sum_{i=1}^n\left( p_i - \frac{1}{n} \right)^2 \ \middle| \   p_i\geq 0, \quad \sum_{i=1}^np_i = 1, \quad \sum_{i=1}^n p_i  \bg_i(\bepsilon) = 0 \right\}.  \nn
% \eea
This optimization problem admits a closed-form solution,
\bea \label{eel-solution}
\widehat{p}_i = \frac{1}{n} + \frac{1}{n}\bar{\bg}^\top(\bepsilon) s^{-1}(\bepsilon) \left[\bar{\bg}(\bepsilon) - \bg_i(\bepsilon) \right], \quad 1 \leq i \leq n,  
\eea
where %$\bar{\bg}({\bepsilon}) =n^{-1}\sum_{i=1}^n \bg_i(\bepsilon)$ and
$$\bar{\bg}({\bepsilon}) =n^{-1}\sum_{i=1}^n \bg_i(\bepsilon) \quad \text{and} \quad
s(\bepsilon) = \frac{1}{n} \sum_{i=1}^n  \left[ \bg_i(\bepsilon)  - \bar{\bg}({\bepsilon}) \right]\left[\bg_i(\bepsilon) - \bar{\bg}({\bepsilon}) \right]^\top. 
$$
% \bea \label{barg}
% \bar{\bg}({\bepsilon;\theta_P}) =n^{-1}\sum_{i=1}^n \bg_i(\bepsilon;\theta_P) 
% \quad \text{and} \quad
% s(\bepsilon;\theta_P) = \frac{1}{n} \sum_{i=1}^n  \left[ \bg_i(\bepsilon;\theta_P)  - \bar{\bg}({\bepsilon;\theta_P}) \right]\left[\bg_i(\bepsilon;\theta_P) - \bar{\bg}({\bepsilon;\theta_P}) \right]^\top.
% \eea
%The empirical Euclidean log-likelihood ratio at $\bepsilon$ is that
% {\color{red}\bea \label{ratioEEL}
% \ell_{EEL}(\bepsilon) = -2 \tilde{\ell}_{EEL}(\bepsilon) . \nn
% \eea}
And then the empirical Euclidean log-likelihood ratio at $\bepsilon$ is that
\(\ell_{EEL}(\bepsilon)  = -2 \tilde{\ell}_{EEL}(\bepsilon) \). We have the following result.
%\noindent
\begin{theo}\label{theoEEL}
Suppose that conditions (\ref{C1}) to (\ref{C5}) hold.
Let $\bepsilon^*$ be the true disparity. 
Then %as $n\to \infty$,
\[
\ell_{EEL}(\bepsilon^*)\tod \chi^2_{m} \quad \text{as} \quad n\to \infty,
\]
where 
$\chi^2_{m}$ is a chi-squared distributed random variable with $m$ degrees of freedom.
\end{theo}

By Theorem \ref{theoEEL},  an EEL-based confidence region for $\bepsilon$ %with asymptotical coverage probability $1 - \alpha$ can be constructed as  
is given by
\bea\label{ci-eel}
\left\{\bepsilon : \ell_{EEL}(\bepsilon) \le z_{\alpha}(m)\right\}. 
\eea
% Let $z_{\alpha}(m)$ denote the $\alpha$-quantile satisfying $\mathbb{P} (\chi^2_{m} \ge z_{\alpha}(m)) = \alpha$ for $0 < \alpha < 1$. 
Similarly, this confidence region can be applied to hypothesis testing concerning the target disparity $\bzero$. 
Meanwhile, we can use $p$-values to determine whether the null hypothesis $H_0:\bepsilon=\bzero$ is rejected or not. The EEL-based $p$-value is calculated as:
\[
p^{EEL} = \mathbb{P} \{ \chi^2_m \geq \ell_{EEL}(\bzero)\} = 1 - F_m(\ell_{EEL}(\bzero))
\]
where \( F_m(\cdot) \) denotes the cumulative distribution function of $\chi^2_m$. If $ p^{EEL} \geq \alpha $, we fail to reject the null hypothesis; otherwise, reject it. 
%Analogously to the plug-in EL method, the preceding theorem continues to hold when $\theta_P$ is replaced by its estimator $\htheta$.
The EEL certification algorithm is given in the Supplemental Material (Algorithm~\ref{algo-EEL_certification}).

%============================================================
\section{Flagging performance disparities} \label{sec-flagging}

The previous section investigates collective fairness assessment across all subgroups. 
To identify which specific subgroups are discriminated against,  
we develop a methodology to flag individual subgroups with performance disparities in this section. 
This setting can be formulated as a multiple testing problem. It may happen that when testing many subgroups simultaneously, some may be falsely flagged as unfair simply by chance. 
To address this, we aim to achieve asymptotic control of the false flagging rate (FFR), which bounds the expected proportion of false-flagged subgroups. 
As established in \cite{BenjaminiHochberg1995}, FFR control requires:
\[
\mathbb{E}\left[ \frac{\left| \text{falsely flagged } G \in \mathcal{G} \right|}{\left| \text{flagged } G \in \mathcal{G} \right| \vee 1} \right] \leq \alpha \quad \text{as } n \to \infty.
\]
A widely adopted strategy for achieving FFR control is the Benjamini–Hochberg (BH) procedure, which uses $p$-values for each subgroup while controlling the proportion of false discoveries. 
Our flagging method employs $p$-values derived from the asymptotic theory of empirical likelihood for each subgroup. Subgroups whose corresponding null hypotheses are rejected are flagged and deemed unfair.

To broaden the applicability of empirical likelihood-based fairness auditing, we consider %four frameworks targeting different practical scenarios:
two different practical scenarios:

\begin{enumerate}
    \item[(1)] \( H_0(G): \epsilon_{G} = \epsilon_0 \) versus \( H_1(G): \epsilon_G \neq \epsilon_0 \): detecting subgroups with disparities significantly different from a specific level.

    % \item \( \epsilon_G \geq \epsilon_0 \) versus \( \epsilon_G < \epsilon_0 \): flagging disadvantaged subgroups whose disparities exceed the tolerance threshold.

    \item[(2)] \( H_0(G): \epsilon_G \leq \epsilon_0 \) versus \( H_1(G): \epsilon_G > \epsilon_0 \): flagging over-protected subgroups with disparities below the tolerance level.

    % \item \( \epsilon_1 \leq \epsilon_G \leq \epsilon_2 \) versus \( \epsilon_G < \epsilon_1 \) or \( \epsilon_G > \epsilon_2 \): flagging subgroups outside an acceptable tolerance range.
\end{enumerate}

Scenario (1) represents a simple hypothesis under equality constraints, while scenario (2) represents composite hypotheses under inequality constraints. 
The choice among these frameworks depends on the application. %The choice of framework depends on the application domain. 
%In hiring or lending, higher acceptance rates are desirable; a subgroup is considered disadvantaged when its acceptance rate falls below a specified level, corresponding to framework (2). 
In recidivism prediction, a lower predicted risk is preferable; a subgroup is disadvantaged when its predicted risk exceeds the threshold, corresponding to Scenario (2). 
%In medical dosing, both excessively high and excessively low values may indicate unfairness, as over-treatment and under-treatment can both cause harm; framework (4) addresses such cases. 
Two extended statistical tests are considered. One test of \( H_0(G): \epsilon_G \ge \epsilon_0 \) versus \( H_1(G): \epsilon_G < \epsilon_0 \), which flags disadvantaged subgroups whose disparities exceed the tolerance threshold, and another test of \( H_0(G): \epsilon_1 \leq \epsilon_G \leq \epsilon_2 \) versus \( H_1(G): \epsilon_G < \epsilon_1 \) or \( \epsilon_G > \epsilon_2 \), which flags subgroups outside an acceptable tolerance range. 
%The discussion of two extended tests are presented in the Supplemental Material (Section~\ref{appsec-extended-tests}).

We will show that the limiting distributions of the empirical likelihood ratio tests for framework (2) converge to a weighted mixture of chi-square distributions. 
It is worth mentioning that, unlike \cite{ChenShi2011}, our group-wise estimating equations involve the product of the mean model and indicator functions, which leads to a more intricate proof. 
Specifically, we must account for the asymptotic normality of group-wise means and the probability that the random variables $(X, Y)$ fall within the set $G$. 
The details can be found in proofs given in the Supplemental Material (Sections~\ref{appsec-proof-sec4}).
%Our asymptotic powers depend on this probability, as shown in Corollaries~\ref{coro2}--\ref{coro4}.

% To facilitate the discussion, we define some useful notation. Given constants $\epsilon_0, \epsilon_1, \epsilon_2 \in \mathbb{R}^1$ satisfying $\epsilon_1 < \epsilon_2$, we define the sets $\Omega_0 = \{\epsilon : \epsilon = \epsilon_0\}$, $\Omega_1 = \{\epsilon : \epsilon \ge \epsilon_0\}$, $\Omega_2 = \{\epsilon : \epsilon \le \epsilon_0\}$, $\Omega_3 = \{\epsilon : \epsilon_1 \leq \epsilon \leq \epsilon_2\}$, and $\Omega_4 = \{\epsilon : \epsilon \in \mathbb{R}^1\}$. Furthermore, we denote the hypotheses $\epsilon \in \Omega_i - \Omega_j$ as $H_{ji}$ (where $0 \leq j < i \leq 4$). 

\subsection{Equality constraints}\label{sec-4.1}
%{\color{red}It seems to me that we already demonstrate the result in this part in Theorem \ref{theoEL2}.}
This subsection focuses on the scenario (1), testing the hypothesis 
$$H_0(G): \epsilon_G = \epsilon_0 \quad \text{versus} \quad H_1(G): \epsilon_G \neq \epsilon_0.$$ %against the alternative $H_1: \epsilon_G \neq \epsilon_0$.
We derive the asymptotic distribution of the empirical likelihood ratio statistic for each subgroup's performance disparity, from which we obtain the corresponding $p$-values.
To facilitate subsequent extensions to other testing problems, it is necessary to reintroduce relevant notation. 

Following the notation in Section \ref{certifying}, we specialize to the single-group case ($m=1$) and get the empirical likelihood function for the group-wise performance disparity $\epsilon_G$:
\bea \label{L}
L_{EL}(\epsilon_G)=\sup_{p_i,1\le i\le n} \left\{ \prod^n_{i=1} p_i \ \Big| \ p_i\geq 0, \,\, \sum_{i=1}^np_i = 1, \,\, \sum_{i=1}^n p_i  g_i(\epsilon_G) = 0 \right\},
\eea
where %$$g_i(\epsilon_G) = g(x_i,y_i;\epsilon_G,\theta_P)=\left[M(h(x_i),y_i) - \theta_P - \epsilon_G \right] \cdot \mathbbm{1}_G(x_i) .$$
\bea \label{gi_epsilon_G}
g_i(\epsilon_G) = g(x_i,y_i;\epsilon_G,\theta_P)=\left[M(h(x_i),y_i) - \theta_P - \epsilon_G \right] \cdot \mathbbm{1}_G(x_i) .
\eea
% Using the method of Lagrange multipliers, the maximization in (\ref{L}) is achieved by:
% \bea \label{pi_epsilon_G}
% \hat{p}_i = \hat{p}_i(\epsilon_G) = \frac{1}{n\left(1 + \lambda g_i(\epsilon_G)\right)}, \quad 1 \leq i \leq n, 
% \eea
% where the multiplier $\lambda \in \mathbb{R}^1$ satisfies the equation:
% \bea \label{lambda_epsilon_G}
% \sum_{i=1}^n \frac{g_i(\epsilon_G)}{1 + \lambda g_i(\epsilon_G)} = 0. 
% \eea

% \begin{lemm} \label{lemm_ell}
% The empirical log-likelihood function \( \dot{\ell}_{EL}(\epsilon_G) \) is upper convex with respect to \( \epsilon_G \).
% \end{lemm}

% \begin{lemm} \label{lemm_ell}
% The function \( \log L_{EL}(\epsilon_G;\theta_P) \) is upper convex with respect to \( \epsilon_G \).
% \end{lemm}

The following lemma establishes that the the empirical likelihood function has a unique optimal solution and is locally monotonic at non-optimal points.
\begin{lemm} \label{lemm_ell}
The function \( L_{EL}(\epsilon_G) \) is upper convex with respect to \( \epsilon_G \).
\end{lemm}
Lemma \ref{lemm_ell} implies that, without fairness constraints imposed on $\epsilon_G \in \mathbb{R}^1$,  $L_{EL}(\epsilon_G) $ attains its maximum at 
%$$\hepsilon_G = |G|^{-1}\sum_{i = 1}^{n}L_i \mathbbm{1}_G(x_i) - \theta_P,$$ 
%= \frac{\sum_{i = 1}^{n}L_i \mathbbm{1}_G - \theta_P}{\sum_{i = 1}^{n} \mathbbm{1}_G(x_i)},$$
\bea \label{hat_epsilon_G}
\hepsilon_G = |G|^{-1}\sum_{i = 1}^{n}L_i \mathbbm{1}_G(x_i) - \theta_P,
\eea 
where $\lambda = 0$, $p_i(\hepsilon_G) = n^{-1}$ ($1 \leq i \leq n$), and $L_{EL}(\hepsilon_G) = 1$. 
For the simple null hypothesis $H_0(G): \epsilon_G = \epsilon_0$, we employ the  empirical log-likelihood ratio as the test statistic:
%statistic can be written as
% \bea 
%  T_{=} = \ell_{EL}(\epsilon_G) = -2 \log \frac{\sup_{\epsilon_G \in \Omega_0} L_{EL}(\epsilon_G)}{\sup_{\epsilon_G \in \Omega_4} L_{EL}(\epsilon_G)} = -2 \log \frac{ L_{EL}(\epsilon_0)}{ L_{EL}(\hepsilon_G)}  \nn
% \eea
% \bea \label{ell_epsilon_G}
%  T(G) := 
%  = -2 \log \frac{\sup_{\epsilon_G = \epsilon_0} L_{EL}(\epsilon_G)}{\sup_{\epsilon_G \in \mathbb{R}} L_{EL}(\epsilon_G)} = -2 \log \frac{ L_{EL}(\epsilon_0)}{ L_{EL}(\hepsilon_G)}
% \eea
\begin{equation}
    \begin{split}
        T(G) %&:= -2 \log \frac{\sup_{H_0(G)} L_{EL}(\epsilon_G)}{\sup_{H_0(G)+H_1(G)} L_{EL}(\epsilon_G)}  \\
             &:= -2 \log \frac{\sup_{\epsilon_G = \epsilon_0} L_{EL}(\epsilon_G)}{\sup_{\epsilon_G \in \mathbb{R}} L_{EL}(\epsilon_G)}  = -2 \log \frac{ L_{EL}(\epsilon_0)}{ L_{EL}(\hepsilon_G)} \nn
    \end{split}
\end{equation}

Based on the expressions above, 
%we can readily extend the fairness certification to various inequality constraints and establish their asymptotic distributions. 
Noted that $T(G) = \ell_{EL}(\bepsilon_0; \theta_P)$ with $\{G\}=\mathcal{G}$.
Hence a direct application of Theorem \ref{theoEL} with $m=1$ immediately yields the following result. 
\noindent
\begin{coll}\label{coro1}
%\tecr{Assume that $\mathbb{P}(G) $ and $Var (L | G)$ are bounded away from $0$ for any  $G \in \mathcal{G}$, $\theta_P$ is a-priori known.}
Assume that conditions (\ref{C1}) to (\ref{C2}) hold. 
Then, under the null hypothesis $H_0(G): \epsilon_G = \epsilon_0$, it follows that, %as $n \to \infty$,
\[
T(G)  \tod \chi^2_{1} \quad \text{as} \quad n \to \infty,
\]
where $\chi^2_{1}$ is a chi-squared distributed random variable with $1$ degrees of freedom.
\end{coll}

From this result, the EL based confidence region for $\epsilon_{G}$ with asymptotically correct coverage probability $1 - \alpha$ can be constructed as
$
 \{ \epsilon_{G}: T(G) \leq z_{\alpha}(1) \}.
$
Here $z_{\alpha}(1)$ denote the upper $\alpha$-quantile of $\chi^2_{1}$.
%This confidence interval can be directly applied to hypothesis testing concerning the true disparity $\epsilon_{G_0}$. For instance, to test the hypothesis \( H_0(G): \epsilon_G = 0 \), we first construct a confidence interval for \( \epsilon_G \) and then examine whether \( 0 \) falls within this interval. If \( 0 \) is contained within the interval, we fail to reject the null hypothesis; otherwise, we reject it.
On the other hand, 
%we can use $p$-values to determine whether to reject the null hypothesis. 
the $p$-value is calculated as:
\bea \label{p04}
p = \mathbb{P} \{ \chi_1^2 > T(G)\} = 1 - F_1( T(G) ), 
\eea
where \( F_1(\cdot) \) denotes the cumulative distribution function of $\chi_1^2$. 
%If $ p_{04} \geq \alpha $ , we fail to reject the null hypothesis; otherwise, we reject it.

% \begin{algorithm}
% \caption{EL $p$-value under the null hypothesis $H_0(G) : \epsilon_G = \epsilon_0$ }
% \label{algo-p04}
% \begin{algorithmic}[1]
% \Require Subpopulation set $\mathcal{G}$, holdout dataset $\mathcal{D}$, target $\theta_P$, confidence level $\alpha$, parameter $\epsilon_0$
% \State Compute: $g(\mathcal{D}; \epsilon_0)$ as in \eqref{L};
% \State Compute: $\lambda$ and $p_i, 1 \le i \le n$, as in \eqref{lambda_epsilon_G} and \eqref{pi_epsilon_G} respectively;
% \State Compute: $ R_{EL}(\epsilon_G)$ as in \eqref{R_epsilon_G};
% \State Compute: $T_{04} = -2 \log R_{EL}(\epsilon_G)$;
% \State Compute: $p_{04}$ as in \eqref{p04};
% \State \Return $p_{04}$
% \end{algorithmic}
% \end{algorithm}

\subsection{Inequality constraints} \label{one-sided}

% Next, this subsection focuses on the following tests: $H_1$ against $H_{14}$, $H_2$ against $H_{24}$ and $H_3$ against $H_{34}$.
% That is,  we  test the one-sided hypotheses \( \epsilon_G \geq \epsilon_0 \) and  \( \epsilon_G \leq \epsilon_0 \), as well as the two-sided hypothesis of the form \( \epsilon_1 \leq \epsilon_G \leq \epsilon_2 \).

% \subsubsection{One-sided hypotheses on disparity} \label{One-sided}

Consider the one-side hypothesis testing problem in Senerio (2):
$$H_0(G):  \epsilon_G \le \epsilon_0 \quad \text{versus} \quad H_1(G):  \epsilon_G > \epsilon_0. $$
According to the arguments in Section~\ref{sec-4.1}, The empirical likelihood ratio test statistic is given by
\[
T(G) :=  -2 \log \frac{\sup_{\epsilon_G \le \epsilon_0}L_{EL}(\epsilon_G)}{\sup_{\epsilon_G \in \mathbb{R}^1} L_{EL}(\epsilon_G)}.
\]
% Consider two types of one-sided testing problems.
% Firstly,  for $  H_1 :  \{ \epsilon_G \ge \epsilon_0 \}$ versus $  H_{14} : \{ \epsilon_G < \epsilon_0 \}$, the empirical likelihood ratio is given by  
% \[
% T_{14} := -2 \log \frac{\sup_{\epsilon_G \in \Omega_1} L_{EL}(\epsilon_G)}{\sup_{\epsilon_G \in \Omega_4} L_{EL}(\epsilon_G)} = -2 \log \frac{\sup_{\epsilon_G \ge \epsilon_0}L_{EL}(\epsilon_G)}{\sup_{\epsilon_G \in \mathbb{R}^1} L_{EL}(\epsilon_G)}.
% \]
The application of  Lemma \ref{lemm_ell} yields that \(  L_{EL}(\epsilon_G) \) attains a unique maximum over \( \mathbb{R}^1 \) at \( \hepsilon_G = |G|^{-1}\sum_{i = 1}^nM_i \mathbbm{1}_G - \theta_P \), and then
\[
\sup_{\epsilon_G \leq \epsilon_0}  L_{EL}(\epsilon_G) = 
\begin{cases}
L_{EL}(\epsilon_0), & \ \mathrm{if} \  \  \hepsilon_G >\epsilon_0,\\
           n^{-n}, & \ \mathrm{if} \  \ \hepsilon_G \le \epsilon_0.
\end{cases}
\]
Hence, it is clear that 
% \[
% T_{14} = \left( -2 \log \frac{L_{EL}(\epsilon_0)}{L_{EL}(\hepsilon_G)} \right) I(\hepsilon_G < \epsilon_0)  +  \left( -2 \log \frac{L_{EL}(\hepsilon_G)}{L_{EL}(\hepsilon_G)} \right) I(\hepsilon_G \ge \epsilon_0) = T_{04} I(\hepsilon_G < \epsilon_0).
% \]
\[
T(G) = \left( -2 \log \frac{L_{EL}(\epsilon_0)}{L_{EL}(\hepsilon_G)} \right) \mathbbm{1}(\hepsilon_G > \epsilon_0).  %= T_{=}(G) \mathbbm{1}(\hepsilon_G < \epsilon_0).
\]
%We now state the main results. 
Under the null hypothesis, the limiting distribution of $T(G)$ depends critically on the location of the true disparity $\epsilon_G^*$. 
The following theorems characterize the asymptotic behaviors of the statistic. 
%From Theorem \ref{theo5}, when the true parameter lies on the boundary of the null hypothesis, the empirical likelihood ratio statistic follows a mixture chi-square distribution. Specifically, $ T_{14} $ admits a two-point mixture, taking the value 0 with probability 0.5 and following a $ \chi^2_{1} $ distribution with probability 0.5.
% \begin{theo}\label{theo5}
% Assume that $\mathbb{P}(G) $ and $Var (L | G)$ are bounded away from $0$ for any  $G \in \mathcal{G}$, $\theta_P$ is a-priori known. 
% Then, under the null hypothesis $H_1(G): \epsilon_G \ge \epsilon_0$ with $\epsilon^*$ being the true    group-wise disparity, it follows that as $n \to \infty$,
% \[
% T_{14} \tod \frac{1}{2}\chi^2_{0} + \frac{1}{2}\chi^2_{1} \quad \mathrm{for} \ \epsilon^* = \epsilon_0 ,
% \] 
% where $\chi^2_{0}$ is a degenerate random variable with mass 1 at the point zero.  
% \end{theo}

\begin{theo}\label{theo5}
Assume that conditions (\ref{C1}) to (\ref{C2}) hold. 
Then, under the null hypothesis $H_0(G): \epsilon_G \le \epsilon_0$, it follows that for the least favorable case $\epsilon_G = \epsilon_0$,
\[
T(G) \tod \frac{1}{2}\chi^2_{0} + \frac{1}{2}\chi^2_{1} \quad \mathrm{as} \quad n \to \infty,
\] 
where $\chi^2_{0}$ is a degenerate random variable with mass 1 at the point zero.  
\end{theo}

Theorem \ref{theo5} shows that at the boundary of the null hypothesis, the limiting distribution of the empirical likelihood ratio statistic $ T(G) $ is a mixture: with probability 0.5 it equals 0, and with probability 0.5 it follows a $ \chi^2_{1} $ law.

\begin{theo}\label{theo6}
Assume that conditions (\ref{C1}) to (\ref{C2}) hold.
%$\mathbb{P}(G) $ and $Var (L | G)$ are bounded away from $0$ for any  $G \in \mathcal{G}$, $\theta_P$ is a-priori known. 
Then, under the null hypothesis $H_0(G): \epsilon_G \le \epsilon_0$, we have
\bea
\lim_{n \to \infty} \mathbb{P}\left\{ T(G) > z_{2\alpha}(1) \right\} =
\begin{cases}
\alpha,  \,\, \mathrm{if} \  \  \epsilon_G^* = \epsilon_0; \\
0,           \,\, \mathrm{if} \  \  \epsilon_G^* < \epsilon_0, \nn
\end{cases}
\eea 
where $z_{2\alpha}(1)$ is the upper $2\alpha$-quantile of $\chi^2_{1}$ distribution.
\end{theo}

Theorem \ref{theo6} establishes the Type I error control properties of the testing procedure. 
In the boundary case, the probability of Type I error converges asymptotically to the nominal level \( \alpha \), demonstrating proper error control. 
In the case of the interior point of the null hypothesis, the Type I error probability converges to 0 asymptotically, indicating that the test rarely rejects the null hypothesis when it is true.
In the context of fairness auditing, this conservative behavior implies that when the true performance disparity strictly less than \( \epsilon_0 \), the risk of falsely rejecting the null hypothesis diminishes to zero. This property aligns with practical expectations and provides desirable safety guarantees for real-world applications.

According to \cite{LehmannRomano2005}, Theorem \ref{theo6} demonstrates that $\epsilon_0$ is the least favorable point under the null hypothesis.
Hence the limiting distribution of $T(G)$ at the boundary point $\epsilon_0$ is employed to determine the critical value for the rejection region. 
%More specifically, the null hypothesis $H_1$ is rejected if $T(G) > c_\alpha$ for a given significance level $\alpha$, \tecr{where $c_\alpha$ is defined by $\mathbb{P} \left( \frac{1}{2} \chi^2_0 + \frac{1}{2} \chi^2_1 \geq c_\alpha \right) = \alpha$. It can be inferred that $c_\alpha = z_{2\alpha}(1)$, where $z_{\alpha}(1)$ satisfy $\mathbb{P} (\chi^2_{1} \ge z_{\alpha}(1)) = \alpha$ for $0 < \alpha < 1$.??}
Then the EL-based confidence region for $\epsilon_{G}$ with asymptotically coverage $1 - \alpha$ can be constructed as
\[
\{ \epsilon_{G}: T(G)\leq z_{2\alpha}(1) \}.
\]
In practice, the $p$-values are computed conservatively under the boundary $\epsilon_G = \epsilon_0$, which represents the least favorable case for rejection. Therefore, for any observed $T(G) > 0$, the $p$-value is given by
\bea 
p = \mathbb{P} \left\{ \frac{1}{2} \chi^2_0 + \frac{1}{2} \chi^2_1 > T(G)\right\} = \frac{1}{2} \big( 1 - F_1( T(G) ) \big), \label{p24}
\eea
where \( F_1(\cdot) \) is the cumulative distribution function (CDF) of $\chi_1^2$. 
%If $ p_{14} \geq \alpha $ , we fail to reject the null hypothesis; otherwise, we reject it. 
%In the case of $T(G) = 0$, we have $ p = 1 $ , leading to non-rejection of the null hypothesis.

The next theorem examines the asymptotic local power of the test statistic $T(G)$.  
\begin{coll}\label{coro2}
For the test $H_0(G):  \epsilon_G \le \epsilon_0$  versus $H_1(G):  \epsilon_G > \epsilon_0$,  
if the true disparity is \( \epsilon_G^* = \epsilon_0 + \tau n^{-1/2} \sigma \), where \( \tau > 0\) and \( \sigma^2 = { \mathbb{P}(G) }Var(L\ |\ G) \), then we have 
\[
\lim_{n \to \infty} \mathbb{P}\left( T(G) > z_{2\alpha}(1) \mid \epsilon_G^* \right) = \Phi\left( \mathbb{P}(G)  \tau - \sqrt{z_{2\alpha}(1)} \right).
\]  
Therefore, for any fixed \( \epsilon_G^* > \epsilon_0 \), $\lim_{n \to \infty} \mathbb{P}\left\{ T(G) > z_{2\alpha}(1) \mid H_1 \right\} = 1.$
\end{coll}

Two extended statistical tests are presented in the Supplemental Material (Section~\ref{appsec-extended-tests}).

\subsection{False flagging rate control}

% Building upon the discussion of $p$-values under different hypothesis testing frameworks, we propose an empirical likelihood-based BH (ELBH) procedure. 
Building upon the discussion of testing group-wise performance disparity, we further control overall false flagging rate via adjusting $p$-value thresholds.
We consider multiple testing of group-wise performance disparity: 
$H_{0}(G) \,\, \text{versus} \,\, H_{1}(G)$ 
% $$H_{0}(G): \epsilon_G \le \epsilon_0 \quad \text{versus} \quad H_{1}(G): \epsilon_G > \epsilon_0 $$
with $G \in \mathcal{G} = \{ G_1, \dots, G_m\}.$
For each $H_{0}(G_i) \ \text{versus} \ H_{1}(G_i)$, we construct the corresponding empirical likelihood ratio statistic and compute the empirical $p$-value as $p_i^{EL}$. 
Then we employ the Benjamini-Hochberg (BH, \cite{BenjaminiHochberg1995}) procedure %全称加引用
to control overall false flagging:
%Following the classical BH procedure,  the ELBH procedure is then applied as follows:
\begin{enumerate}
    \item Sort the empirical $p$-values in ascending order: $p_{(1)}^{EL} \le p_{(2)}^{EL} \le \dots \le p_{(m)}^{EL}$, where $m$ denotes the total number of hypotheses;
    \item Determine the maximum index $k$ such that $p_{(k)}^{EL} \le \frac{k}{m} \alpha$, where $\alpha$ is the pre-specified false flagging rate (FFR) level;
    \item Flag all subgroups whose corresponding null hypotheses are rejected, i.e., those with $p$-values less than or equal to $p_{(k)}^{EL}$, as groups exhibiting significant performance disparities.
\end{enumerate}
 
The following theorem shows that the proposed auditing procedure controls false flagging or false flagging.
 \begin{theo}[False flagging rate control] \label{theoFFR}
Assume that for all $G \in \mathcal{G}$, $\mathbb{P}(G)$ and $\text{Var}(L|G)$ are bounded away from zero, and the parameter $\theta_P$ is known a priori. Furthermore, assume that the groups $\{G\}_{G \in \mathcal{G}}$ are disjoint. Then, if the BH($\alpha$) procedure is applied to the collection $\{G\}_{G \in \mathcal{G}}$, the FFR is asymptotically controlled at level $\alpha$.
\end{theo}

For disjoint groups $\{G\}_{G \in \mathcal{G}}$, $p$-values are independent, ensuring validity.  
Theorem \ref{theoFFR} remains robust even when estimating $\theta_P$.
%We expect Theorem \ref{theoFFR} to hold robustly under certain assumption violations. For example, FFR control is maintained even when estimating $\theta_P$, as supported by ELBH experiments in Section \ref{sec-simu}.

\section{Simulations} \label{sec-simu}

Our simulation studies are divided into two parts. 
The first part empirically validates the theory and coverage guarantees of the fairness certification (see Section~\ref{certifying}) with both homoscedastic and heteroscedastic data-generating processes. 
%designs homoskedastic and heteroskedastic models to validate the theoretical performance of fairness certification methods presented in Section~\ref{certifying}. 
The second part evaluates the the statistical power and Type I error control of the unfairness flagging procedures in Section~\ref{sec-flagging}. 
%designs experiments to validate the theoretical performance of unfairness flagging methods presented in Section~\ref{sec-flagging}. 
All code required to reproduce these simulations and analyses is publicly available at \if1\anon at \url{https://github.com/Tang-Jay/EL-for-fairness-auditing}.
\else
upon request.
\fi

\subsection{Certifying}
In this section, we validate  Theorem~\ref{theoEL} and \ref{theoEEL}, and the corresponding confidence interval (\ref{ci-EL}) and (\ref{ci-eel}). 
The evaluation is based on three metrics: distribution matching, coverage probability and computational speed.
We performed 2,000 replications for all simulation studies. 
We compare our proposed methods with the bootstrap approach presented in Algorithm 1 of \cite{CherianCandes2024}.
%The results are presented in Tables \ref{tab-CP-model1-0.95} to \ref{tab-Time-model1}. 

%We initially consider a homoskedastic linear model. We sample $(X_i, Y_i)$ from  
Consider two linear models: 
\bea 
X_i  \toiid \mathrm{Unif}(0, 1), \quad Y_i  \toiid N (\beta_0X_i, 1), \quad \beta_0=2,  \label{model1}\\
X_i  \toiid \mathrm{Unif}(0, 1), \quad  Y_i  \toiid N (\beta_0X_i, X_i), \quad \beta_0=2.   \label{model2}
\eea
We adopt a partitioned interval structure where each subset $G_j$
is defined with $[ {j-1 \over m}, {j \over m}).$
The performance metric of interest is squared-error loss, i.e., $M(f(X), Y) = (Y -f (X))^2$ and the target $\theta_P = 0$.
Hence the true value of performance disparities is $\epsilon_{G_j} = 1$ for (\ref{model1}) and $\epsilon_{G_j} = {2j-1 \over 2m}$ for (\ref{model2}). 
%We then obtain $f (x) = \hat{\beta}x $ via ordinary least-squares on $n$ training points sampled from this distribution with $\beta_0 = 2$. 
%The performance metric of interest is squared-error loss, i.e., $M(f(X), Y) = (Y -f (X))^2$ and the target $\theta_P = 0$. 
%We adopt a partitioned interval structure where each subset $G_j$ is defined with endpoints in the interval $[ {j-1 \over m}, {j \over m})$ for $j = 1, \dots, m$. Correspondingly, we assign the parameter $\epsilon_{G_j} = 1$ for each such subset.
% In the second experiment, we validate our audit using a heteroskedastic linear model, 
% \bea \label{model2}
% X_i  \toiid \mathrm{Unif}(0, 1), \quad  Y_i  \toiid N (\beta_0X_i, X_i).   
% \eea
% The model $f(X)$ and metric $M(f(X), Y)$ are identical to the previous synthetic experiment. The subsets are also partitioned in a similar manner as above, but with $\epsilon_{G_j} = {2j-1 \over 2m}$.

\begin{figure}
\centering
\includegraphics[width=1\textwidth]{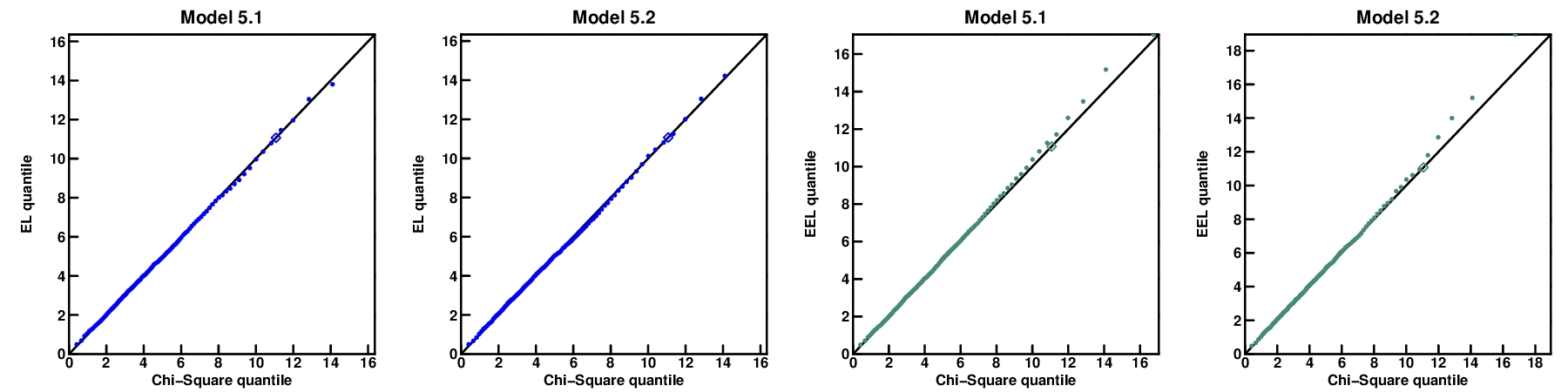}\\
\caption{Q-Q plots of $\ell_{EL}(\bepsilon^*)$ versus $\chi^2_m$ and of $\ell_{EEL}(\bepsilon^*)$ versus $\chi^2_m$ under Models \eqref{model1} and \eqref{model2} with $n=2000$ and $m=5$.}
\label{fig-QQ-EL-EEL}
\end{figure}

In terms of coverage probability, as shown in Table~\ref{tab-CP-0.95}, the coverage rates of our proposed EL and EEL methods approach the nominal level $1- \alpha = 0.95$ increasingly closely as the sample size~$n$ increases. Furthermore, the Q-Q plots in Figures~\ref{fig-QQ-EL-EEL}, which compare the EL and EEL statistics against theoretical chi-square quantiles, confirm our theoretical conclusions. These figures also indicate the need for a larger sample size as the number of groups increases.

% Table: CP 0.95 (Model 1 and Model 2)
\begin{table}[H]
\centering
\footnotesize
\caption{Coverage probabilities of the Bootstrap, EL, and EEL confidence regions under Model (\ref{model1}) and Model (\ref{model2}) at $\alpha = 0.05$.}
\label{tab-CP-0.95}
\setlength{\tabcolsep}{3pt}
\resizebox{\textwidth}{!}{%
\begin{tabular}{@{}c c *{9}{c}@{}}
\toprule
\multirow{2}{*}{Model} & \multirow{2}{*}{$m$} & \multicolumn{3}{c}{$n=2000$} & \multicolumn{3}{c}{$n=4000$} & \multicolumn{3}{c}{$n=8000$} \\
\cmidrule(lr){3-5} \cmidrule(lr){6-8} \cmidrule(lr){9-11}
 & & Bootstrap & EL & EEL & Bootstrap & EL & EEL & Bootstrap & EL & EEL \\
\midrule
\multirow{3}{*}{\eqref{model1}} & 2 & 0.9000 & {\bf 0.9475} & 0.9465 & 0.9040 & 0.9545 & {\bf 0.9520} & 0.8985 & {\bf 0.9495} & 0.9480 \\
 & 5 & 0.9285 & {\bf 0.9480} & 0.9405 & 0.9125 & {\bf 0.9505} & 0.9430 & 0.9170 & 0.9465 & {\bf 0.9485} \\
 & 10 & 0.9330 & {\bf 0.9405} & 0.9130 & 0.9290 & {\bf 0.9415} & 0.9260 & 0.9280 & {\bf 0.9510} & {\bf 0.9490} \\
\midrule
\multirow{3}{*}{\eqref{model2}} & 2 & 0.8940 & {\bf 0.9485} & 0.9460 & 0.9125 & 0.9520 & {\bf 0.9490} & 0.8985 & 0.9545 & {\bf 0.9520} \\
 & 5 & 0.9145 & {\bf 0.9510} & 0.9470 & 0.8965 & {\bf 0.9480} & 0.9440 & 0.8980 & 0.9440 & {\bf 0.9460} \\
 & 10 & 0.9250 & {\bf 0.9365} & 0.9095 & 0.9180 & {\bf 0.9415} & 0.9290 & 0.9150 & {\bf 0.9485} & 0.9440 \\
\bottomrule
\end{tabular}%
}
\end{table}

% Table3 operator time
\begin{table}[H]
\centering
\small
\caption{Code runtime of coverage probabilities of the Bootstrap, EL, and EEL under Model (\ref{model1}). (mins)}
\label{tab-Time-model1}
\begin{tabular}{@{}c *{9}{c}@{}}
  \toprule
  \multirow{2}{*}{$m$} & \multicolumn{3}{c}{$n=400$} & \multicolumn{3}{c}{$n=2000$} & \multicolumn{3}{c}{$n=4000$} \\
  \cmidrule(lr){2-4} \cmidrule(lr){5-7} \cmidrule(lr){8-10}
  & Bootstrap & EL & EEL & Bootstrap & EL & EEL & Bootstrap & EL & EEL \\
  \midrule
%    1 & 3.3862 & 0.1241 & 0.0026 & 15.3059 & 0.1636 & 0.0043 & 31.1060  & 0.2202 & 0.0058 \\
    2 & 3.5173 & 0.2926 & 0.0028 & 16.5068 & 0.5462 & 0.0047 & 34.7999 & 0.9382 & 0.0110 \\
    5 & 3.7941 & 1.4438 & 0.0042 & 20.4461 & 3.7460 & 0.0125 & 41.0674 & 5.8104 & 0.0189 \\
  10 & 4.0742 & 5.5183 & 0.0073 & 22.6387 & 16.2920 & 0.0177 & 50.3828 & 28.2695 & 0.0390 \\
  \bottomrule
\end{tabular}
\end{table}

Regarding computational speed, Table~\ref{tab-Time-model1} clearly demonstrates that the EEL method is computationally very efficient, whereas the Bootstrap method is comparatively time-consuming.

In summary, our simulation results demonstrate that the proposed EL and EEL methods achieve coverage probabilities closer to the nominal level. 
The EEL method exhibits superior computational efficiency, requiring substantially less runtime than both the EL and bootstrap methods. 
%While the bootstrap method produces slightly shorter confidence intervals, 
Hence our proposed methods provide better theoretical alignment and more robust performance across different model specifications, rendering them particularly well-suited for practical fairness auditing applications.

% Figure EL EEL QQ

\subsection{Flagging}

\begin{figure}[tphb]
\centering
\begin{subfigure}[b]{0.24\textwidth}
    \centering
    \includegraphics[width=\textwidth]{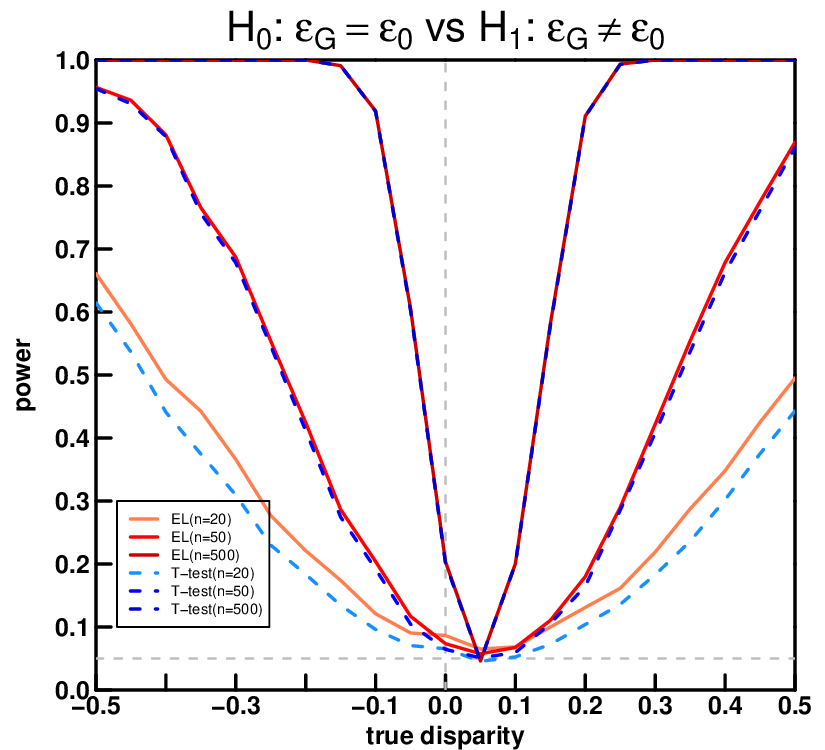}
    \caption{}
    \label{fig-Power-1a}
\end{subfigure}
\hfill
\begin{subfigure}[b]{0.24\textwidth}
    \centering
    \includegraphics[width=\textwidth]{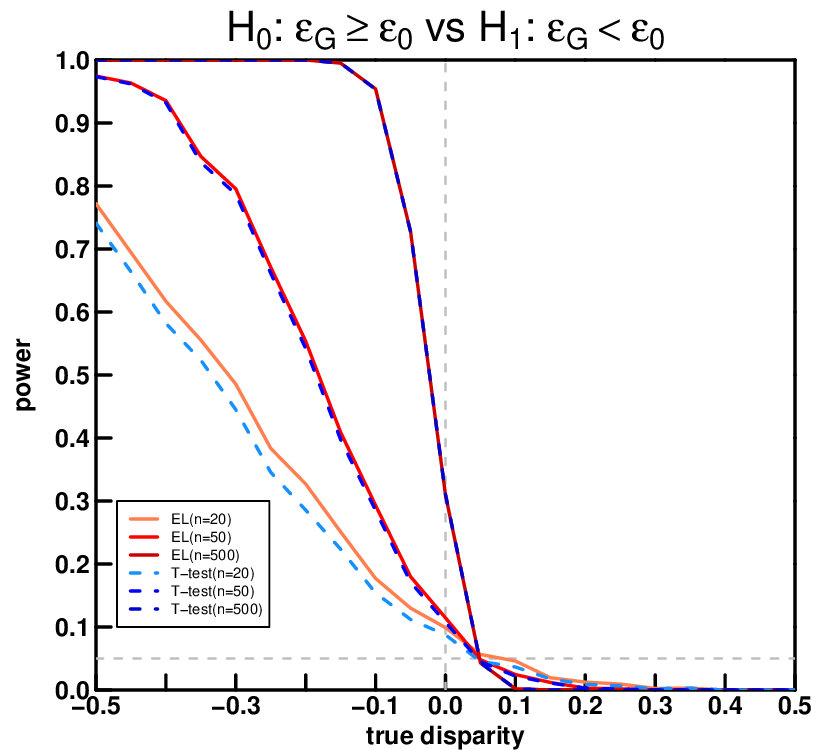}
    \caption{}
    \label{fig-Power-1b}
\end{subfigure}
\hfill
\begin{subfigure}[b]{0.24\textwidth}
    \centering
    \includegraphics[width=\textwidth]{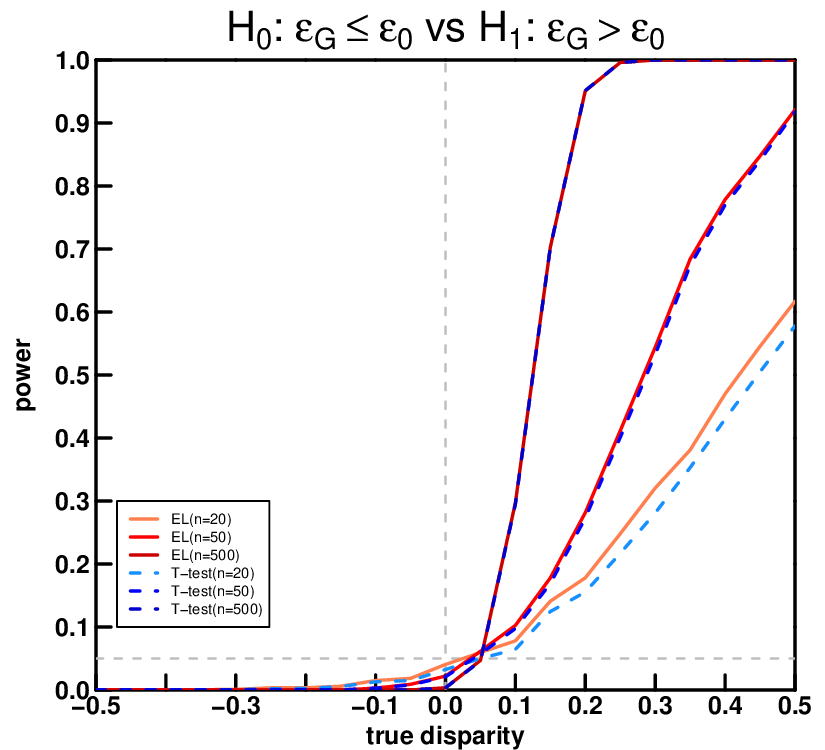}
    \caption{}
    \label{fig-Power-1c}
\end{subfigure}
\hfill
\begin{subfigure}[b]{0.24\textwidth}
    \centering
    \includegraphics[width=\textwidth]{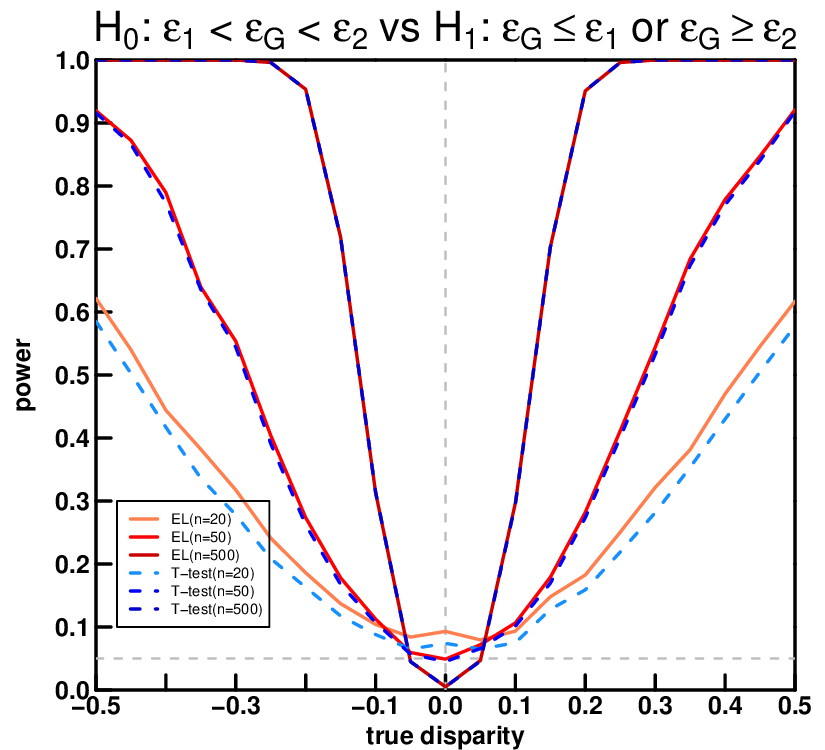}
    \caption{}
    \label{fig-Power-2d}
\end{subfigure}
\caption{Power plots under Model \eqref{model3}: (a) to (d) show results for different sample sizes with $\nu \sim N (0, 1)$. The horizontal line in the figure corresponds to \( \alpha= 0.05 \).}
\label{fig-Power-N}
\end{figure}

\begin{figure}[H]
\centering
\begin{subfigure}[b]{0.24\textwidth}
    \centering
    \includegraphics[width=\textwidth]{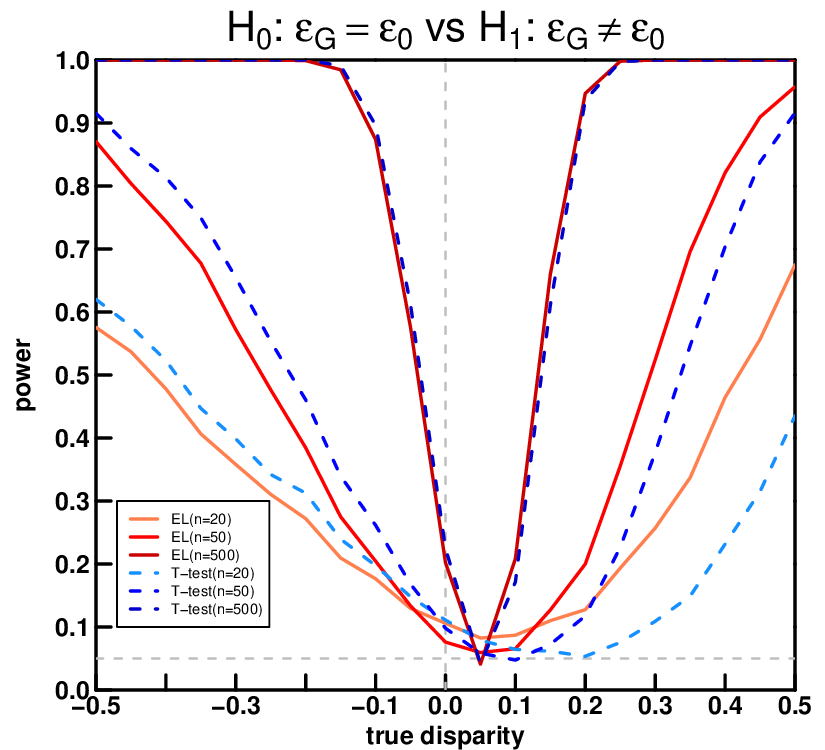}
    \caption{}
    \label{fig-Power-3a}
\end{subfigure}
\hfill
\begin{subfigure}[b]{0.24\textwidth}
    \centering
    \includegraphics[width=\textwidth]{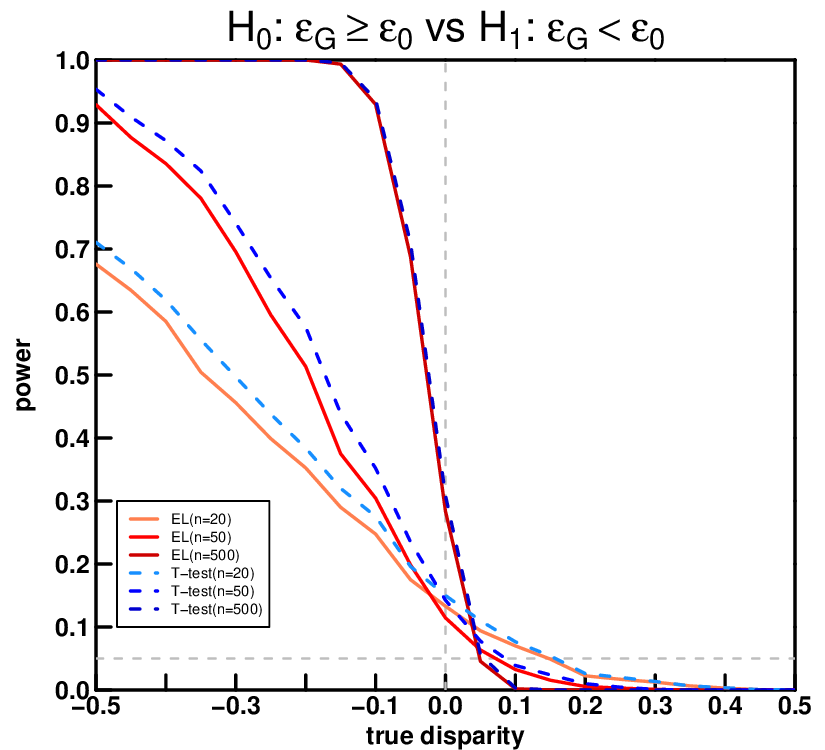}
    \caption{}
    \label{fig-Power-3b}
\end{subfigure}
\hfill
\begin{subfigure}[b]{0.24\textwidth}
    \centering
    \includegraphics[width=\textwidth]{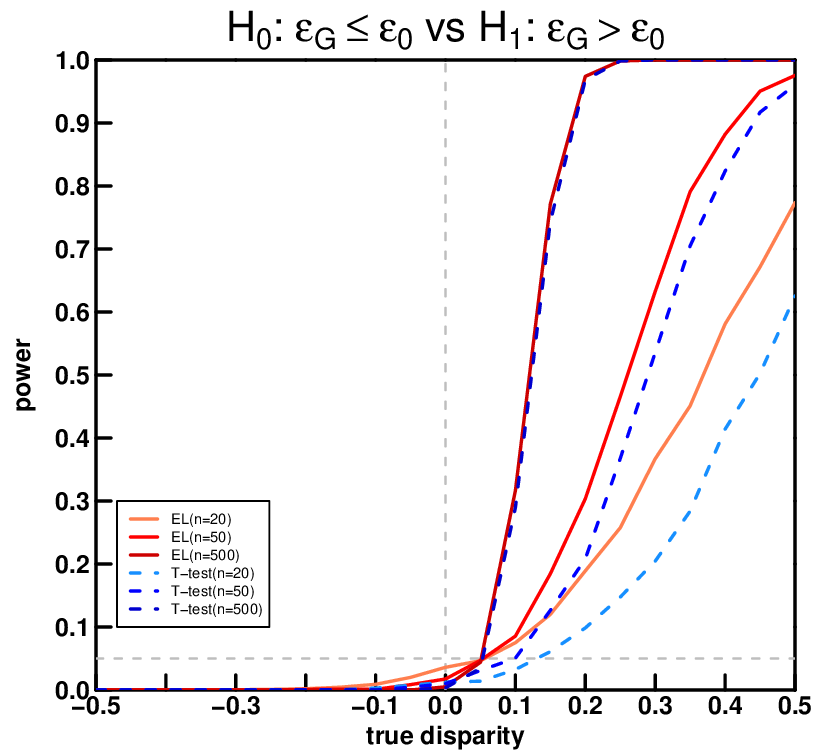}
    \caption{}
    \label{fig-Power-4a}
\end{subfigure}
\hfill
\begin{subfigure}[b]{0.24\textwidth}
    \centering
    \includegraphics[width=\textwidth]{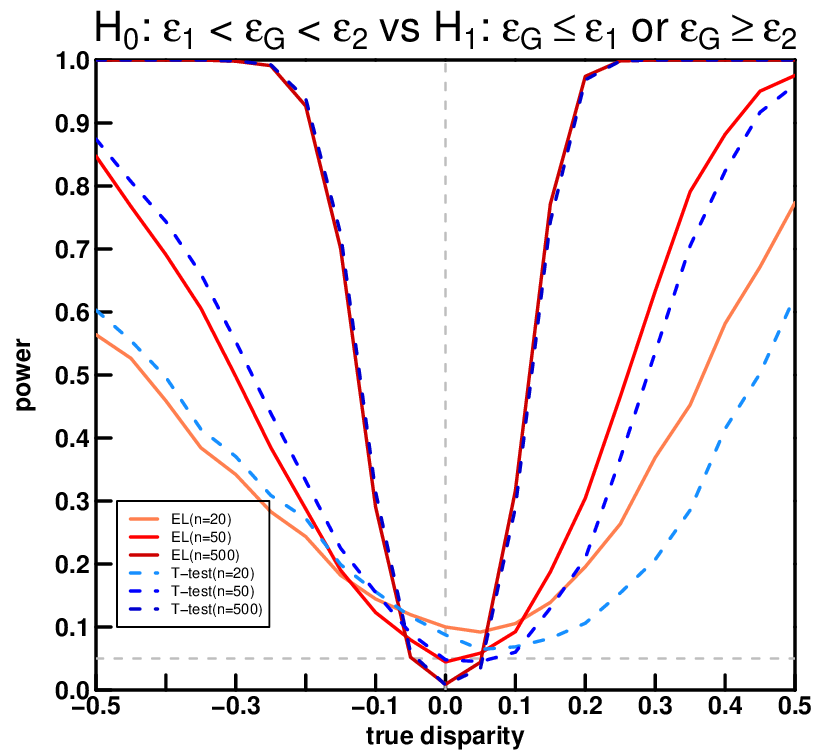}
    \caption{}
    \label{fig-Power-4b}
\end{subfigure}
\caption{Power plots under Model \eqref{model3}: (a) to (d) show results for different sample sizes with $\nu \sim Exp (1) - 1 $. The horizontal line in the figure corresponds to \( \alpha= 0.05 \).}
\label{fig-Power-E}
\end{figure}

Next, we evaluate the validity of the asymptotic predictions derived from Theorems~\ref{theo6} to~\ref{theoFFR} in finite samples. This assessment consists of two parts: the first examines the performance of the EL test under different hypotheses applied to the same dataset, while the second investigates the FFR and statistical power across \(m\) groups.  
All simulation studies were conducted with 2,000 replications. The corresponding results are presented in Figures~\ref{fig-Power-N}--\ref{fig-Power-E} and Table~\ref{tab-FFR}.

To allow $\epsilon_G$ to take negative values and to more directly illustrate the power of the EL test, we consider the following model. We generate $n$ independent samples $(X_i, Y_i)$ from
\bea \label{model3}
X_i  \toiid \mathrm{Unif}(0, 1), \quad Y_i  = \beta_0X_i + \nu_i .  
\eea
We set $\beta_0 = 2$, $\beta_1 = \beta_0 - 2\tau$, $\theta_P = 0$, $f(X) = \beta_1 X$ and $M(f (X), Y ) = Y - f(X)$.
We adopt a partitioned interval structure where each subset $G_j$
is defined with endpoints in the interval $[ {j-1 \over m}, {j \over m})$ for $j = 1, \dots, m$.

To validate Theorems \ref{theo6} and Corollary \ref{coro2}, we examine the power of the EL test under different hypotheses. 
Owing to the high computational cost of the bootstrap, we compare the EL test with the $T$-test, which is computationally cheaper.
We show the performance of EL powers for the four tests with $m=1$. For both $\nu_i \toiid N(0, 1)$ and $\nu_i \toiid \text{Exp}(1) - 1$, a direct computation reveals the true disparity is $\epsilon^*_{G_1} = \tau$ when $m=1$. 
The $T$-test is implemented using the \texttt{t.test} function in R. 
Figures~\ref{fig-Power-N}-\ref{fig-Power-E} present the power curves of the four tests for the following hypothesis pairs:
(1) \( H_0: \epsilon_G = \epsilon_0 \) versus \( H_1: \epsilon_G \neq \epsilon_0 \) with \( \epsilon_0 = 0.05 \);
(2) \( H_0: \epsilon_G \geq \epsilon_0 \) versus \( H_1: \epsilon_G < \epsilon_0 \) with \( \epsilon_0 = 0.05 \);
(3) \( H_0: \epsilon_G \leq \epsilon_0 \) versus \( H_1: \epsilon_G > \epsilon_0 \) with \( \epsilon_0 = 0.05 \);
(4) \( H_0: \epsilon_1 < \epsilon_G < \epsilon_2 \) versus \( H_1: \epsilon_G \leq \epsilon_1 \) or \( \epsilon_G \geq \epsilon_2 \) with \( \epsilon_1 = -0.05 \) and \( \epsilon_2 = 0.05 \).

% Fig3

% Fig4

The following observations can be drawn from Figures~\ref{fig-Power-N}-\ref{fig-Power-E}.
First, the EL test exhibits robust performance across different error distributions, maintaining consistent power characteristics for both symmetric (e.g., normal) and asymmetric (e.g., exponential) measurement errors. 
Under asymmetric samples, the $T$-test is consistent with the EL test for large sample sizes, but exhibits significant deviation and lacks robustness under moderate and small sample sizes.
Second, under the null hypothesis, the test exhibits proper size control: at the boundary points \( \epsilon_0, \epsilon_1, \epsilon_2 \), which represent the least favorable cases, the rejection probabilities remain close to the nominal significance level of 0.05 and decrease rapidly toward zero as \( \epsilon_G^* \) moves into the interior of the null parameter space.
Third, under the alternative hypothesis, the test demonstrates good power properties: the rejection probabilities increase monotonically toward one, with the convergence rate accelerating as the sample size increases.
In summary, both methods exhibit consistent performance under large sample sizes. For interval tests, the EL method achieves power convergence to one faster than the $T$-test. Moreover, under skewed data distributions, the EL method demonstrates superior robustness compared to the $T$-test.

% Table FFR
\begin{table}[H]
\centering
\renewcommand{\arraystretch}{0.7}
\caption{FFR control and statistical power under Model (\ref{model3})}
\label{tab-FFR}
\begin{tabular}{ccccc|ccccc}
\toprule
$\tau$ & $\epsilon^*_{G_1}$ & $\epsilon^*_{G_2}$ & FFR & Power & $\tau$ & $\epsilon^*_{G_1}$ & $\epsilon^*_{G_2}$ & FFR & Power \\
\midrule
-0.15 & -0.075 & -0.225 & 0.0000 & 0.0000 & 0.25 & 0.125 & 0.375 & 0.0000 & 0.7555 \\
-0.10 & -0.050 & -0.150 & 0.0000 & 0.0000 & 0.30 & 0.150 & 0.450 & 0.0000 & 0.8560 \\
-0.05 & -0.025 & -0.075 & 0.0000 & 0.0000 & 0.35 & 0.175 & 0.525 & 0.0000 & 0.9340 \\
0.00 & 0.000 & 0.000 & 0.0015 & 0.0000 & 0.40 & 0.200 & 0.600 & 0.0000 & 0.9783 \\
0.05 & 0.025 & 0.075 & 0.0043 & 0.0735 & 0.45 & 0.225 & 0.675 & 0.0000 & 0.9940 \\
0.10 & 0.050 & 0.150 & 0.0163 & 0.4805 & 0.50 & 0.250 & 0.750 & 0.0000 & 0.9985\\
0.15 & 0.075 & 0.225 & 0.0000 & 0.5558 & 0.55 & 0.275 & 0.825 & 0.0000 & 1.0000 \\
0.20 & 0.100 & 0.300 & 0.0000 & 0.6475 & 0.60 & 0.300 & 0.900 & 0.0000 & 0.9998 \\
\bottomrule
\end{tabular}
\end{table}

Finally, to empirically validate Theorems~\ref{theoFFR}, we investigate the FFR and statistical power across \(m\) groups. Here, power refers to the proportion of false hypotheses that are correctly rejected. We assess their performance in a practical scenario involving the test of \(H_0(G_j): \epsilon_{G_j} \leq \epsilon_0\) versus \(H_1: \epsilon_{G_j} > \epsilon_0\), with \(\epsilon_0 = 0.05\) and \(j = 1, 2, \dots, m\). Assuming \(\nu_i \stackrel{\text{i.i.d.}}{\sim} N(0, 1)\), it can be directly computed that for \(m = 2\), the true disparities are \(\epsilon^*_{G_1} = 0.5\tau\) and \(\epsilon^*_{G_2} = 1.5\tau\). Using Algorithm~\ref{algo-p24} to compute the \(p\)-values for each group and applying the ELBH procedure, we summarize the experimental results in Table~\ref{tab-FFR}.

As shown in Table~\ref{tab-FFR}, the FFR are all below 0.05, indicating that the false flagging rate is well controlled. When $\tau$ is less than 0.05, both $\epsilon^*_{G_j}, j=1,2$ are no greater than 0.05, consistent with the null hypothesis. In this case, the power is zero, meaning that none of the null hypotheses are rejected, which is reasonable. When $\tau$ takes values of 0.05 and 0.1, $\epsilon^*_{G_1} \leq 0.05$ while $\epsilon^*_{G_2} > 0.05$, indicating that the null hypothesis holds for the first group but not for the second. Here, the power approaches 0.5, implying that about half of the hypotheses are rejected. For $\tau$ greater than 0.1, both $\epsilon^*_{G_j}, j=1,2$ exceed 0.05, violating the null hypothesis, and the power rapidly converges to 1, meaning all null hypotheses are rejected, as expected. For a more intuitive understanding of the trends in FFR and Power, Figure~\ref {fig-FFR} is presented. In summary, ELBH is sensitive to the true signal locations and provides effective FFR control while maintaining good power.

\section{Real Data Analysis}
\label{sec-real-data}

In this section, we present an empirical evaluation of the proposed ELFA framework using a real-world dataset.  
Our analysis is based on the well-known COMPAS recidivism prediction instrument, which assigns each defendant a risk score ranging from 1 to 10 that represents the estimated likelihood of reoffending. 
The dataset, publicly released by ProPublica at \url{https://raw.githubusercontent.com/propublica/compas-analysis/master/compas-scores-two-years.csv}, has been widely used in the literature to investigate algorithmic bias within criminal justice systems. 
Prior studies have documented systematic disparities, showing that African-American defendants are more likely to receive higher risk scores than their Caucasian counterparts \citep{AngwinEtAl2016, CherianCandes2024}. 
Building upon this benchmark, we apply our ELFA procedure to quantify such disparities and to construct EL-based bounds that provide statistically rigorous guarantees for fairness assessment.

To examine fairness under the \textit{predictive parity} criterion, we focus on the requirement that the \textit{positive predictive value} (PPV) of a classifier be consistent across demographic groups. Formally, predictive parity for a group $G$ holds when
\[
\mathbb{P}(Y = 1 \mid f(X) = 1, X \in G) = \mathbb{P}(Y = 1 \mid f(X) = 1, X \in G^c),
\]
meaning that a positive prediction is equally informative across groups.  

For our empirical study, we use the COMPAS dataset and binarize the risk scores by defining $f(X) = 1$ if the recidivism prediction instrument score is greater than or equal to 5, and $f(X) = 0$ otherwise. We restrict our analysis to the subset of the holdout data with $f(X) = 1$ and where the recorded race attribute $X_{\text{race}}$ corresponds to either African-American or Caucasian defendants, resulting in $n = 2174$ observations.  
In our empirical likelihood audit, the loss variable $M$ indicates whether the predicted label agrees with the true outcome $Y$. The protected group $G$ consists of African-American defendants, while the reference quantity $\theta_P$ denotes the PPV for the non-protected group (White defendants):
\begin{gather*}
M(f (X), Y ) = Y, \quad G =  \{ \{(X, Y) \mid X_{\text{race}} = \text{African-American},\ f(X) = 1\}  \}, \\
\theta_P = \mathbb{E}[Y = 1 \mid X_{\text{race}} = \text{Caucasian}].
\end{gather*}
The parameter $\theta_P$ is estimated by its empirical conditional expectation, denoted as
\[
\hat{\theta} = \hat{\mathbb{E}}_n [Y = 1 \mid X_{\text{race}} = \text{Caucasian}].
\]
This setup allows us to quantify and test for disparities in predictive parity using our framework.

\begin{figure}[H]
  \centering
  \begin{subfigure}[b]{0.45\textwidth}
    \includegraphics[width=0.95\textwidth]{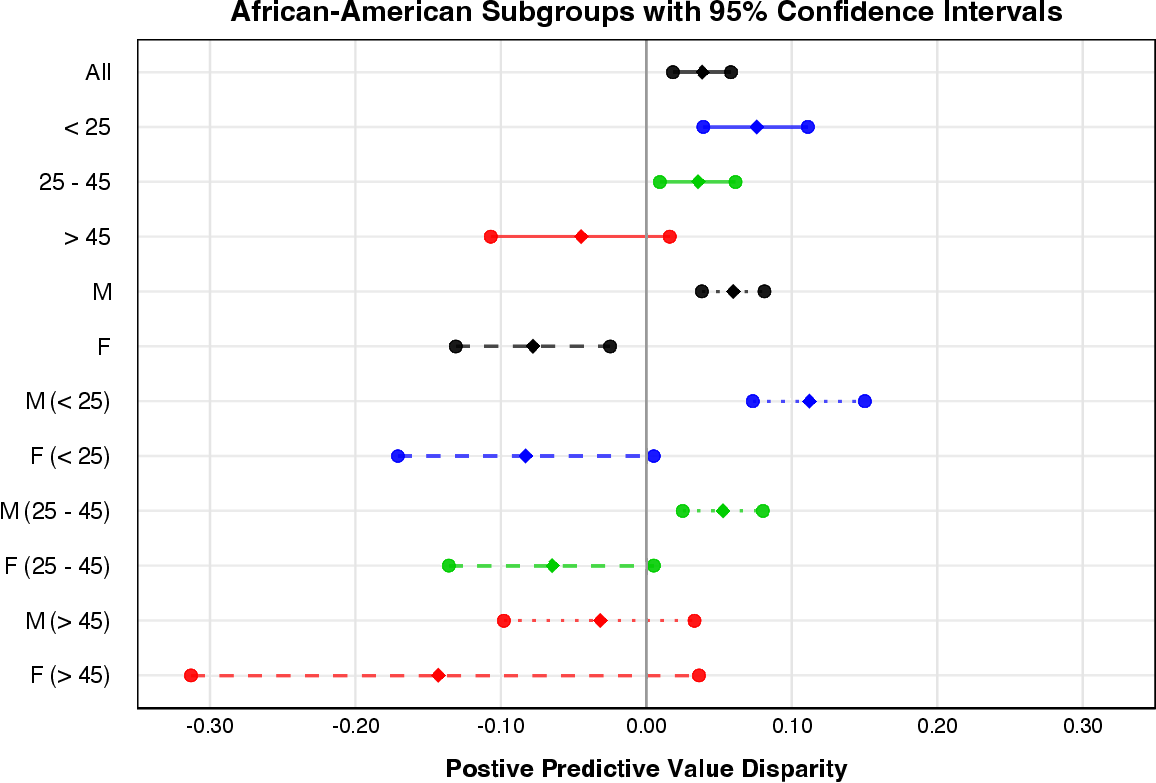}
    \caption{}
    \label{fig-PPV_African-American_Disparity-0.95}
  \end{subfigure}
  \hfill  
  \begin{subfigure}[b]{0.45\textwidth}
    \includegraphics[width=0.95\textwidth]{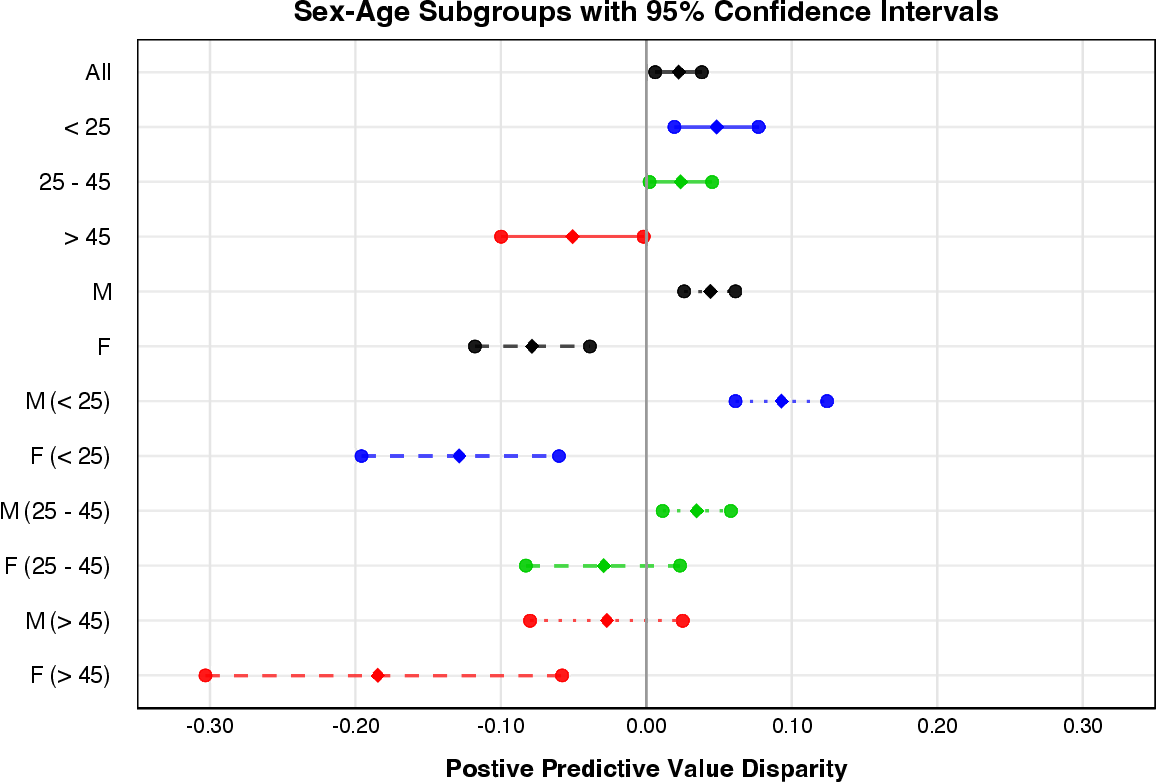}
    \caption{}
    \label{fig-PPV_Sex-Age_Disparity-0.95}
\end{subfigure}
  \caption{(a) plots confidence intervals for the difference in COMPAS PPV between each African-American subgroup and the entire Caucasian subgroup. (b) plots confidence intervals for the difference in COMPAS PPV between each subgroup formed by intersections of age and sex and the entire Caucasian subgroup.}
  \label{fig-Sex-Age Disparity}
\end{figure}

Firstly, to quantify the magnitude of this disparity, we construct empirical likelihood-based confidence intervals for
\[
\epsilon(G) := \mathbb{P}(Y = 1 \mid f(X) = 1, X \in G) - \mathbb{P}(Y = 1 \mid f(X) = 1, X \in G^c),
\]
which represents the difference in PPV between the two groups. We then conduct the hypothesis test
\[
H_0(G): \epsilon(G) = 0 \quad \text{versus} \quad H_1(G): \epsilon(G) \neq 0,
\]
where the null hypothesis corresponds to perfect predictive parity. Applying our ELFA procedure, we obtain 90\% and 95\% empirical likelihood confidence intervals for $\epsilon(G)$ of $[0.022,\, 0.055]$ and $[0.018,\, 0.058]$, respectively. Both intervals exclude zero, providing statistically significant evidence of a predictive disparity between African-American and Caucasian defendants under the COMPAS risk assessment. These results demonstrate that the proposed ELFA method effectively captures and quantifies fairness discrepancies with rigorous statistical guarantees.

Secondly, again using the PPV for Caucasian defendants as the reference, we construct simultaneous   95\% confidence intervals for the PPV disparity across all subpopulations defined by the intersection of African-American, sex, and age. 
Figure~\ref{fig-PPV_African-American_Disparity-0.95} indicates that, at the 95\% level, the intervals exclude zero for \textit{All}, $\!<25$, $25$--$45$, \textit{M}, \textit{M}($<25$), and \textit{M}($25$--$45$), implying higher PPV for these African-American groups relative to the Caucasian reference; the interval for \textit{F} also excludes zero but in the negative direction. 
%At the 90\% level, two additional female-by-age groups, \textit{F}($<25$) and \textit{F}($25$--$45$), show negative disparities with intervals excluding zero. 
Overall, ELFA suggests that COMPAS predictions are parity-consistent for several, though not all, African-American subpopulations, with departures from parity concentrated among younger male groups that exhibit higher PPV than the Caucasian reference.

\begin{figure}[H]
  \centering
  \begin{subfigure}[b]{0.49\textwidth}
    \includegraphics[width=0.95\textwidth]{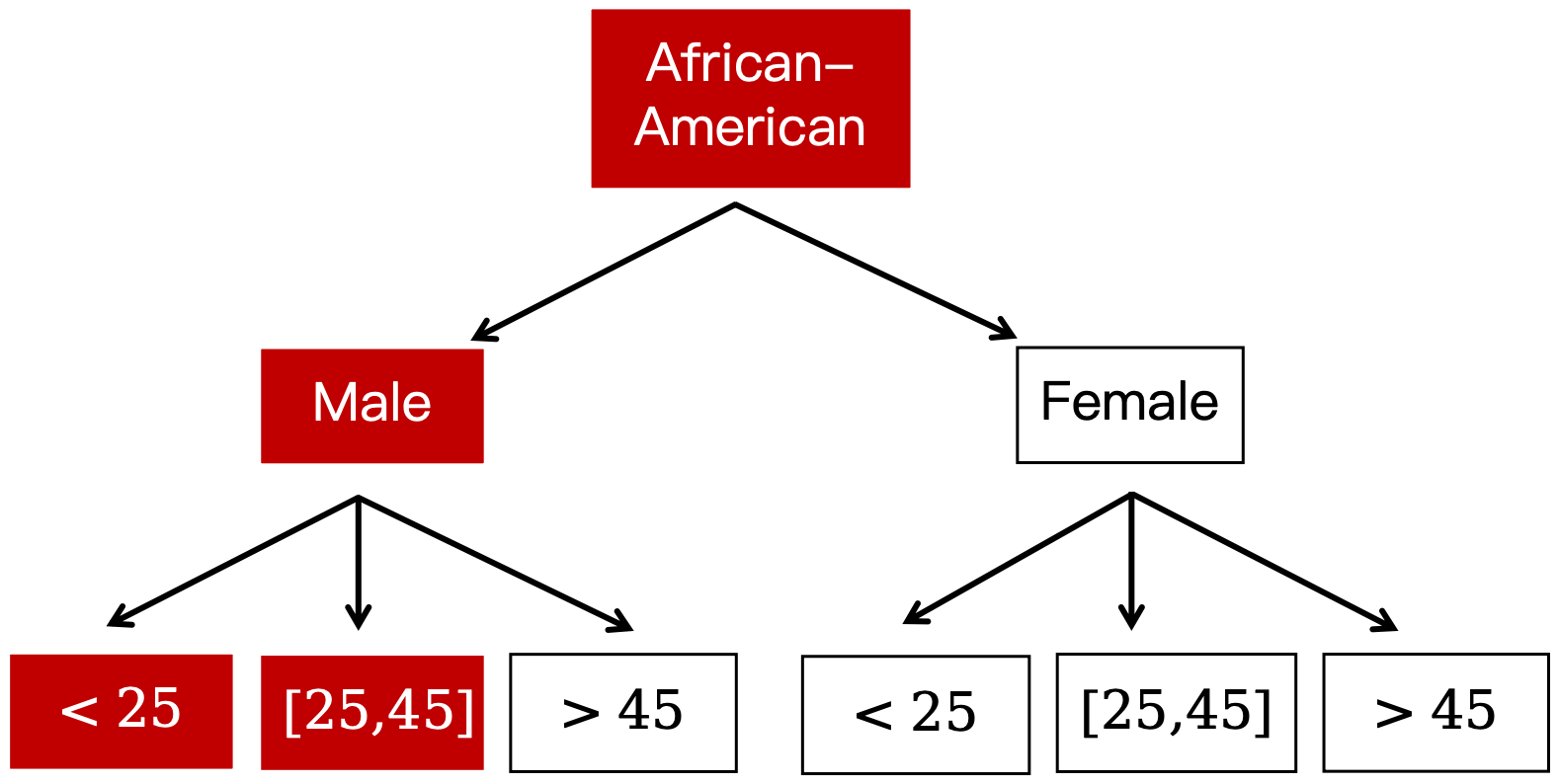}
    \caption{}
    \label{fig-flag-1a}
  \end{subfigure}
  \hfill  
  \begin{subfigure}[b]{0.49\textwidth}
    \includegraphics[width=0.95\textwidth]{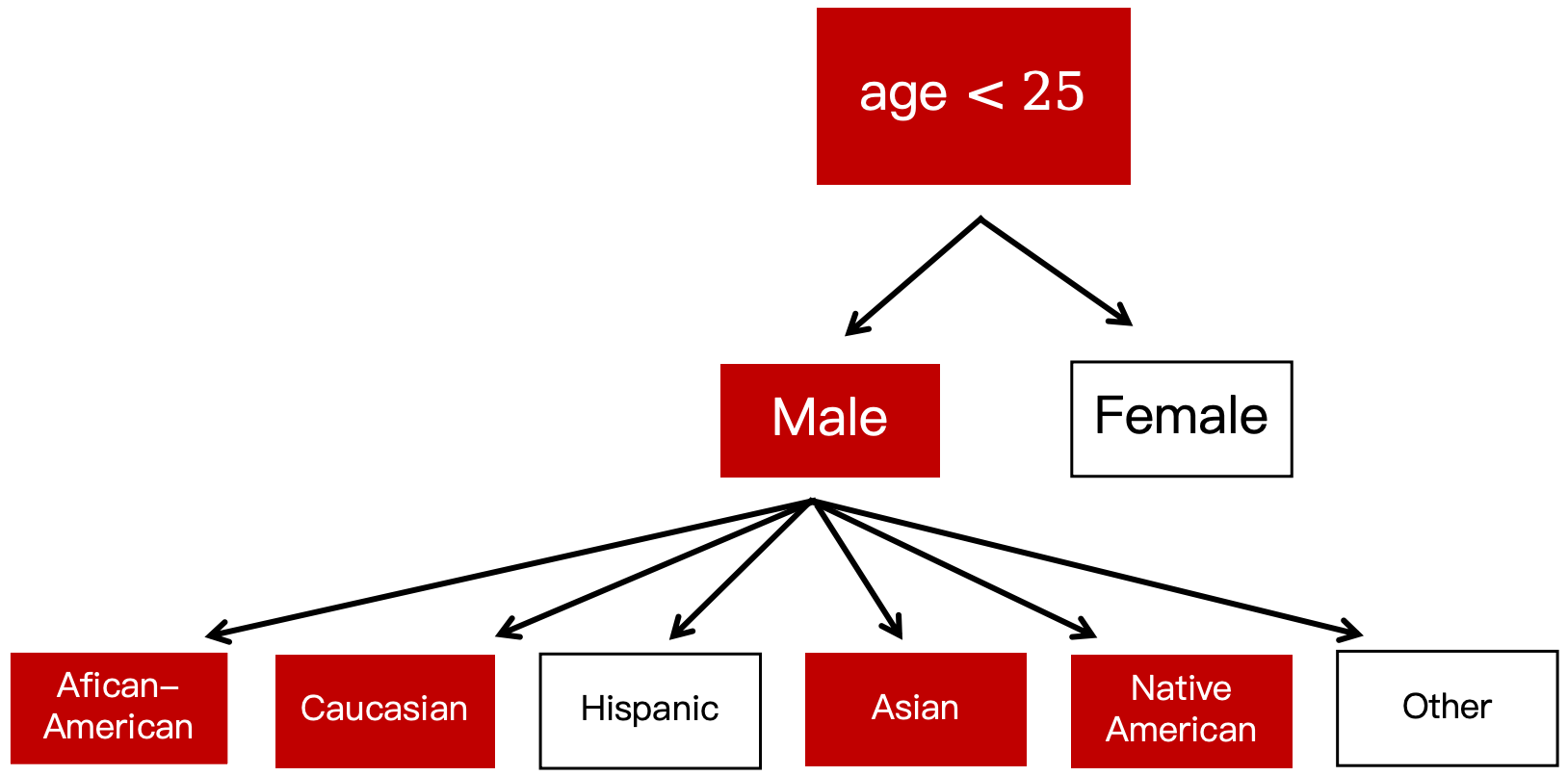}
    \caption{}
    \label{fig-flag-1b}
  \end{subfigure}
  \caption{The figure shows the flagging subpopulations of the COMPAS dataset, which possess disparity over the PPV for Caucasian defendants. The red boxes correspond to groups flagged as having substantially higher-than-target false positive rates.}
  \label{fig-flag-1}
\end{figure}

Thirdly, taking the PPV for Caucasian defendants as the reference, we construct simultaneous 95\% confidence intervals for the PPV disparity across all subpopulations defined by the intersection of sex and age alone, based on $n=3317$ observations. In summary, conditional on a positive COMPAS prediction, the PPV for Caucasian defendants is lower than the overall PPV pooled across all racial categories. As shown in Figure~\ref{fig-PPV_Sex-Age_Disparity-0.95}, this below-average pattern is most pronounced in key subgroups such as for \textit{All}, $\!<25$, $25$--$45$, \textit{M}, \textit{M}($<25$), and \textit{M}($25$--$45$), with the Caucasian PPV consistently falling below the all-group benchmark.

\begin{figure}[H]
  \centering
  \begin{subfigure}[b]{0.39\textwidth}
    \includegraphics[width=0.95\textwidth]{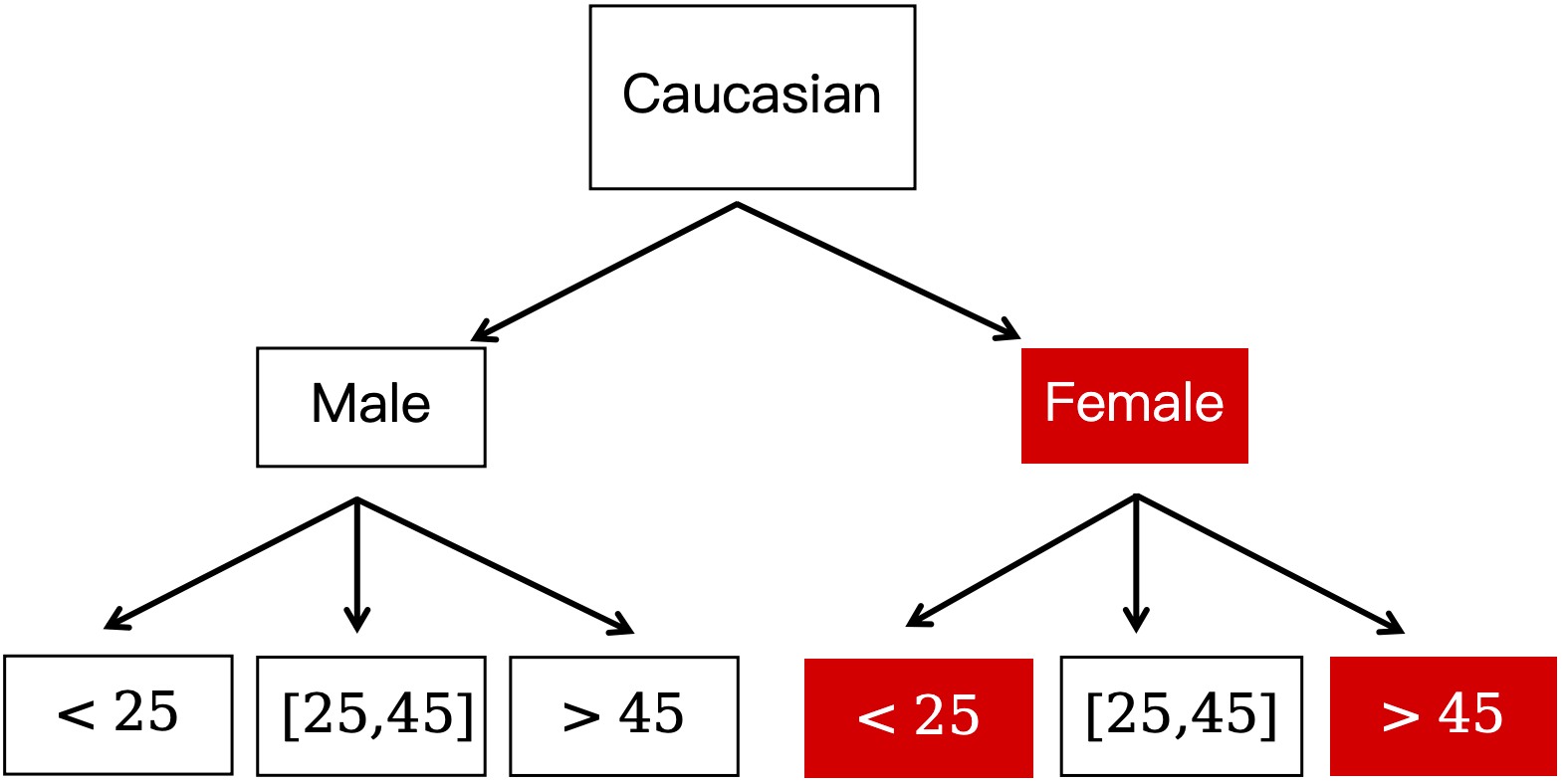}
    \caption{}
    \label{fig-flag-2a}
  \end{subfigure}
  \hfill  
  \begin{subfigure}[b]{0.59\textwidth}
    \includegraphics[width=0.95\textwidth]{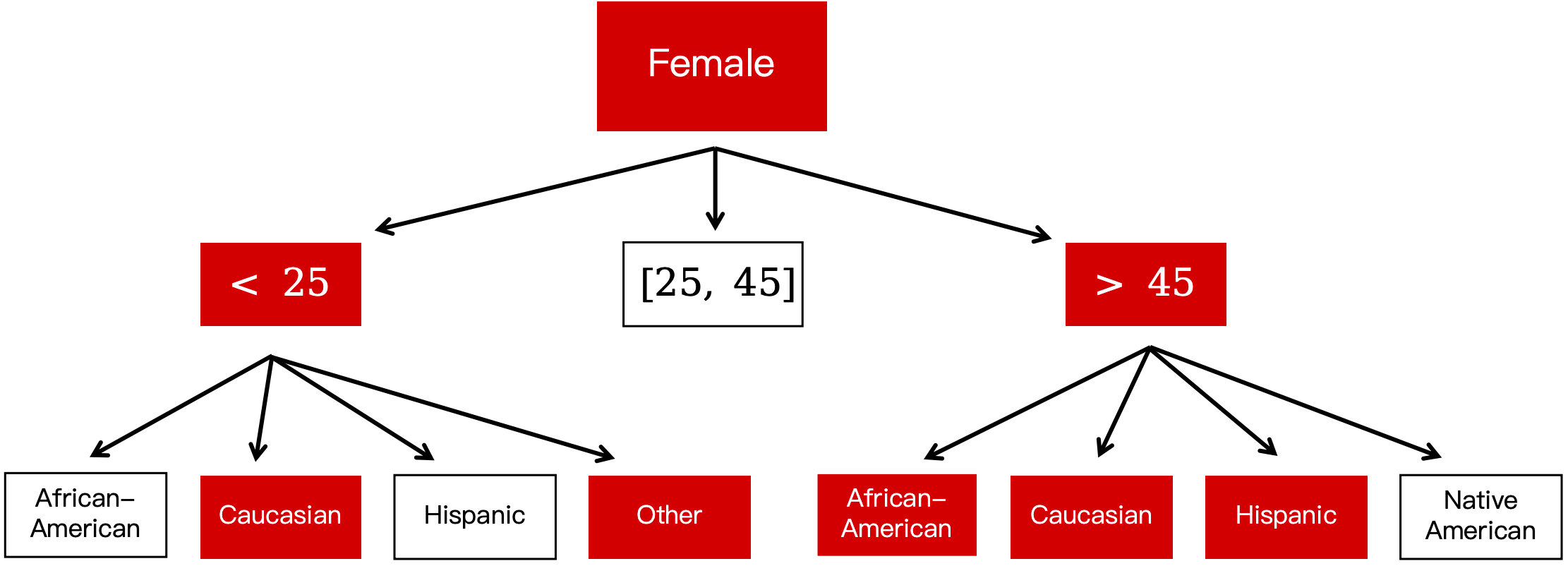}
    \caption{}
    \label{fig-flag-2b}
  \end{subfigure}
  \caption{The figure shows the flagging subpopulations of the COMPAS dataset, which possess disparity over the PPV for the average. The red boxes correspond to groups flagged as having substantially lower-than-target false positive rates.}
  \label{fig-flag-2}
\end{figure}

Finally, we use the ELBH procedure to investigate whether any demographic groups exhibit harmful PPV disparities. Let $\epsilon_0 = 0.01$ denote the disparity threshold. We conduct the hypothesis test:
\[
H_0(G_j): \epsilon(G) \le \epsilon_0 \quad \text{vs} \quad H_1(G_j): \epsilon(G) > \epsilon_0, \quad j = 1, \dots, |\mathcal{G}|
\]
using the PPV for Caucasian defendants as the reference, i.e., $\hat{\theta} = \hat{\mathbb{E}}_n [Y = 1 \mid X_{\text{race}} = \text{Caucasian}]$. A rejection of \(H_0(G_j)\) indicates that group \(G_j\) has a PPV at least 0.01 higher than the Caucasian benchmark. As shown in Figure~\ref{fig-flag-1}, the subpopulation of African-American males under age 25 is flagged. We also conduct the hypothesis test:
\[
H_0(G_j): \epsilon(G) \ge -\epsilon_0 \quad \text{vs} \quad H_1(G_j): \epsilon(G) < -\epsilon_0, \quad j = 1, \dots, |\mathcal{G}|
\]
using the overall average PPV as the reference, i.e., $\hat{\theta} = \hat{\mathbb{E}}_n [Y = 1]$. This is equivalent to flagging groups whose PPV is at least 0.01 lower than the overall average. As shown in Figure~\ref{fig-flag-2}, the subpopulations of females under age 25 and Caucasians over age 45 are flagged under this deficit criterion.

In summary, the ELFA identifies two significant disparities in the COMPAS algorithm: a higher positive prediction rate for African-American males (notably those under 25) compared to Caucasians, and a lower rate for Caucasian females than the average.

\section{Concluding remarks}
\label{sec-concluding}

This paper introduced ELFA, a novel empirical likelihood-based framework designed to address the dual challenges of certifying fairness compliance and flagging unfair treatment in black-box algorithmic systems. By leveraging asymptotic distributions and the BH procedure, ELFA enables third-party auditors to assess models without requiring access to internal architectures or making restrictive distributional assumptions.

Our framework offers two certification paths (EL and EEL) and four flexible flagging structures tailored to diverse auditing scenarios. Simulation results confirm that ELFA outperforms bootstrap alternatives in both statistical coverage and computational speed. Furthermore, our application to the COMPAS dataset demonstrates the framework’s practical utility in uncovering critical intersectional disparities, such as those affecting young African-American males and Caucasian females. Ultimately, ELFA provides a rigorous yet efficient foundation for enhancing transparency and accountability in algorithmic decision-making.

%%\vskip1cm
\if1\anon
\bigskip
\noindent {\bf Acknowledgments}
Chuanlong Xie was supported by the National Natural Science Foundation of China (No.12201048). Lixing Zhu was supported by the grants  (NSFC12131006, NSFC12471276) from the National Natural Scientific Foundation of China and the grant (CI2023C063YLL) from the Scientific and Technological Innovation Project of China Academy of Chinese Medical Science.
\else
%[Funding information omitted for anonymous review.]
\fi

\bigskip
\noindent {\bf Supplemental Materials}
The Supplemental Materials include proofs, further simulation results, algorithms, and discussion.
\if1\anon
R code is available at \url{https://github.com/Tang-Jay/EL-for-fairness-auditing}.
\else
R code is available upon request.
\fi

% \bigskip
% \noindent {\bf Disclosure statement}
% No potential conflict of interest was reported by the author(s).

\bibliography{reference.bib}

@article{Luo1994,
title={Large Sample Properties of the Empirical Euclidean Likelihood Estimation for Semiparametric Model},
author={Luo, Xu},
journal={Chinese Journal of Applied Probability and Statistics},
volume={10},
number={4},
pages={344--352},
year={1994},
publisher={Editorial Office of Chinese Journal of Applied Probability and Statistics}
}

@article{BenjaminiHochberg1995,
title={Controlling the false discovery rate: a practical and powerful approach to multiple testing},
author={Benjamini, Yoav and Hochberg, Yosef},
journal={Journal of the Royal statistical society: series B (Methodological)},
volume={57},
number={1},
pages={289--300},
year={1995},
publisher={Wiley Online Library}
}

@article{QinLawless1994,
title={Empirical likelihood and general estimating equations},
author={Qin, Jin and Lawless, Jerry},
journal={the Annals of Statistics},
volume={22},
number={1},
pages={300--325},
year={1994},
publisher={Institute of Mathematical Statistics}
}

@book{LehmannRomano2005,
title={Testing statistical hypotheses},
author={Lehmann, Erich Leo and Romano, Joseph P},
year={2005},
publisher={Springer}
}

@misc{MaityEtAl2021,
title={Statistical inference for individual fairness}, 
author={Subha Maity and Songkai Xue and Mikhail Yurochkin and Yuekai Sun},
year={2021},
eprint={2103.16714},
archivePrefix={arXiv},
primaryClass={stat.ML},
url={https://arxiv.org/abs/2103.16714}, 
}

@article{Owen1991,
author= {Owen, A. B.},
title = {Empirical likelihood for linear models},
journal = {Annals of Statistics},
year= {1991},
volume= {19},
pages = {1725--1747}
}

@article{Chen1996,
author= {Chen, S. X.},
title = {Empirical likelihood confidence intervals for nonparametric density estimation},
journal = {Biometrika},
year= {1996},
volume= {83},
pages = {329--341}
}

@article{CherianCandes2024,
author= {John J. Cherian and Emmanuel J. Cand{{\`e}}s},
title = {Statistical Inference for Fairness Auditing},
journal = {Journal of Machine Learning Research},
year= {2024},
volume= {25},
number= {149},
pages = {1--49},
url = {http://jmlr.org/papers/v25/23-0739.html}
}

@incollection{AngwinEtAl2016,
author= {Angwin, J. and Larson, J. and Mattu, S. and Kirchner, L.},
title = {Machine Bias},
booktitle = {Ethics of Data and Analytics},
pages = {254--264},
publisher = {Auerbach Publications},
year= {2016}
}

@misc{BrundageEtAl2020,
 author = {Brundage, M. and Avin, S. and Wang, J. and Belfield, H. and Krueger, G. and Hadfield, G. and Khlaaf, H. and Yang, J. and Toner, H. and Fong, R. and others},
 title= {Toward Trustworthy AI Development: Mechanisms for Supporting Verifiable Claims},
howpublished = {arXiv preprint arXiv:2004.07213},
 year = {2020}
}

@incollection{Dastin2018,
author= {Dastin, J.},
title = {Amazon Scraps Secret AI Recruiting Tool That Showed Bias Against Women},
booktitle = {Ethics of Data and Analytics},
pages = {296--299},
publisher = {Auerbach Publications},
year= {2018}
}

@inproceedings{DiciccioEtAl2020,
author= {DiCiccio, C. and Vasudevan, S. and Basu, K. and Kenthapadi, K. and Agarwal, D.},
title = {Evaluating Fairness Using Permutation Tests},
booktitle = {Proceedings of the 26th ACM SIGKDD International Conference on Knowledge Discovery \& Data Mining},
pages = {1467--1477},
year= {2020},
address = {New York, NY, USA},
publisher = {Association for Computing Machinery},
isbn= {9781450379984}
}

@inproceedings{DworkEtAl2012,
author= {Dwork, C. and Hardt, M. and Pitassi, T. and Reingold, O. and Zemel, R.},
title = {Fairness Through Awareness},
booktitle = {Proceedings of the 3rd Innovations in Theoretical Computer Science Conference},
pages = {214--226},
year= {2012},
address = {New York, NY, USA},
publisher = {Association for Computing Machinery},
isbn= {9781450311151},
doi = {10.1145/2090236.2090255}
}

@misc{MorinaEtAl2019,
author = {Morina, G. and Oliinyk, V. and Waton, J. and Marusic, I. and Georgatzis, K.},
title= {Auditing and Achieving Intersectional Fairness in Classification Problems},
howpublished = {arXiv preprint arXiv:1911.01468},
year = {2019}
}

@misc{RoyMohapatra2023,
author = {Roy, A. and Mohapatra, P.},
title= {Fairness Uncertainty Quantification: How Certain Are You That the Model is Fair?},
howpublished = {arXiv preprint arXiv:2110.01052},
year = {2023}
}

@misc{SchaakeClark2022,
author = {Schaake, M. and Clark, J.},
title= {Stanford Launches AI Audit Challenge},
year = {2022},
month= jul,
note = {URL \url{https://hai.stanford.edu/news/stanford-launches-ai-audit-challenge}}
}

@inproceedings{SiEtAl2021,
author= {Si, N. and Murthy, K. and Blanchet, J. and Nguyen, V. A.},
title = {Testing Group Fairness via Optimal Transport Projections},
booktitle = {International Conference on Machine Learning},
pages = {9649--9659},
year= {2021},
publisher = {PMLR}
}

@inproceedings{TaskesenEtAl2021,
author= {Taskesen, B. and Blanchet, J. and Kuhn, D. and Nguyen, V. A.},
title = {A Statistical Test for Probabilistic Fairness},
booktitle = {Proceedings of the 2021 ACM Conference on Fairness, Accountability, and Transparency},
pages = {648--665},
year= {2021}
}

@misc{vonZahnEtAl2023,
author = {von Zahn, M. and Hinz, O. and Feuerriegel, S.},
title= {Locating Disparities in Machine Learning},
howpublished = {arXiv preprint arXiv:2208.06680},
year = {2023}
}

@inproceedings{XueEtAl2020,
author= {Xue, S. and Yurochkin, M. and Sun, Y.},
title = {Auditing {ML} Models for Individual Bias and Unfairness},
booktitle = {International Conference on Artificial Intelligence and Statistics},
pages = {4552--4562},
year= {2020},
publisher = {PMLR}
}

@inproceedings{YanZhang2022,
author= {Yan, T. and Zhang, C.},
title = {Active Fairness Auditing},
booktitle = {International Conference on Machine Learning},
pages = {24929--24962},
year= {2022},
publisher = {PMLR}
}

@incollection{Dudley1984,
author= {Dudley, R. M.},
title = {A course on empirical processes},
booktitle = {Ecole d'{\'e}t{\'e} de Probabilit{\'e}s de Saint-Flour XII-1982},
pages = {1--142},
year= {1984},
publisher = {Springer},
series= {Lecture Notes in Mathematics},
volume= {1097}
}

@article{GineZinn1984,
author= {Gin{\'e}, E. and Zinn, J.},
title = {Some limit theorems for empirical processes},
journal = {The Annals of Probability},
year= {1984},
pages = {929--989}
}

@book{Ferguson1996,
author= {Ferguson, T. S.},
title = {A Course in Large Sample Theory},
edition = {1st},
year= {1996},
publisher = {Routledge},
doi = {10.1201/9781315136288},
url = {https://doi.org/10.1201/9781315136288}
}

@book{vanderVaart2000,
author= {A. W. van der Vaart},
title = {Asymptotic Statistics},
series= {Cambridge Series in Statistical and Probabilistic Mathematics},
volume= {3},
publisher = {Cambridge University Press},
year= {2000}
}

@book{vanderVaartWellner1996,
author= {van der Vaart A. W. and WellnerJ. A.},
title = {Weak Convergence and Empirical Processes},
publisher = {Springer},
year= {1996}
}

@inproceedings{LiuZhao2024,
 title={Empirical Likelihood for Fair Classification},
 author={Liu, P. P. and Zhao, Y. C.},
 booktitle={The Twelfth International Conference on Learning Representations},
 year={2024},
 url={https://openreview.net/forum?id=GACjMj1MS1}
}

@article{Owen1988,
author= {Owen, A. B.},
title = {Empirical likelihood ratio confidence intervals for a single functional},
journal = {Biometrika},
year= {1988},
volume= {75},
pages = {237--249}
}

@article{Owen1990,
author= {Owen, A. B.},
title = {Empirical likelihood ratio confidence regions},
journal = {Annals of Statistics},
year= {1990},
volume= {18},
pages = {90--120},
note= {Abbreviated as Ann. Statist. in citation}
}

@inproceedings{TramerEtAl2017,
author= {F. Tram{\`e}r and V. Atlidakis and R. Geambasu and D. Hsu and J. Hubaux and M. Humbert and A. Juels and H. Lin},
title = {FairTest: Discovering Unwarranted Associations in Data-Driven Applications},
booktitle = {IEEE European Symposium on Security and Privacy (EuroS\&P)},
pages = {401--416},
year= {2017}
}

@inproceedings{GargEtAl2020,
author= {Garg, P. and Villasenor, J. and Foggo, V.},
title = {Fairness Metrics: A Comparative Analysis},
booktitle = {IEEE International Conference on Big Data (Big Data)},
pages = {3662--3666},
year= {2020},
publisher = {IEEE}
}

@inproceedings{VermaRubin2018a,
author= {Verma, S. and Rubin, J.},
title = {Fairness Definitions Explained},
booktitle = {Proceedings of the International Workshop on Software Fairness},
pages = {1--7},
year= {2018}
}

@article{MehrabiEtAl2021,
author= {Mehrabi, N. and Morstatter, F. and Saxena, N. and Lerman, K. and Galstyan, A.},
title = {A Survey on Bias and Fairness in Machine Learning},
journal = {ACM Computing Surveys (CSUR)},
volume= {54},
number= {6},
pages = {1--35},
year= {2021}
}

@misc{Simonite2015,
author= {Simonite, T.},
title = {Probing the dark side of Google's ad-targeting system},
journal = {MIT Technology Review},
year= {2015},
url = {https://www.technologyreview.com/s/539021/probing-the-dark-side-of-googles-ad-targeting-system/}
}

@article{PessachShmueli2022,
author= {Pessach, D. and Shmueli, E.},
title = {A review on fairness in machine learning},
journal = {ACM Computing Surveys},
volume= {55},
number= {3},
pages = {1--44},
year= {2022}
}

@article{LengTang2012,
title = {Penalized empirical likelihood and growing dimensional general estimating equations},
author = {Leng, Chenlei and Tang, Cheng Yong},
journal = {Biometrika},
volume = {99},
number = {3},
pages = {703--716},
year = {2012},
publisher = {Oxford University Press}
}

@article{DiCiccioEtAl1991,
title = {Empirical Likelihood is Bartlett-Correctable},
author = {DiCiccio, Thomas J. and Hall, Peter and Romano, Joseph P.},
journal = {Annals of Statistics},
volume = {19},
number = {3},
pages = {1053--1061},
year = {1991},
publisher = {Institute of Mathematical Statistics},
doi = {10.1214/aos/1176348137}
}

@article{LandersBehrend2022,
title = {Auditing the AI auditors: A framework for evaluating fairness and bias in high stakes AI predictive models},
author = {Landers, R. and Behrend, Tara S.},
journal = {The American Psychologist},
year = {2022},
doi = {10.1037/amp0000972}
}

@article{LaineEtAl2024,
title = {Ethics-based AI auditing: A systematic literature review on conceptualizations of ethical principles and knowledge contributions to stakeholders},
author = {Laine, Joakim and Minkkinen, Matti and Mäntymäki, Matti},
journal = {Information \& Management},
year = {2024},
volume = {61},
pages = {103969},
doi = {10.1016/j.im.2024.103969}
}

@article{LacmanovićŠkare2025,
title = {Artificial intelligence bias auditing -- current approaches, challenges and lessons from practice},
author = {Lacmanović, Sabina and Škare, M.},
journal = {Review of Accounting and Finance},
year = {2025},
doi = {10.1108/raf-01-2025-0006}
}

@article{ChenShi2011,
title = {Empirical likelihood hypothesis test on mean with inequality constraints},
author = {Chen, Li and Shi, Jian},
journal = {Science China Mathematics},
volume = {54},
number = {9},
pages = {1847--1857},
year = {2011},
publisher = {Springer}
}

\bigskip

\newpage
\begin{center}
{\Large\bfseries SUPPLEMENTAL MATERIAL}
\end{center}

\appendix
\setcounter{algorithm}{0}
\renewcommand{\thealgorithm}{\thesection.\arabic{algorithm}}
\setcounter{figure}{0}
\renewcommand{\thefigure}{\thesection.\arabic{figure}}
\setcounter{lemm}{0}
\renewcommand{\thelemm}{\thesection.\arabic{lemm}}
\makeatletter
\@addtoreset{lemm}{section}
\makeatother

% ---------------------------------
% % 放附录或者和4.1合并

% If one need to obtain the confidence interval for  subgroup disparity measure $ \epsilon_{G_j} $ only, we  can let  
% \[
% \ell_{EL2}( \epsilon_{G_j}):=2\log\{{R}_{EL}(\hat{\bepsilon})\}-2\log\{{R}_{EL}(\epsilon_{G_j},  \{ \hepsilon_{G_i}\}_{i=1,i\neq j}^m)\},
% \] 
% where $\hat{\bepsilon}$ and $\{ \hepsilon_{G_i}\}_{i=1,i\neq j}^m$ are the EL estimators of $\ \bepsilon$ and $\{ \epsilon_{G_i}\}_{i=1,i\neq j}^m $ 
% (with $ \epsilon_{G_j}$ fixed for the later estimator), respectively. 
% We have the following result. 

% \begin{theo}\label{theoEL2}
% Suppose that conditions (\ref{C1}) to (\ref{C6})  are satisfied. Let $\epsilon_{0_{G_j}}$ be the true disparity performance of subgroup $G_j$. Then as $n\to \infty$,
% \[
% \ell_{EL2}( \epsilon_{0_{G_j}}) \tod \chi^2_{1}.
% \]
% \end{theo}

% The result follows by adapting the proof of Corollary $5$ in \cite{QinLawless1994}. From this result, the EL based confidence region for $\epsilon_{0_{G_j}}$ with asymptotically correct coverage probability $1 - \alpha$ can be constructed as
% \[
% \{ \epsilon_{G_j}: \ell_{EL2}( \epsilon_{G_j}) \leq z_{\alpha}(1) \}.
% \]

% -----------------------

\section{Extended Hypothesis Tests for Flagging} \label{appsec-extended-tests}

We consider the test of \( H_0(G): \epsilon_G \ge \epsilon_0 \) versus \( H_1(G): \epsilon_G < \epsilon_0 \), and the test of \( H_0(G): \epsilon_1 \leq \epsilon_G \leq \epsilon_2 \) versus \( H_1(G): \epsilon_G < \epsilon_1 \) or \( \epsilon_G > \epsilon_2 \). 
The boundary point serves as the least favorable point under the null hypothesis $H_0$, where the limiting distribution of the test statistic follows a mixture chi-square distribution. 
As the true disparity deviates further from the boundary point in the direction of the alternative hypothesis, the test power approaches 1. 

\subsection{One-sided test for over-protected subgroups}  \label{appsec-one-sided}

This section presents the second one-sided test problem: $ H_0(G) : \{ \epsilon_G \ge \epsilon_0 \}$ versus $ H_1(G) : \{ \epsilon_G < \epsilon_0 \}$. According to the arguments in Section~\ref{sec-4.1}, the empirical likelihood ratio test statistic for this hypothesis test is given by  
\[
T(G) :=  -2 \log \frac{\sup_{\epsilon_G \ge \epsilon_0}L_{EL}(\epsilon_G)}{\sup_{\epsilon_G \in \mathbb{R}^1} L_{EL}(\epsilon_G)}.
\]
From Lemma \ref{lemm_ell}, we have
\[
\sup_{\epsilon_G \ge \epsilon_0}  L_{EL}(\epsilon_G) = 
\begin{cases}
L_{EL}(\epsilon_0), & \ \mathrm{if} \  \ \hepsilon_G < \epsilon_0,\\
                      n^{-n}, & \ \mathrm{if} \  \  \hepsilon_G \ge \epsilon_0.
\end{cases}
\]
Hence, it is clear that 
\[ 
T(G)  =  \left( -2 \log \frac{L_{EL}(\epsilon_0)}{L_{EL}(\hepsilon_G)} \right) \mathbbm{1}(\hepsilon_G < \epsilon_0).
\]

We now state the main results. Under the null hypothesis, the limiting distribution of $T(G)$ also depends critically on the location of the true disparity $\epsilon_G^*$. More specifically, the null hypothesis is rejected if $T(G) > z_{2\alpha}(1)$ for a given significance level $\alpha$.

Similarly to Theorem~\ref{theo5}, under the null hypothesis $H_0(G): \epsilon_G \ge \epsilon_0$, it follows that in the least favorable case $\epsilon_G^* = \epsilon_0$,
$
T(G) \tod \frac{1}{2}\chi^2_{0} + \frac{1}{2}\chi^2_{1} \quad \mathrm{as} \quad n \to \infty,
$ where $\chi^2_{0}$ is a degenerate random variable with mass 1 at the point zero.  
The following Theorem \ref{theo7} establishes the Type I error control properties of the testing procedure. 
\begin{theo}\label{theo7}
Assume that conditions (\ref{C1}) to (\ref{C2}) hold.
%For any fixed true group-wise performance disparity $\epsilon_G^* \in \Omega_2$, we have $\theta_P$ is a-priori known. 
Then, under the null hypothesis $H_0(G): \epsilon_G \ge \epsilon_0$, we have
\bea
\lim_{n \to \infty} \mathbb{P}\left\{ T(G) > z_{2\alpha}(1) \right\} =
\begin{cases}
\alpha,  \,\, \mathrm{if} \  \  \epsilon_G^* = \epsilon_0; \\
0,           \,\, \mathrm{if} \  \  \epsilon_G^* > \epsilon_0, \nn
\end{cases}
\eea 
where $z_{2\alpha}(1)$ is the upper $2\alpha$-quantile of $\chi^2_{1}$ distribution.
\end{theo}
The above theorem demonstrates that $\epsilon_0$ is the least favorable point under the null hypothesis. 
As discussed previously, the EL based confidence region for $\epsilon_{G}$ with asymptotically correct coverage probability $1 - \alpha$ can be constructed as
\[
\{ \epsilon_{G}: T(G)\leq z_{2\alpha}(1) \}.
\]
Moreover, for any $T(G) > 0$, the $p$-value is given by equation~\eqref{p24}. 
If $ p_{24} \geq \alpha $, we fail to reject the null hypothesis; otherwise, we reject it. 
In the case of $T(G) = 0$, we have $ p = 1 $ , leading to non-rejection of the null hypothesis.
The next theorem examines the asymptotic local power of the test statistic $T(G)$. 
\begin{coll}\label{coro3}
For the test $H_0(G):  \epsilon_G \ge \epsilon_0$  versus $H_1(G):  \epsilon_G < \epsilon_0$,  
 if the true disparity is \( \epsilon_G^* = \epsilon_0 - \tau \sigma n^{-1/2}  \) , where \( \tau > 0\) and \( \sigma^2 = { \mathbb{P}(G) }Var(L\ |\ G) \), then we have
\[
\lim_{n \to \infty} \mathbb{P}\left( T(G) > z_{2\alpha}(1) \mid \epsilon_G^* \right) = \Phi\left( \mathbb{P}(G)  \tau - \sqrt{z_{2\alpha}(1)} \right).
\]  
Therefore, for any fixed \( \epsilon_G^* > \epsilon_0 \), $\lim_{n \to \infty} \mathbb{P}\left\{ T(G) > z_{2\alpha}(1) \mid H_1 \right\} = 1.$
\end{coll}

%--------------------------------

\subsection{Two-sided tests of hypotheses on disparity} \label{appsec-two-sided}

In this subsection, we extend our framework to test whether disparities lie within a tolerance interval, relaxing the strict equality constraint of Scenario~(1). 
Many practical fairness criteria permit bounded deviations rather than requiring exact equality. 
For instance, Statistical Parity allows prediction probabilities to differ by at most a tolerance $\epsilon$ across groups \citep{PessachShmueli2022}:
\[
| \mathbb{P}(f(X)=1\mid X \in G)- \mathbb{P}(f(X)=1\mid X \in G^c) | \le \epsilon.
\]
This corresponds to testing whether the disparity $\epsilon_G$ lies within the symmetric interval $[-\epsilon, \epsilon]$. 
More generally, we consider asymmetric tolerance intervals $[\epsilon_1, \epsilon_2]$.

Formally, we test $ H_0(G): \epsilon_1 \le \epsilon_G \le \epsilon_2  $ versus \( H_1(G): \epsilon_G < \epsilon_1 \) or \( \epsilon_G > \epsilon_2 \). 
The empirical log-likelihood ratio statistic is defined as
\[
T(G)  :=  -2 \log \frac{\sup_{\epsilon_1 \leq \epsilon_G \leq \epsilon_2} L_{EL}(\epsilon_G)}{\sup_{\epsilon_G \in \mathbb{R}^1} L_{EL}(\epsilon_G)}.
\]  
From Lemma \ref{lemm_ell}, we have
\[
\sup_{\epsilon_1 \leq \epsilon_G \leq \epsilon_2} L_{EL}(\epsilon_G) = L_{EL}(\epsilon_1) \mathbbm{1}(\hepsilon_G < \epsilon_1) + L_{EL}(\hepsilon_G) \mathbbm{1}(\epsilon_1 \leq \hepsilon_G \leq \epsilon_2) + L_{EL}(\epsilon_2) \mathbbm{1}(\hepsilon_G > \epsilon_2).  
\]
Then, we have
\[
\begin{aligned}
T(G) &= \left( -2 \log \frac{L_{EL}(\epsilon_1)}{L_{EL}(\hepsilon_G)} \right) \mathbbm{1}(\hepsilon_G < \epsilon_1) + \left( -2 \log \frac{L_{EL}(\epsilon_2)}{L_{EL}(\hepsilon_G)} \right) \mathbbm{1}(\hepsilon_G > \epsilon_2).
\end{aligned}
\]

We now state the main results. To determine the rejection region for the two sided test, by analogy with the one-sided test, there are two least favorable points \( \epsilon_1 \) and \( \epsilon_2 \) in for the two sided test such that the limiting distribution of the test statistic \( T(G) \) is \( \frac{1}{2} \chi^2_0 + \frac{1}{2} \chi^2_1 \). 

\begin{theo}\label{theo8}
Assume that conditions (\ref{C1}) to (\ref{C2}) hold. 
Then, under the null hypothesis $H_0(G): \epsilon_1 \le  \epsilon_G \le \epsilon_2$ with $\epsilon_G^*$ being the true group-wise disparity, it follows that as $n \to \infty$,
\[
T(G) \tod \frac{1}{2}\chi^2_{0} + \frac{1}{2}\chi^2_{1} \quad \mathrm{for} \ \epsilon_G^* = \epsilon_1 \ \mathrm{or} \ \epsilon_2.
\]
\end{theo}

\begin{theo}\label{theo9}
Assume that conditions (\ref{C1}) to (\ref{C2}) hold.
Then, under the null hypothesis $H_0(G): H_0(G): \epsilon_1 \le  \epsilon_G \le \epsilon_2$, we have
% For any fixed true value \( \epsilon_G^* \in \Omega_3 \),  
\[
\lim_{n \to \infty} \mathbb{P}\left\{ T(G) > z_{2\alpha}(1) \right\} = 
\begin{cases} 
\alpha, & \ \mathrm{if} \  \epsilon_G^* = \epsilon_1 \ \mathrm{or} \ \epsilon_2; \\
0, & \ \mathrm{if} \  \epsilon_1 < \epsilon_G^* < \epsilon_2.
\end{cases}  
\]  
\end{theo}

According to \cite{LehmannRomano2005}, the above theorem demonstrates that $\epsilon_1$ and $\epsilon_2$ are the least favorable point under the null hypothesis. 
As discussed previously, the EL based confidence region for $\epsilon_{G}$ with asymptotically correct coverage probability $1 - \alpha$ can be constructed as
\[
\{ \epsilon_{G}: T(G)\leq z_{2\alpha}(1) \}.
\]
Moreover, for any $T(G) \ge 0$, the $p$-value of the two-sided test is given by equation~\eqref{p24}. 

Analogous to Corollary \ref{coro3}, the asymptotic power of the two-sided test statistics is given as follows.
\begin{coll}\label{coro4}
For the test $H_0(G): \epsilon_1 \le  \epsilon_G \le \epsilon_2$  versus $H_1(G): \epsilon_G < \epsilon_1 $ or $ \epsilon_G > \epsilon_2 $, 
 if the true disparity is \( \epsilon_G^* = \epsilon_1 - \tau \sigma n^{-1/2}  \) or \( \epsilon_G^* = \epsilon_2 + \tau \sigma n^{-1/2}  \) , where \( \tau > 0\) and \( \sigma^2 = { \mathbb{P}(G) }Var(L\ |\ G) \), then we have
\[
\lim_{n \to \infty} \mathbb{P}\left\{ T(G) > z_{2\alpha}(1) \mid \epsilon_G^* \right\} = \Phi\left( \mathbb{P}(G) \tau - \sqrt{z_{2\alpha}(1)} \right).
\] 
Therefore, for any fixed \( \epsilon_G^* < \epsilon_1 \) or \( \epsilon_G^* > \epsilon_2 \), $\lim_{n \to \infty} \mathbb{P}\left\{ T(G) > z_{2\alpha}(1) \mid H_1 \right\} = 1.$
\end{coll}

% ----------------------------------

\section{Proof of Theorems in Section 3} \label{appsec-proof-sec3}
\subsection{Proof of Theorem \ref{theoEL}}

\begin{lemm}\label{lemm1}
Let $\eta_1, \eta_2,\cdots, \eta_n$  be a sequence of stationary random variables,
 with $E|\eta_1|^s<\infty$ for some constants $s>0$ and $C>0$.  Then
\[
\max_{1\leq i \leq n }|\eta_i|=o(n^{1/s}), \ \   a.s.
\]
\end{lemm}

{ \sc Proof.} Using Borel-Cantelli lemma and following the proof of (2.3) in  \cite{Owen1990}, one can prove Lemma \ref{lemm1}.

\begin{lemm}\label{lemm2} (CLT)
Let $\bX_1, \bX_2, \cdots$ be independent and identically distributed $d$-dimensional random vectors with mean $\bmu$, and finite covariance matrix $\mathbf{\Sigma}$. Let $\bar{\bX}_n = {1 \over n}\sum_{j=1}^m\bX_j$, then $\sqrt{n}(\bar{\bX}_n - \bmu) \tod N(\mathbf{0}, \mathbf{\Sigma})$.
\end{lemm}

{ \sc Proof.} See Theorem 5 in \cite{Ferguson1996}. 

\begin{lemm}\label{lemm3}
$\mathcal{F}_h = \{ h \cdot \mathbbm{1}_{G} : G \in \mathcal{G} \} $ is P-Glivenko-Cantelli class so long as $\mathbb{E}_P[|h|] < \infty$.
\end{lemm}

{ \sc Proof.} See Corollary 3 in \cite{GineZinn1984}.

\begin{lemm}\label{lemmEL}
Suppose that conditions (\ref{C1}) to (\ref{C5}) are satisfied, then as $n \to \infty$
\bea \label{EL1}
 g^{*}(\bepsilon) :=\max_{1 \leq i \leq n}\| \bg_i(\bepsilon; \theta_P) \|= o(\sqrt{n})\ \ a.s., 
\eea
\bea \label{EL2}
{1 \over \sqrt{n}}{ \sum_{i=1}^n  \bg_i(\bepsilon; \theta_P) } \tod N(0, \Sigma_{EL}) , 
\eea
\bea \label{EL3}
\hat{\Sigma}_{EL} :={1 \over n}\sum_{i=1}^n  \bg_i(\bepsilon; \theta_P) (\bg_i(\bepsilon; \theta_P))^{\tau}= \Sigma_{EL} +o_P(1),
\eea
where  $\Sigma_{EL}$ in (\ref{SigmaEL}) and use $\|b\|$ to denote the $L_2$-norm of  a vector $b$.
\end{lemm}

{ \sc Proof.} Let $\bs_i = ( \mathbbm{1}_{G_1}, \dots,  \mathbbm{1}_{G_m})^{\tau}$ , we have
\bea
g^{*}(\bepsilon) \le \max_{1 \leq i \leq n}\|L_i\bs_i\| + |\theta_P|\cdot \max_{1 \leq i \leq n}\|\bs_i\| +  \|\bepsilon  \| \cdot \max_{1 \leq i \leq n}\| \bs_i\|. \nn
\eea
By condition~(\ref{C2}), $\mathbb{P}(G) $ is bounded away from 0 all $G \in \mathcal{G}$, we have
\[
\max_{1 \leq i \leq n}\|\bs_i\| = O(1)
\]
By condition~(\ref{C3}), $\mathbb{E}_P[M^2] < \infty$ and Lemma \ref{lemm1}, we have
\[
\max_{1 \leq i \leq n}\|L_i\bs_i\| \le \max_{1 \leq i \leq n}|L_i | \cdot \max_{1 \leq i \leq n}\| \bs_i \|= o(\sqrt{n})\ \ a.s.
\]
By condition~(\ref{C4}), ${\epsilon_G}$ are  uniformly bounded in absolute value for all $G \in \mathcal{G}$, and those above relations, thus $g^{*}(\bepsilon)  = o_P(\sqrt{n})$. The equation (\ref{EL1}) is proved.

Next, we will prove  (\ref{EL2}). 
To simplify notation, let $\bg_i(\bepsilon)$ be $\bg_i(\bepsilon; \theta_P)$. 
By condition~(\ref{C1}), ${ (X_i, Y_i), 1 \le i \le n}$ are independent and identically distributed random  variables, we can get $\bg_i(\bepsilon), 1 \le i \le n$  are independent and identically distributed random variables and 
\[ 
\mathbb{E} \left( {1 \over \sqrt{n}} \sum_{i=1}^n \bg_i(\bepsilon) \right) ={1 \over \sqrt{n}} \sum_{i=1}^n \mathbb{E}   \left( \bg_i(\bepsilon) \right) = 0,
\]
since 
\[
\mathbb{E} \left( g_i(\epsilon_{G_j}) \right) = \mathbb{E}\left\{  \left[ M_i - \theta_P - \epsilon_{G_j} \right]\cdot \mathbbm{1}_{G_j} \right\}  = 0, \ j = 1, \dots, m.
\]
We now derive the variance of ${1 \over \sqrt{n}}\sum_{i=1}^n\bg_i(\bepsilon)$.
Note that
\bea
Cov \left( {1 \over \sqrt{n}}\sum_{i=1}^n \bg_i(\bepsilon) \right) 
=  {1 \over n} \sum_{i=1}^n \sum_{j=1}^n Cov \left(  \bg_i(\bepsilon),  g_j(\bepsilon) \right) 
=  {1 \over n} \sum_{i=1}^n  Cov \left(  \bg_i(\bepsilon),  \bg_i(\bepsilon) \right), \nn
\eea
since $i \neq j$, $Cov \left(  \bg_i(\bepsilon),  g_j(\bepsilon) \right) = 0$.
Let
\(
Cov \left( {1 \over \sqrt{n}}\sum_{i=1}^n \bg_i(\bepsilon) \right)
= \left( \sigma^2_{kj} \right)_{m \times m}, \nn
\)
we have the off-diagonal entries equal
\bea
\sigma^2_{kj} 
= {1 \over n} \sum_{i=1}^n Cov  \left( g_i(\epsilon_{G_k}),  g_i(\epsilon_{G_j}) \right)
= { \mathbb{E}\left[(M - \theta_P - \epsilon_{G_k})(M - \theta_P - \epsilon_{G_j}) \mathbbm{1}_{G_k \cap  G_j }\right]  } \nn
\eea
and the diagonal entries equal
\bea
\sigma^2_{kk} = { \mathbb{E} \left[(M - \theta_P - \epsilon_{G_k})^2 \mathbbm{1}_{G_k} \right] } = { \mathbb{P}(G_k) }Var(M \mid G_k). \nn
\eea
Thus, $ Cov \left( {1 \over \sqrt{n}}\sum_{i=1}^n \bg_i(\bepsilon) \right) = \Sigma_{EL} $, where  $\Sigma_{EL}$ in (\ref{SigmaEL}).
Under condition (\ref{C5}), the covariance matrix $\Sigma_{EL}$ is guaranteed to be finite-dimensional and positive definite, thereby satisfying the regularity requirements of the CLT.
By Lemma \ref{lemm2}, we directly derive the asymptotic result stated in equation (\ref{EL2}).

Continually,  we will prove  equation (\ref{EL3}). 
Let us define the function $h$ through the following equalities:
\bea
h = [L-\theta_P - \epsilon_{G_k} ]  [L-\theta_P - \epsilon_{G_j}] ,\nn
\eea
where ${G_k \cap G_j} \in \mathcal{G}$ for ${G_k} \in \mathcal{G}$ and ${G_j} \in \mathcal{G}$.
It is evident that conditions (\ref{C3}) and (\ref{C4}) jointly ensure the finiteness condition 
\[
\mathbb{E}_P[|h|] \le \mathbb{E}_P[M^2] + \theta_P^2 + \left(\sup_{ G \in \mathcal{G}}\|\epsilon_G\|\right)^2 + 2\|L\|_{\infty} \cdot \left(\theta_P + \sup_{ G \in \mathcal{G}}\|\epsilon_G\|\right)  + 2 \theta_P \cdot \sup_{ G \in \mathcal{G}}\|\epsilon_G\| < \infty.
\] 
Applying Lemma \ref{lemm3}, we have the 
$ \{ [L-\theta_P - \epsilon_{G_k} ]  [L-\theta_P - \epsilon_{G_j}] \cdot \mathbbm{1}_{G_k \cap G_j} : {G_k \cap G_j} \in \mathcal{G} \} $ 
is a $P$-Glivenko-Cantelli class and hence 
\bea \label{sup}
\sup_{ {G_k \cap G_j} \in \mathcal{G}}(\mathbb{P}_n - P  )\left[ [L-\theta_P - \epsilon_{G_k} ]  [L-\theta_P - \epsilon_{G_j}] \cdot \mathbbm{1}_{G_k \cap G_j} \right] = o_P(1). 
\eea
Establishing the validity of equation (\ref{EL3}) is mathematically equivalent to demonstrating the uniform stochastic convergence  
\[
\hat{\Sigma}_{EL}(k, j) - \Sigma_{EL}(k, j) = o_P(1),  
\]  
where \(\hat{\Sigma}_{EL}(k, j)\) denotes the \((k,j)\)-th element of  \(\hat{\Sigma}_{EL}\),
i.e.,
\bea \label{EL3-1}
{1 \over n}\sum_{i=1}^n  \left\{ [L_i - \theta_P - \epsilon_{G_k} ]  [L_i - \theta_P - \epsilon_{G_j} ] \cdot \mathbbm{1}_{G_k \cap G_j}  \right\}  - \sigma_{kj}^2 = o_P(1), 
\eea
where $\sigma_{kj}^2$ in (\ref{SigmaEL}).
The left side of above equation (\ref{EL3-1}) is equal to 
\[
(\mathbb{P}_n - P  )\left[ [L-\theta_P - \epsilon_{G_k} ]  [L-\theta_P - \epsilon_{G_j}] \cdot \mathbbm{1}_{G_k \cap G_j} \right].
\]
By (\ref{sup}), we  establish the validity of equation (\ref{EL3-1}) for ${G_k} \in \mathcal{G}$ and ${G_j} \in \mathcal{G}$.
The proof of (\ref{EL3}) is thus complete.

We are now in a position to prove the first result in this article.

{\bf Proof of Theorem \ref{theoEL}.} Let $\blambda=\blambda({\bepsilon}), \rho_0=\|\blambda \|,
\blambda=\rho_0\eta_0$ and $\bg_i(\bepsilon) = \bg_i(\bepsilon;\theta_P) $. From (\ref{lambda}), we have
\[
\frac{\eta_0^{\tau}}{n}\sum_{j=1}^{n}g_j({\bepsilon})-\frac{\rho_0}{n}\sum_{j=1}^{n}
{(\eta_0^{\tau}g_j({\bepsilon}))^2\over
1+\blambda^{\tau} g_j({\bepsilon})}=0.
\]
It follows that
\[
|\eta_0^{\tau}\bar{\bg}|\geq
{\rho_0\over 1+\rho_0
  g^{*}({\bepsilon})}\lambda_{min}(\hat{\Sigma}_{EL}),
\]
where $ g^{*}({\bepsilon}) $ is defined in (\ref{EL1}), $\hat{\Sigma}_{EL}$ in (\ref{EL3}), and $\bar{\bg}=n^{-1}\sum_{i=1}^n \bg_i(\bepsilon)$.
Combining condition~(\ref{C5}) with  (\ref{EL2}), we have
\[
|\eta_0^{\tau}\bar{\bg}|
= |\eta_0^{\tau}\Sigma_{EL}^{1/2}\Sigma_{EL}^{-1/2}\bar{\bg}| 
\le \lambda_{max}(\Sigma_{EL}^{1/2}) \|\eta_0\|\|\Sigma_{EL}^{-1/2}\bar{\bg}\|=O_p(n^{-1/2}).
\]
Thus,
\[
{\rho_0\over 1+\rho_0    g^{*}({\bepsilon})}=O_p(n^{-1/2}).
\]
From  (\ref{EL1}), we have
\[
\rho_0=O_p(n^{-1/2}),
\]
i.e.,
\[
\blambda=O_p(n^{-1/2}).
\]
Let
$\gamma_i=\blambda^{\tau}\bg_i(\bepsilon)$. Then \be \max_{1\leq i \leq
n}|\gamma_i|=o_p(1).  \label{maxl}
\ee
According to
\bea
{g_j({\bepsilon})\over 1+\blambda^{\tau} g_j({\bepsilon})} 
&=& g_j({\bepsilon}) 
- { g_j({\bepsilon}) \blambda^{\tau} g_j({\bepsilon})  } 
+ { g_j({\bepsilon}) \blambda^{\tau} g_j({\bepsilon}) \blambda^{\tau} g_j({\bepsilon}) \over 1+\blambda^{\tau} g_j({\bepsilon})} \nn\\
&=& g_j({\bepsilon}) 
- { g_j({\bepsilon})  g_j^{\tau}({\bepsilon}) \blambda  }
+ { g_j({\bepsilon}) \gamma_i^2 \over 1+\blambda^{\tau} g_j({\bepsilon})}, \nn
\eea
using (\ref{lambda})  again, we have
\begin{eqnarray*}
0 & = & {1\over {n}}\sum_{j=1}^{n}{g_j({\bepsilon})\over 1+\blambda^{\tau} g_j({\bepsilon})}\\
& = & {1\over {n}}\sum_{j=1}^{n} g_j({\bepsilon}) -  \left\{{1\over
{n}}\sum_{j=1}^{n}g_j({\bepsilon})g_j^{\tau}({\bepsilon}) \right\}\blambda+ {1\over {n} }
\sum_{j=1}^{n}\frac{g_j({\bepsilon})\gamma_j^2}{1+\gamma_j}\\
&=& \bar{\bg}- \hat{\Sigma}_{EL}\blambda+{1\over {n} }
\sum_{j=1}^{n}\frac{g_j({\bepsilon})\gamma_j^2}{1+\gamma_j}.
\end{eqnarray*}
Combining with Lemma \ref{lemmEL},  we may write
\be
\blambda=\hat{\Sigma}_{EL}^{-1}\bar{\bg}+\varsigma,  \label{lambda express}
\ee
 where  $\|\varsigma\|$ is  bounded by
\[
 \lambda_{max}(\hat{\Sigma}_{EL}^{-1}) \cdot {1 \over n}\sum_{j=1}^{n}\|g_j({\bepsilon})\|^3\cdot \|\blambda\|^2
=O_p(1)O_p(n^{-1}) = O_p(n^{-1}).
\]
By equation (\ref{maxl}), we may
expand $\log(1+\gamma_i)=\gamma_i-\gamma_i^2/2+\nu_i$  where, for
some finite $B>0$,
$$
\mathbb{P}(|\nu_i|\leq B|\gamma_i|^3, 1\leq i \leq n)\rightarrow 1, \hbox{ as } n\rightarrow \infty.
$$
Therefore, from equation (\ref{lambda express}) and Taylor expansion, we have
\begin{eqnarray*}
\ell({\bepsilon})& = & 
2\sum_{j=1}^{n} \log(1+\gamma_j)=2\sum_{j=1}^{n}\gamma_j-\sum_{j=1}^{n}\gamma_j^2+2\sum_{j=1}^{n}\nu_j\\
& = & 2n\blambda^{\tau}\bar{\bg}-n\blambda^{\tau}\hat{\Sigma}_{EL}\blambda+2\sum_{j=1}^{n}\nu_j\\
&=&2n(\hat{\Sigma}_{EL}^{-1}\bar{\bg})^{\tau}\bar{\bg}+2n\varsigma^{\tau}\bar{\bg}
-n\bar{\bg}^{\tau}\hat{\Sigma}_{EL}^{-1}\bar{\bg}\\
&&-2n\varsigma^{\tau}\bar{\bg}
-n\varsigma^{\tau}\hat{\Sigma}_{EL}\varsigma+2\sum_{j=1}^{n}\nu_j\\
&=& n\bar{\bg}^{\tau}\hat{\Sigma}_{EL}^{-1}\bar{\bg}- n\varsigma^{\tau}\hat{\Sigma}_{EL}\varsigma+2\sum_{j=1}^{n}\nu_j\nn\\
&=&  \{n^{1/2}\Sigma^{-1/2}_{EL}\bar{\bg}\}^{\tau} \{\Sigma^{-1/2}_{EL}\hat{\Sigma}_{EL}\Sigma^{-1/2}_{EL}\}^{-1} \{n^{1/2} \Sigma^{-1/2}_{EL}\bar{\bg}\}\nn\\
 && - n\varsigma^{\tau}\hat{\Sigma}_{EL}\varsigma+2\sum_{j=1}^{n}\nu_j .\nn
\end{eqnarray*}
From Lemma \ref{lemmEL} and condition~(\ref{C5}), we have
\[
  \{n^{1/2}\Sigma^{-1/2}_{EL}\bar{\bg}\}^{\tau} \{\Sigma^{-1/2}_{EL}\hat{\Sigma}_{EL}\Sigma^{-1/2}_{EL}\}^{-1} \{n^{1/2} \Sigma^{-1/2}_{EL}\bar{\bg}\} \tod \chi^2_{m}.
\]
On the other hand, using Lemma \ref{lemmEL} and above derivations, we can see that $n\varsigma^{\tau}\hat{\Sigma}_{EL}\varsigma=O_p(n^{-1})=o_p(1)$
and
\[
|\sum_{j=1}^{n}\nu_j|\leq B \|\blambda\|^3\sum_{j=1}^{n}\|g_j({\bepsilon})\|^3 =O_p(n^{-3/2})O_p(n)=O_p(n^{-1/2})=o_p(1).
\]
 The proof of Theorem \ref{theoEL} is thus complete.

\subsection{Proof of Theorem \ref{theoPIEL}}

\begin{lemm}\label{lemm5}
Under suitable measurability assumptions, 
$\mathcal{F} := \{ (x, y)  \mapsto \mathbbm{1}\{(x, y) \in  G\} : G \in \mathcal{G} \} $ is Donsker if and only if $VC(G) < \infty$.
\end{lemm}

{ \sc Proof.} See Theorem 11.4.1 in \cite{Dudley1984}. When $\mathcal{F}$ is a $P$-Donsker class, this means that $\sqrt{n}(\mathbb{P}_n - P )[ \mathbbm{1}_G ] \dto \mathbb{G}[\mathbbm{1}_G]$. If $\mathcal{F}$ is a $P$-Donsker class,  then it is also $P$-Glivenko-Cantelli \citep{vanderVaart2000}, i.e., $\sup_{G \in \mathcal{G}} | \mathbb{P}_n(G) - \mathbb{P}(G) | \pto 0$.

\begin{lemm}\label{lemm6}
If $\mathcal{F}$ is a Donsker class with $\| P \|_{\mathcal{F}} < \infty $ and $g$, a uniformly bounded, measurable function, then $\mathcal{F} \cdot g$ is Donsker. 
\end{lemm}
{ \sc Proof.} See Example 2.10.10 in \cite{vanderVaartWellner1996}. 

\begin{lemm}\label{lemm7}
If $\mathcal{F}$ and $\mathcal{G}$ are uniformly bounded Donsker classes, then the pairwise products $\mathcal{F} \cdot \mathcal{G}$ form a Donsker class. 
\end{lemm}
{ \sc Proof.} See Example 2.10.8 in \cite{vanderVaartWellner1996}. 

\begin{lemm}\label{lemmPI}
Suppose that conditions (\ref{C1}) to (\ref{C6}) are satisfied,
then
\bea \label{PI1}
 \tilde{g}^{*}(\bepsilon) :=\max_{1 \leq i \leq n}\| \hat{\bg}_i(\bepsilon) \|= o(\sqrt{n})\ \ a.s., 
\eea
\be \label{PI2}
{1 \over \sqrt{n}}{ \sum_{i=1}^n  \hat{\bg}_i(\bepsilon) }\tod N(0, \Sigma_{EL}) , 
\ee
\bea \label{PI3}
{1 \over n}\sum_{i=1}^n  \hat{\bg}_i(\bepsilon) (\hat{\bg}_i(\bepsilon))^{\tau}= \Sigma_{EL} +o_P(1),
\eea
where $\hat{\bg}_i(\bepsilon) := \bg_i(\bepsilon; \hat\theta)$.
\end{lemm}

{ \sc Proof.} Let $\bs_i = ( \mathbbm{1}_{G_1}, \dots,  \mathbbm{1}_{G_m})^{\tau}$ , we have  
\[
\hat{\bg}_i(\bepsilon) =  \bg_i(\bepsilon)  - (\htheta - \theta_P)  \bs_i .
\]
and then
\[
\tilde{g}^{*}(\bepsilon) \le \max_{1 \leq i \leq n}\| \bg_i(\bepsilon) \| + |\htheta - \theta_P|\cdot \max_{1 \leq i \leq n}\| \bs_i \|.
\]
By the relation (\ref{EL1}) and condition~(\ref{C6}), we have (\ref{PI1}).

Next, we will prove  (\ref{PI2}). 
Firstly, we will show 
\bea \label{E}
{1 \over \sqrt{n}}{ \sum_{i=1}^n  \hat{\bg}_i(\bepsilon) } - {1 \over \sqrt{n}}{ \sum_{i=1}^n  \bg_i(\bepsilon) } = o_P(1).
\eea
For any given $l = (l_1, \dots, l_m)^\tau \in \mathbb{R}^m$ with $\| l \| =1$, where $l_j \in \mathbb{R}^1, j = 1, \dots, m$. Then
By definition, we have
\bea
{1 \over \sqrt{n}}{ \sum_{i=1}^n  l^{\tau}\hat{\bg}_i(\bepsilon) } - {1 \over \sqrt{n}}{ \sum_{i=1}^n   l^{\tau}\bg_i(\bepsilon) }
&=&   {1 \over \sqrt{n}}{ \sum_{i=1}^n    (\theta_P - \htheta )  l^{\tau}\bs_i  }.\nn
\eea
Let $ S_n = { \sum_{i=1}^n    (\theta_P - \htheta )  l^{\tau}\bs_i  }$, we will prove ${1 \over \sqrt{n}}S_n = o_P(1)$.
It suffices to prove  ${1 \over n}Var(S_n) \to 0$. Note that
\bea
{1 \over n}Var(S_n) 
& = & {1 \over n}(\htheta - \theta_P )^2 Var \left({ \sum_{i=1}^n  l^{\tau}\bs_i  }\right) \nn\\
& = & {1 \over n}(\htheta - \theta_P )^2 \sum_{i=1}^n \sum_{j=1}^m Var \left({   l_j\mathbbm{1}_{(X_i, Y_i) \in G_j}  }\right) \nn\\
& = &  (\htheta - \theta_P )^2   \sum_{j=1}^m l_j^2  Var \left({ \mathbbm{1}_{(X, Y) \in G_j}  }\right) \nn\\
& = &  (\htheta - \theta_P )^2   \sum_{j=1}^m l_j^2  \mathbb{P}(G_j)(1 - \mathbb{P}(G_j)), \nn
\eea
where $ \sum_{j=1}^m l_j^2  \mathbb{P}(G_j)(1 - \mathbb{P}(G_j)) = O(1)$ and $(\htheta - \theta_P )^2 = O_P({1 \over n})$, 
thus, we have completed the proof of (\ref{E}) by Markov's inequality.
And then, by Slutsky's theorem, the convergence in probability of the difference between two sequences to zero implies their limiting distributions coincide. 
Thus, by equations (\ref{EL2}) and (\ref{E}), we have completed the proof of (\ref{PI2}).

Finally, we will show (\ref{PI3}). We begin by analyzing the sample covariance structure through the following decomposition:
\bea
{1 \over n}\sum_{i=1}^n  \hat{\bg}_i(\bepsilon) (\hat{\bg}_i(\bepsilon))^{\tau} 
&=& {1 \over n}\sum_{i=1}^n  \bg_i(\bepsilon) (\bg_i(\bepsilon))^{\tau} + (\htheta - \theta_P)^2 {1 \over n}\sum_{i=1}^n \bs_i \bs_i^{\tau}  \nn\\
&& -2(\htheta - \theta_P){1 \over n}\sum_{i=1}^n\bg_i(\bepsilon)\bs_i^{\tau}, \nn
\eea
where 
${1 \over n}\sum_{i=1}^n  \bg_i(\bepsilon) (\bg_i(\bepsilon))^{\tau} = \Sigma_{EL} + o_P(1)$ in (\ref{EL2}).
To establish the asymptotic equivalence, it suffices to demonstrate:
\bea \label{PI3-1}
(\htheta - \theta_P)^2 {1 \over n}\sum_{i=1}^n \bs_i \bs_i^{\tau}  = o_P(1),
\eea
and 
\bea \label{PI3-2}
(\htheta - \theta_P){1 \over n}\sum_{i=1}^n\bg_i(\bepsilon)\bs_i^{\tau} = o_P(1).
\eea

By Lemma \ref{lemm5}, we  establish the uniform convergence $\sup_{G \in \mathcal{G}} | \mathbb{P}_n(G) - \mathbb{P}(G) | \pto 0$. Consequently, the quadratic form can be analyzed as:
\[
l^{\tau}\left( {1 \over n}\sum_{i=1}^n \bs_i \bs_i^{\tau}\right) l 
= \sum_{k=1}^m\sum_{j=1}^m l_k \left( {1 \over n}\sum_{i=1}^n \mathbbm{1}_{G_k \cap G_j} \right) l_j  
\pto  
\sum_{k=1}^m\sum_{j=1}^m l_k  \mathbb{P}({G_k \cap G_j})  l_j 
< \infty.
\]
Combining the parametric convergence rate $(\htheta - \theta_P)^2 = O_P(n^{-1})$ with the established stochastic order results, we have
\[
(\htheta - \theta_P)^2 {1 \over n}\sum_{i=1}^n \bs_i \bs_i^{\tau} = O_P(n^{-1}).
\]
Thus, we obtain the required conclusion in (\ref{PI3-1}). 

Similarly, we note that, according to Lemma \ref{lemm6}, the function  class $\mathcal{F}_1 = \{(L-\theta_P - \epsilon_G ) \cdot \mathbbm{1}_{G} \mid G \in \mathcal{G}\}$ is a $P$-Donsker class because $ \{ L-\theta_P - \epsilon_G \}$ is uniformly bounded by conditions (\ref{C1})--(\ref{C4}) and $\mathcal{F}_2 = \{ \mathbbm{1}_{G} \mid G \in \mathcal{G}\}$ is a uniformly Donsker class by Lemma \ref{lemm5}.
Then, according to Lemma \ref{lemm7}, $ \mathcal{F}_1 \cdot \mathcal{F}_2 = \{(L-\theta_P - \epsilon_{G_k} )\mathbbm{1}_{G_k} \cdot \mathbbm{1}_{G_j} \mid G_k, G_j \in \mathcal{G}\}$ is a Donsker class because the subset of an elementwise product of two uniformly bounded Donsker classes is a Donsker class. If $ \mathcal{F} = \mathcal{F}_1 \cdot \mathcal{F}_2$ is a $P$-Donsker class,  then it is also $P$-Glivenko-Cantelli  \citep{vanderVaart2000}, i.e., $\sup_{f \in \mathcal{F} } |(\mathbb{P}_n - P)[f]| \pto 0$. Consequently, the quadratic form can be analyzed as:
\bea
l^{\tau}\left( {1 \over n}\sum_{i=1}^n\bg_i(\bepsilon)\bs_i^{\tau} \right) l
&=& \sum_{k=1}^m\sum_{j=1}^m l_k \left( {1 \over n}\sum_{i=1}^n (L_i -\theta_P - \epsilon_{G_k} )\mathbbm{1}_{G_k \cap G_j} \right) l_j  \nn \\
&\pto  &
\sum_{k=1}^m\sum_{j=1}^m l_k  \mathbb{E}\left[ (L_i -\theta_P - \epsilon_{G_k} )\mathbbm{1}_{G_k \cap G_j} \right]  l_j 
< \infty. \nn
\eea
Combining  $\htheta - \theta_P = O_P(n^{-1/2})$ with the above results, we have
\[
(\htheta - \theta_P){1 \over n}\sum_{i=1}^n\bg_i(\bepsilon)\bs_i^{\tau} = O_P(n^{-1}),
\]
and then we obtain the required conclusion in (\ref{PI3-2}).
So,  we have completed the proof of Lemma \ref{lemmPI}.

We are next in the position to prove the limit results of the plugged-in EL ratio in this article.

{\bf Proof of Theorem \ref{theoPIEL}.} The proof follows analogous arguments to those established in Theorem \ref{theoEL}, and thus we omit the technical reiteration here.

\subsection{Proof of Theorem \ref{theoEEL}}

{\bf Proof of Theorem \ref{theoEEL}.} 
When $\theta_P$ is known a priori, let $\bg_i(\bepsilon) := \bg_i(\bepsilon;\theta_P)$, \(\bar{\bg}(\bepsilon) = \frac{1}{n} \sum_{i=1}^{n} \bg_i(\bepsilon)\), and $\hat{\Sigma}_{EL} = {1 \over n}\sum_{i=1}^n \bg_i(\bepsilon) (\bg_i(\bepsilon))^{\tau}$. 
Under equations \eqref{EL2}--\eqref{EL3} in Lemma \ref{lemmEL}, we have 
$\sqrt{n} \bar{\bg}(\bepsilon) \tod N(0, \Sigma_{EL})$ and 
$\hat{\Sigma}_{EL} \stackrel{p}{\longrightarrow} \Sigma_{EL} $.  
(When $\theta_P$ is unknown, let $\hat\bg_i(\bepsilon) = \bg_i(\bepsilon; \hat\theta)$. Using equations~\eqref{PI2}--\eqref{PI3} in Lemma~\ref{lemmPI}, the remainder of the proof follows analogously.) 
The Lagrange multiplier method is used to derive the expression for \( L_{EEL}(\bepsilon) \). For this purpose, let
\bea \label{eq:EEL-1}
G = -\frac{1}{2} \sum_{i=1}^{n} (np_i - 1)^2 - n \bt^{\tau} \sum_{i=1}^{n} p_i \bg_i(\bepsilon) - \lambda \left( \sum_{i=1}^{n} p_i - 1 \right)  
\eea
where \( \bt \in \mathbb{R}^{m} \). Taking the partial derivative of \( G \) with respect to \( p_i \) yields
\[
\frac{\partial G}{\partial p_i} = -n (np_i - 1) - n \bt^{\tau} \bg_i(\bepsilon) - \lambda
\]
Setting \( \frac{\partial G}{\partial p_i} = 0 \) gives
\bea \label{eq:EEL-2}
- n^2 p_i + n - n \bt^{\tau} \bg_i(\bepsilon) - \lambda = 0 
\eea
Summing both sides of equation \eqref{eq:EEL-2} over \( i \) from 1 to \( n \), we obtain
\[
- n^2 \sum_{i=1}^{n} p_i + \sum_{i=1}^{n} n - \sum_{i=1}^{n} n \bt^{\tau} \bg_i(\bepsilon) - \sum_{i=1}^{n} \lambda = 0.
\]
Using \( \sum_{i=1}^{n} p_i = 1 \), we get
\bea \label{eq:EEL-3}
\lambda = -\sum_{i=1}^{n} \bt^{\tau} \bg_i(\bepsilon) = - \bt^{\tau} \sum_{i=1}^{n} \bg_i(\bepsilon) = - n \bt^{\tau} \bar{\bg}(\bepsilon).  
\eea
Substituting \eqref{eq:EEL-3} into \eqref{eq:EEL-2} gives
\[
- n^2 p_i + n - n \bt^{\tau} \bg_i(\bepsilon) + n \bt^{\tau} \bar{\bg}(\bepsilon) = 0
\]
Simplifying the above equation yields
\bea \label{eq:EEL-4}
p_i = \frac{1}{n} + \frac{1}{n} \bt^{\tau} [\bar{\bg}(\bepsilon) - \bg_i(\bepsilon)], \quad 1 \le i \le n.  
\eea
From \eqref{eq:EEL-4}, we obtain
\bea
\sum_{i=1}^{n} p_i \bg_i(\bepsilon) &=& \sum_{i=1}^{n} \left\{ \frac{1}{n} + \frac{1}{n} \bt^{\tau} [\bar{\bg}(\bepsilon) - \bg_i(\bepsilon)] \right\} \bg_i(\bepsilon) \nn \\
&=&  \sum_{i=1}^{n} \frac{1}{n} \bg_i(\bepsilon) - \sum_{i=1}^{n} \frac{1}{n} \bt^{\tau} [\bg_i(\bepsilon) - \bar{\bg}(\bepsilon)] \bg_i(\bepsilon) \nn \\
&=&  \bar{\bg}(\bepsilon) - \frac{1}{n} \sum_{i=1}^{n} \bt^{\tau} [\bg_i(\bepsilon) - \bar{\bg}(\bepsilon)] [\bg_i(\bepsilon) - \bar{\bg}(\bepsilon)] \nn \\
&=&  \bar{\bg}(\bepsilon) - \frac{1}{n} \sum_{i=1}^{n} [\bg_i(\bepsilon) - \bar{\bg}(\bepsilon)] [\bg_i(\bepsilon) - \bar{\bg}(\bepsilon)]^{\tau} \bt \nn \\
&=&  \bar{\bg}(\bepsilon) - S(\bepsilon) \bt \nn
\eea
where \(S(\bepsilon) = \frac{1}{n} \sum_{i=1}^{n} [\bg_i(\bepsilon) - \bar{\bg}(\bepsilon)] [\bg_i(\bepsilon) - \bar{\bg}(\bepsilon)]^{\tau}\). 
Using \(\sum_{i=1}^{n} p_i \bg_i(\bepsilon) = 0\), we obtain
\bea \label{eq:EEL-5}
\bt = S^{-1}(\bepsilon) \bar{\bg}(\bepsilon)  
\eea
From \eqref{eq:EEL-4}--\eqref{eq:EEL-5}, we have
\bea
-2L_{EEL}(\bepsilon) &=& \sum_{i=1}^{n} (np_i - 1)^2 = \sum_{i=1}^{n} \left\{ \bt^{\tau}[\bar{\bg}(\bepsilon) - \bg_i(\bepsilon)] \right\}^2 \nn \\
&=& \bt^{\tau} \sum_{i=1}^{n} [\bar{\bg}(\bepsilon) - \bg_i(\bepsilon)] [\bar{\bg}(\bepsilon) - \bg_i(\bepsilon)]^{\tau} \bt \nn \\
&=& n \bt^{\tau} S(\bepsilon) \bt \nn \\
&=& n \left\{ S^{-1}(\bepsilon) \bar{\bg}(\bepsilon) \right\}^{\tau} S(\bepsilon) \left\{ S^{-1}(\bepsilon) \bar{\bg}(\bepsilon) \right\} \nn \\
&=& n \bar{\bg}^{\tau}(\bepsilon) S^{-1}(\bepsilon) \bar{\bg}(\bepsilon) \nn
\eea
By asymptotic normality, we have \(\bar{\bg}(\bepsilon) = O_p(n^{-1/2})\), and therefore \(\bar{\bg}(\bepsilon) \bar{\bg}^{\tau}(\bepsilon) = O_p(n^{-1})\). 
It follows that \(S(\bepsilon) = \hat{\Sigma}_{EL} + o_p(1)\). 
Combining this with the fact that \(\hat{\Sigma}_{EL} \stackrel{p}{\longrightarrow} \Sigma_{EL}\), we obtain \(S(\bepsilon) = \Sigma_{EL} + o_p(1)\). 
Since \(\Sigma_{EL}\) is positive definite, we have \(S^{-1}(\bepsilon) \xrightarrow{p} \Sigma_{EL}^{-1}\). 
Therefore, \(-2L_{EEL}(\bepsilon) \xrightarrow{d} \chi^2_{m}\), which completes the proof of the theorem.

%==============================================================

\section{Proof of Theorems in Section 4} \label{appsec-proof-sec4}

{\bf Proof of Lemma \ref{lemm_ell}. }
Using the method of Lagrange multipliers in section~\ref{appsec-lagrange-multiplier}, the maximization in (\ref{L}) is achieved by:
\bea \label{pi_epsilon_G}
\hat{p}_i = \hat{p}_i(\epsilon_G) = \frac{1}{n\left(1 + \lambda g_i(\epsilon_G;\theta_P)\right)}, \quad 1 \leq i \leq n, 
\eea
where the multiplier $\lambda \in \mathbb{R}^1$ satisfies the equation:
\bea \label{lambda_epsilon_G}
\sum_{i=1}^n \frac{g_i(\epsilon_G)}{1 + \lambda g_i(\epsilon_G)} = 0. 
\eea
Proving that $L_{EL}(\epsilon_G)$ is upper convex in $\epsilon_G$ is equivalent to proving that $\log  L_{EL}(\epsilon_G)$ is upper convex in $\epsilon_G$. 
Ignoring the constant term $-n \log n$, we have:
\bea \label{tilde_ell_epsilon_G}
\dot{\ell}_{EL}(\epsilon_G) := \log  L_{EL}(\epsilon_G) = -\sum_{i=1}^n \log\left[1 + \lambda(\epsilon_G) g_i(\epsilon_G) \right]. \nn
\eea
From equations (\ref{pi_epsilon_G}) and (\ref{lambda_epsilon_G}), it follows that
\[
\dot{\ell}'_{EL}(\epsilon_G) = -\sum_{i=1}^n \frac{\lambda'(\epsilon_G) g_i(\epsilon_G) - \lambda(\epsilon_G)}{1 + \lambda(\epsilon_G) g_i(\epsilon_G)} = n \lambda(\epsilon_G),
\]
where the “$'$” denotes derivative with respect to \( \epsilon_G \). 
By taking the derivative with respect to \( \epsilon_G \) in (\ref{lambda_epsilon_G}), we have
\[
\sum_{i=1}^n \frac{-1 - \lambda'(\epsilon_G) \left(g_i(\epsilon_G)\right)^2}{\left[1 + \lambda(\epsilon_G) g_i(\epsilon_G)\right]^2} = 0 \quad \Rightarrow \quad \lambda'(\epsilon_G) \sum_{i=1}^n p_i^2(\epsilon_G) \left(g_i(\epsilon_G)\right)^2 = -\sum_{i=1}^n p_i^2(\epsilon_G). 
\]
From this, it follows that \( \lambda'(\epsilon_G) < 0 \), and hence \( \dot{\ell}''_{EL}(\epsilon_G) = n \lambda'(\epsilon_G) < 0 \). Therefore, \( \dot{\ell}_{EL}(\epsilon_G) \) is upper convex in \( \epsilon_G \).

% Theorem 4
{\bf Proof of Theorem \ref{theo5}. } 
Let $\ell_{EL}(\epsilon_G) = -2 \log \frac{ L_{EL}(\epsilon_G)}{ L_{EL}(\hepsilon_G)}$. 
From Section \ref{one-sided}, we have $T(G) = \ell_{EL}(\epsilon_0) \mathbbm{1}(\hepsilon_G > \epsilon_0)$. 
Let $\bar{g}(\epsilon_G)=n^{-1}\sum_{i=1}^n g_i(\epsilon_G)$ and \( S_n = n^{-1} \sum_{i=1}^n (g_i (\epsilon_0))^2 \). 
When $\epsilon_G^* = \epsilon_0$, it follows from the proof of Theorem \ref{theoEL} that  
\bea \label{T04}
\ell_{EL}(\epsilon_0) = -2 \log \frac{L_{EL}(\epsilon_0)}{L_{EL}(\hepsilon_n)} = \frac{n \bar{g}^2(\epsilon_0)}{S_n} + o_p(1) = \frac{n \bar{g}^2(\epsilon_0)}{\sigma^2} + o_p(1),
\eea
where $ \sigma^2 = { \mathbb{E} \left[(M - \theta_P - \epsilon_0)^2 \mathbbm{1}_{G} \right] } = { \mathbb{P}(G) }Var(L\ |\ G). $
Moreover, by the central limit theorem, \( \sqrt{n}\bar{g}(\epsilon_0) / \sigma \tod Z \), where \( Z \sim N(0,1) \). Therefore, for any \( t > 0 \),  
\bea \label{T24-1}
\begin{aligned}
P\left\{ T(G) > t \right\} &= P\left\{ \ell_{EL}(\epsilon_0) > t,\; \hepsilon_G > \epsilon_0 \right\} \\
&= P\left\{ \frac{n \bar{g}^2(\epsilon_G)}{\sigma^2} + o_p(1)  > t,\; \frac{\sqrt{n}\bar{g}(\epsilon_0)}{\sigma} > 0 \right\} \\
&\to P\left\{ Z^2 > t,\; Z > 0 \right\} \\
&=  P\left\{ Z  > \sqrt{t}  \right\} \\
&= \frac{1}{2} P\left\{ \chi^2_1 > t \right\},
\end{aligned}\nn
\eea
since $ P\left\{ \chi^2_1 > t \right\} = P\left\{ Z  > \sqrt{t}  \right\} + P\left\{ Z  < -\sqrt{t}  \right\} = 2 P\left\{ Z  > \sqrt{t}  \right\} $.
Hence, we conclude that  
\[
T(G) \  \tod \  \frac{1}{2} \chi^2_0 + \frac{1}{2} \chi^2_1\quad \mathrm{for} \ \epsilon_G^* = \epsilon_0.
\]  
This completes the proof of Theorem \ref{theo5}.

% Theorem 5
{\bf Proof of Theorem \ref{theo6}. }
From Theorem \ref{theo5}, we have that  
$P\left\{ T(G) > z_{2\alpha}(1) \mid \epsilon_G^* = \epsilon_0 \right\} = \alpha$, as \( n \to \infty \).
Let \( \bar{h}(\epsilon_G) := |G|^{-1} \sum_{i=1}^nM_i \cdot \mathbf{1}_G - \theta_P - \epsilon_G \). 
Since \( \sqrt{n}\bar{g}(\epsilon_G^*) / \sigma \stackrel{d}{\to} N(0, 1) \) 
and \( |G|/n \stackrel{p}{\to} \mathbb{P}(G) \), it follows from the Slutsky Theorem that:
\bea \label{barh}
\sqrt{n}\bar{h}(\epsilon_G^*) = \frac{n}{|G|} \cdot  \sqrt{n}\bar{g}(\epsilon_G^*) \tod N\left(0,  \  \frac{\sigma^2}{\mathbb{P}^2(G)}\right).
\eea
%Therefore,
%\[
%\sqrt{n} \mathbb{P}(G) \bar{h}(\epsilon_0) / \sigma \tod N(0,  1).
%\]
Furthermore, for any fixed \( \epsilon_G^* < \epsilon_0 \), it follows from the above limit theorem that:  
\bea \label{T24-2}
\begin{aligned}
P\left\{ T(G) > z_{2\alpha}(1) \mid \epsilon_G^* \right\} 
&= P\left\{ \ell_{EL}(\epsilon_0) > z_{2\alpha}(1), \ \hepsilon_G > \epsilon_0 \mid \epsilon_G^* \right\} \\
&\le P\left\{ \hepsilon_G > \epsilon_0 \mid \epsilon_G^* \right\} \\
&= P\left\{ \frac{\sqrt{n} \mathbb{P}(G) \bar{h}(\epsilon_G^* )}{\sigma} > \frac{\sqrt{n}\mathbb{P}(G)  (\epsilon_0 - \epsilon_G^*)}{\sigma} \;\middle|\; \epsilon_G^* \right\} \\
&\to P\left\{ Z > +\infty \right\} \\
&= 0,
\end{aligned}
\eea
where \( Z \sim N(0, 1) \).  
This completes the proof of Theorem \ref{theo6}.

% Corollary 2
{\bf Proof of Corollary \ref{coro2}. }
Equation (\ref{T04}) remains valid when the true disparity is \( \epsilon_G^* = \epsilon_0 - \tau \sigma n^{-1/2} \), although in this case the true value lies within the alternative parameter space and deviates from \( \epsilon_0 \). From the central limit theorem, it can be deduced that  
\bea \label{tau}
\sqrt{n}\bar{g}(\epsilon_0) /\sigma = \sqrt{n}\bar{g}(\epsilon_G^*)/\sigma - \tau |G| /n \tod Z - \mathbb{P}(G)\tau. 
\eea
Therefore, from (\ref{T04}), (\ref{barh}) and (\ref{tau}), it follows that  
\bea 
\begin{aligned}
P\left\{ T(G) > z_{2\alpha}(1) \mid \epsilon_G^* \right\} &= P\left\{ \ell_{EL}(\epsilon_0) > z_{2\alpha}(1),\; \hepsilon_G >  \epsilon_0  \mid \epsilon_G^* \right\} \\
&= P\left\{ \frac{n \bar{g}^2(\epsilon_0)}{\sigma^2} + o_p(1) > z_{2\alpha}(1),\; \hepsilon_G > \epsilon_G^* - \tau \sigma n^{-1/2}   \right\} \\
&= P\left\{ \frac{n \bar{g}^2(\epsilon_0)}{\sigma^2} + o_p(1) > z_{2\alpha}(1),\; \frac{\sqrt{n} \mathbb{P}(G) \bar{h}(\epsilon_G^* )}{\sigma} >  -\mathbb{P}(G) \tau \right\} \\
&\to P\left\{  (Z -  \mathbb{P}(G)  \tau  )^2 > z_{2\alpha}(1),\; Z + \mathbb{P}(G)  \tau >  0 \right\} \\
&= P\left\{ Z > \sqrt{z_{2\alpha}(1)} - \mathbb{P}(G)  \tau   \right\} \\
&= \Phi\left( \mathbb{P}(G) \tau - \sqrt{z_{2\alpha}(1)} \right).
\end{aligned} \nn 
\eea
This completes the proof of Corollary \ref{coro2}.

% Theorem 6
\begin{lemm}[Theorem 1 in \cite{BenjaminiHochberg1995}] \label{lemm-BH}
For independent test statistics and for any configuration of false null hypotheses, the above procedure controls the FFR at $\alpha$.
\end{lemm}

{\bf Proof of Theorem \ref{theoFFR}.} The proof is completed via Lemma \ref{lemm-BH}.

%==============================================================

\section{Proof of Theorems in Supplemental Material \ref{appsec-extended-tests} } \label{appsec-proof-appsec1}

\subsection{Proof of One-sided test of hypotheses Theorems}

{\bf Proof of Theorem \ref{theo7}. }
From supplemental material \ref{appsec-one-sided}, we have $T(G) = \ell_{EL}(\epsilon_0) \mathbbm{1}(\hepsilon_G < \epsilon_0)$. 
Let $\bar{g}(\epsilon_G)=n^{-1}\sum_{i=1}^n g_i(\epsilon_G)$ and \( S_n = n^{-1} \sum_{i=1}^n (g_i (\epsilon_0))^2 \). 
When $\epsilon_G^* = \epsilon_0$, it follows from the proof of Theorem \ref{theoEL} that  (\ref{T04}) is also vaild and \( \sqrt{n}\bar{g}(\epsilon_0) / \sigma \tod Z \), where \( Z \sim N(0,1) \). Therefore, for any \( t > 0 \),  
\bea
\begin{aligned} \label{T14-1}
P\left\{ T(G) > t \right\} &= P\left\{ \ell_{EL}(\epsilon_0) > t,\; \hepsilon_G < \epsilon_0 \right\} \\
&= P\left\{ \frac{n \bar{g}^2(\epsilon_0)}{\sigma^2} + o_p(1)  > t,\; \frac{\sqrt{n}\bar{g}(\epsilon_0)}{\sigma} < 0 \right\} \\
&\to P\left\{ Z^2 > t,\; Z < 0 \right\} \\
&=  P\left\{ Z  <  - \sqrt{t}  \right\} \\
&= \frac{1}{2} P\left\{ \chi^2_1 > t \right\},
\end{aligned}
\eea
since $ P\left\{ \chi^2_1 > t \right\} = P\left\{ Z  > \sqrt{t}  \right\} + P\left\{ Z  < -\sqrt{t}  \right\} = 2 P\left\{ Z  < - \sqrt{t}  \right\} $.
Hence, we conclude that  
\[
T(G) \  \tod \  \frac{1}{2} \chi^2_0 + \frac{1}{2} \chi^2_1\quad \mathrm{for} \ \epsilon_G^* = \epsilon_0.
\]  
Therefore, we have that  
$P\left\{ T(G) > z_{2\alpha}(1) \mid \epsilon_G^* = \epsilon_0 \right\} = \alpha$, as \( n \to \infty \).
Let \( \bar{h}(\epsilon_G) := |G|^{-1} \sum_{i=1}^nM_i \cdot \mathbf{1}_G - \theta_P - \epsilon_G \). 
Furthermore, for any fixed \( \epsilon_G^* > \epsilon_0 \), it follows from equation~(\ref{barh}) that:  
\[
\begin{aligned}
P\left\{ T(G) > z_{2\alpha}(1) \mid \epsilon_G^* \right\} 
&= P\left\{ \hepsilon_G > \epsilon_0 \mid \epsilon_G^* \right\} \\
&= P\left\{ \frac{\sqrt{n} \mathbb{P}(G) \bar{h}(\epsilon_G^* )}{\sigma} > \frac{\sqrt{n}\mathbb{P}(G)  (\epsilon_0 - \epsilon_G^*)}{\sigma} \;\middle|\; \epsilon_G^* \right\} \\
&\to P\left\{ Z < -\infty \right\} \\
&= 0,
\end{aligned}
\]  
where \( Z \sim N(0, 1) \).  
This completes the proof of Theorem \ref{theo7}.

{\bf Proof of Corollary \ref{coro3}. }
Equation (\ref{T04}) remains valid when the true disparity is \( \epsilon_G^* = \epsilon_0 + \tau \sigma n^{-1/2} \), although in this case the true value lies within the alternative parameter space and deviates from \( \epsilon_0 \). From the central limit theorem, it can be deduced that  
\bea \label{tau2}
\sqrt{n}\bar{g}(\epsilon_0) /\sigma = \sqrt{n}\bar{g}(\epsilon_G^*)/\sigma + \tau |G| /n \tod Z + \mathbb{P}(G)\tau. 
\eea
Therefore, from (\ref{T04}), (\ref{barh}) and (\ref{tau2}), it follows that  
\[
\begin{aligned}
P\left\{ T(G) > z_{2\alpha}(1) \mid \epsilon_G^* \right\} &= P\left\{ \ell_{EL}(\epsilon_0) < z_{2\alpha}(1),\; \hepsilon_G > \epsilon_0  \mid \epsilon_G^* \right\} \\
&= P\left\{ \frac{n \bar{g}^2(\epsilon_0)}{\sigma^2} + o_p(1) > z_{2\alpha}(1),\; \hepsilon_G < \epsilon_G^* + \tau \sigma n^{-1/2}   \right\} \\
&= P\left\{ \frac{n \bar{g}^2(\epsilon_0)}{\sigma^2} + o_p(1) > z_{2\alpha}(1),\; \frac{\sqrt{n} \mathbb{P}(G) \bar{h}(\epsilon_G^* )}{\sigma} <  \mathbb{P}(G) \tau \right\} \\
&\to P\left\{  (Z -  \mathbb{P}(G)  \tau  )^2 > z_{2\alpha}(1),\; Z - \mathbb{P}(G)  \tau > 0 \right\} \\
&= P\left\{ Z < \mathbb{P}(G)  \tau  - \sqrt{z_{2\alpha}(1)} \right\} \\
&= \Phi\left( \mathbb{P}(G) \tau - \sqrt{z_{2\alpha}(1)} \right).
\end{aligned}
\] 
This completes the proof of Corollary \ref{coro3}.

\subsection{Proof of Two-sided test of hypotheses Theorems}

{\bf Proof of Theorem \ref{theo8}. }
Let $\ell_{EL}(\epsilon_G) = -2 \log \frac{ L_{EL}(\epsilon_G)}{ L_{EL}(\hepsilon_G)}$. From Section \ref{appsec-two-sided}, we have 
\[
\begin{aligned}
T(G) &= \ell_{EL}(\epsilon_1)\mathbbm{1}(\hepsilon_G < \epsilon_1) + \ell_{EL}(\epsilon_2) \mathbbm{1}(\hepsilon_G > \epsilon_2).
\end{aligned}
\]
We first assume that the true disparity \(\epsilon_G^* = \epsilon_1\). By applying the central limit theorem and arguments similar to those used in (\ref{T14-1}) from the proof of Theorem \ref{theo7}, we obtain  
\[
 \ell_{EL}(\epsilon_1) \mathbbm{1}(\hepsilon_G < \epsilon_1)  \tod \frac{1}{2} \chi^2_0 + \frac{1}{2} \chi^2_1.
\]  
Similarly, by applying the central limit theorem and arguments similar to those used in (\ref{T24-2}) from the proof of Theorem \ref{theo6}, we have, for any $t > 0$, 
\[
P\left(   \ell_{EL}(\epsilon_2)\mathbbm{1}(\hepsilon_G > \epsilon_2)   > t \mid \epsilon_G^* \right) \le P\left( \hepsilon_G > \epsilon_2  \mid \epsilon_G^* \right)  \longrightarrow  0.
\]  
It follows that  
\[
T(G) \tod \frac{1}{2} \chi^2_0 + \frac{1}{2} \chi^2_1,
\]  
which establishes the conclusion of Theorem \ref{theo8}. The same reasoning applies to the case where \(\epsilon_G^* = \epsilon_2\), thereby completing the proof of Theorem \ref{theo8}.

{\bf Proof of Theorem \ref{theo9}. }By Theorem \ref{theo8}, it follows that as \( n \to \infty \),
\[
P\left\{ T(G) > z_{2\alpha}(1) \mid \epsilon_G^* = \epsilon_1 \text{ or } \epsilon_2 \right\} = \alpha.
\]
Furthermore, for any fixed \( \epsilon_1 < \epsilon_G^* < \epsilon_2 \), by the central limit theorem and (\ref{barh}), we have
\[
\begin{aligned}
P\left\{ T(G) > z_{2\alpha}(1) \mid \epsilon_G^* \right\} 
& = P\left\{ \ell_{EL}(\epsilon_1) > z_{2\alpha}(1), \hepsilon_G < \epsilon_1 \mid \epsilon_G^* \right\} + P\left\{\ell_{EL}(\epsilon_2) > z_{2\alpha}(1),  \hepsilon_G > \epsilon_2 \mid \epsilon_G^* \right\} \\
&\leq P\left\{ \hepsilon_G < \epsilon_1 \mid \epsilon_G^* \right\} + P\left\{ \hepsilon_G > \epsilon_2 \mid \epsilon_G^* \right\} \\
&= P\left\{ \frac{\sqrt{n} \mathbb{P}(G) \bar{h}(\epsilon_G^* )}{\sigma} < \frac{\sqrt{n}\mathbb{P}(G)(\epsilon_1 - \epsilon_G^*)}{\sigma} \;\middle|\; \epsilon_G^* \right\} \\
&\quad + P\left\{ \frac{\sqrt{n} \mathbb{P}(G) \bar{h}(\epsilon_G^* )}{\sigma} > \frac{\sqrt{n}\mathbb{P}(G) (\epsilon_2 - \epsilon_G^*)}{\sigma} \;\middle|\; \epsilon_G^* \right\} \\
&\to P\left\{ Z < -\infty \right\} + P\left\{ Z > +\infty \right\} \\
&\to 0,
\end{aligned}
\]
where \( Z \sim N(0,1) \). This completes the proof of Theorem \ref{theo9}.

{\bf Proof of Corollary \ref{coro4}. } We first assume that the true disparity \(\epsilon_G^* = \epsilon_1 - \tau \sigma n^{-1/2} \). Following a similar argument to the proof of Corollary \ref{coro3}, we obtain 
\bea
\begin{aligned}
P\left\{ T(G) > z_{2\alpha}(1) \mid \epsilon_G^* \right\} 
& = P\left\{ \ell_{EL}(\epsilon_1) > z_{2\alpha}(1), \hepsilon_G < \epsilon_1 \mid \epsilon_G^* \right\} + P\left\{\ell_{EL}(\epsilon_2) > z_{2\alpha}(1),  \hepsilon_G > \epsilon_2 \mid \epsilon_G^* \right\} \\
& \le P\left\{ \ell_{EL}(\epsilon_1) > z_{2\alpha}(1), \hepsilon_G < \epsilon_1 \mid \epsilon_G^* \right\} + P\left\{ \hepsilon_G > \epsilon_2 \mid \epsilon_G^* \right\} \\
&\to P\left\{  (Z -  \mathbb{P}(G)  \tau  )^2 > z_{2\alpha}(1),\; Z - \mathbb{P}(G)  \tau \le 0 \right\} 
   +  P( Z > +\infty )\\
&\to \Phi\left( \mathbb{P}(G) \tau - \sqrt{z_{2\alpha}(1)} \right),
\end{aligned} \nn
\eea
The same result can be established for the true disparity \(\epsilon_G^* = \epsilon_2 + \tau \sigma n^{-1/2}\) through a similar derivation, which completes the proof of Corollary \ref{coro4}.

\section{More Experimental Results} \label{appsec-more-experiments}

\begin{figure}[H]
\centering
\includegraphics[width=0.85\textwidth]{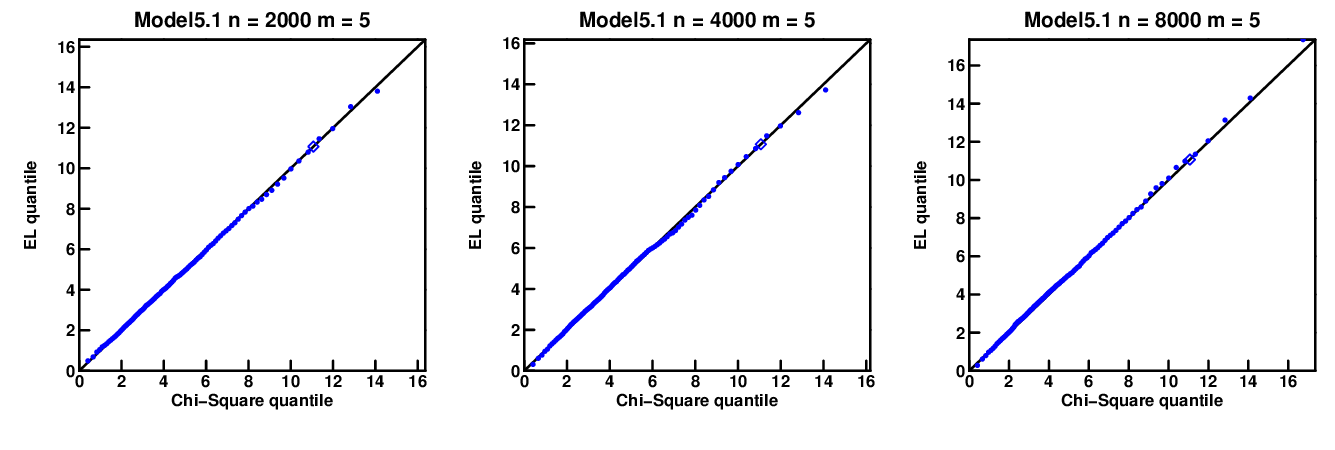}\\
\includegraphics[width=0.85\textwidth]{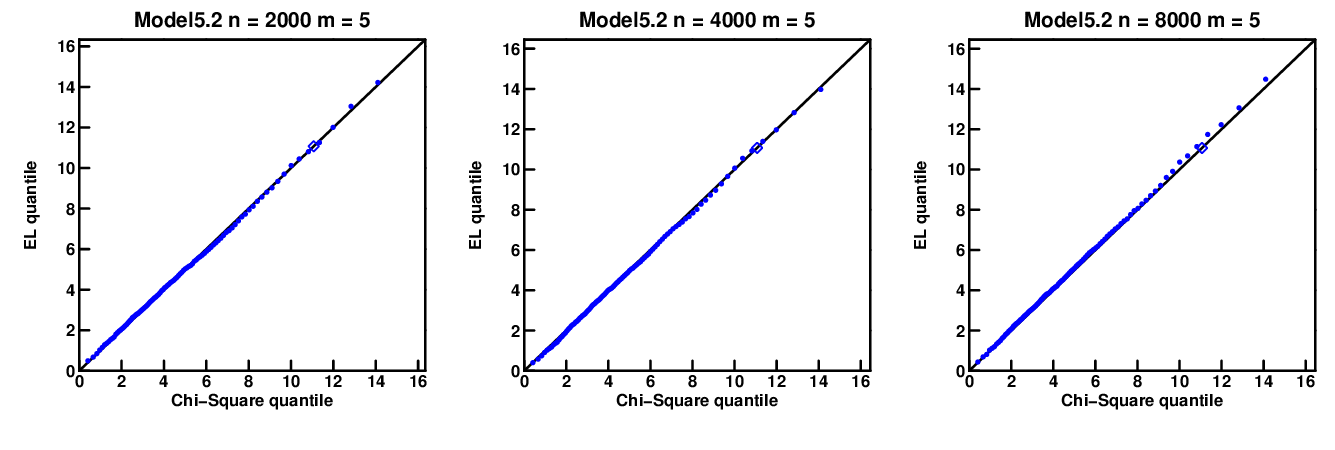}\\
\caption{Q-Q plots of $ \ell_{EL}(\bepsilon^*)$ and $\chi^2_m$ under Model \eqref{model1} and Model \eqref{model2}, respectively. }
\label{fig-QQ-EL}
\end{figure}

% Figure EEL QQ
\begin{figure}[H]
\centering
\includegraphics[width=0.85\textwidth]{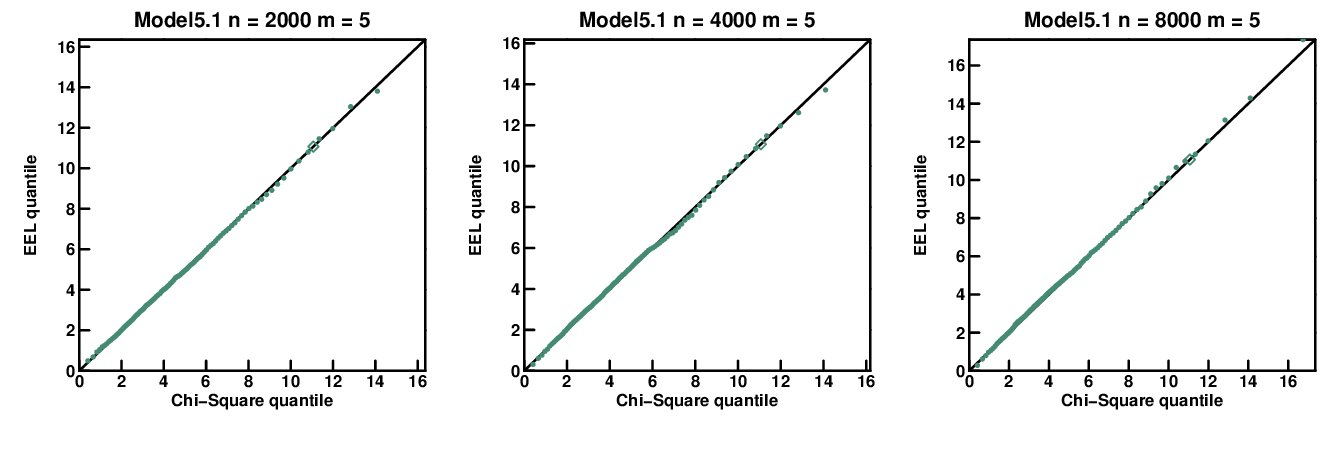}\\
\includegraphics[width=0.85\textwidth]{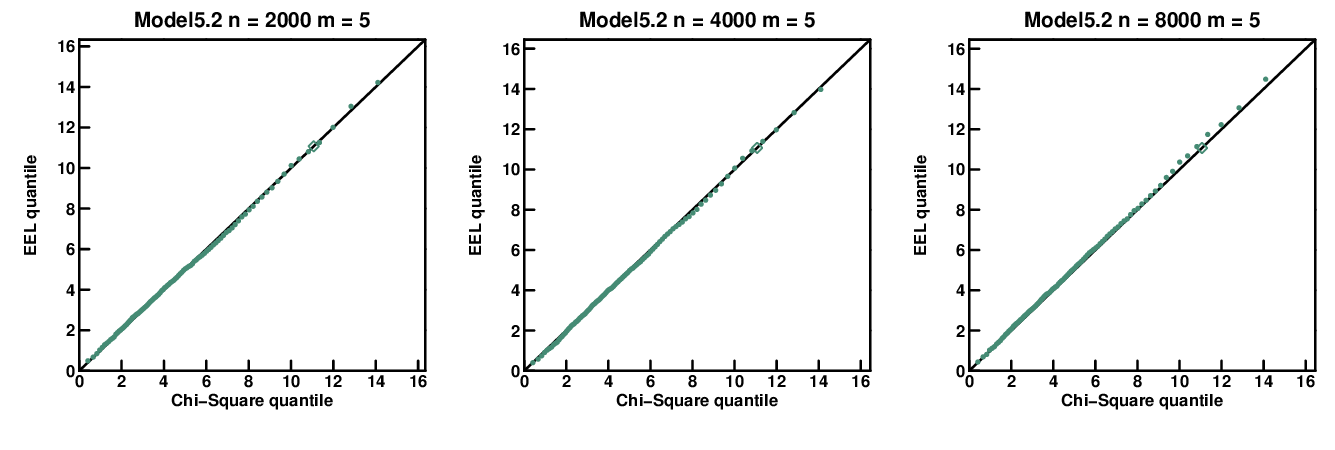}\\
\caption{Q-Q plots of $ \ell_{EEL}(\bepsilon^*)$ and $\chi^2_m$ under Model \eqref{model1} and Model \eqref{model2}, respectively. }
\label{fig-QQ-EEL}
\end{figure}

% Figure ELBH
\begin{figure}[H]
\centering
\includegraphics[width=0.49\textwidth]{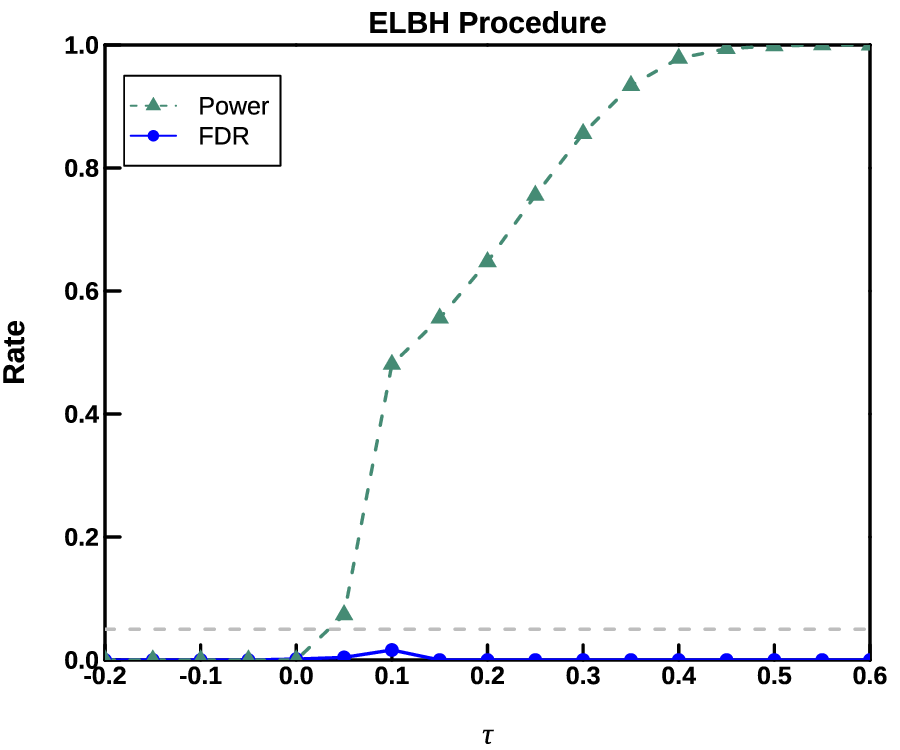 }
\includegraphics[width=0.49\textwidth]{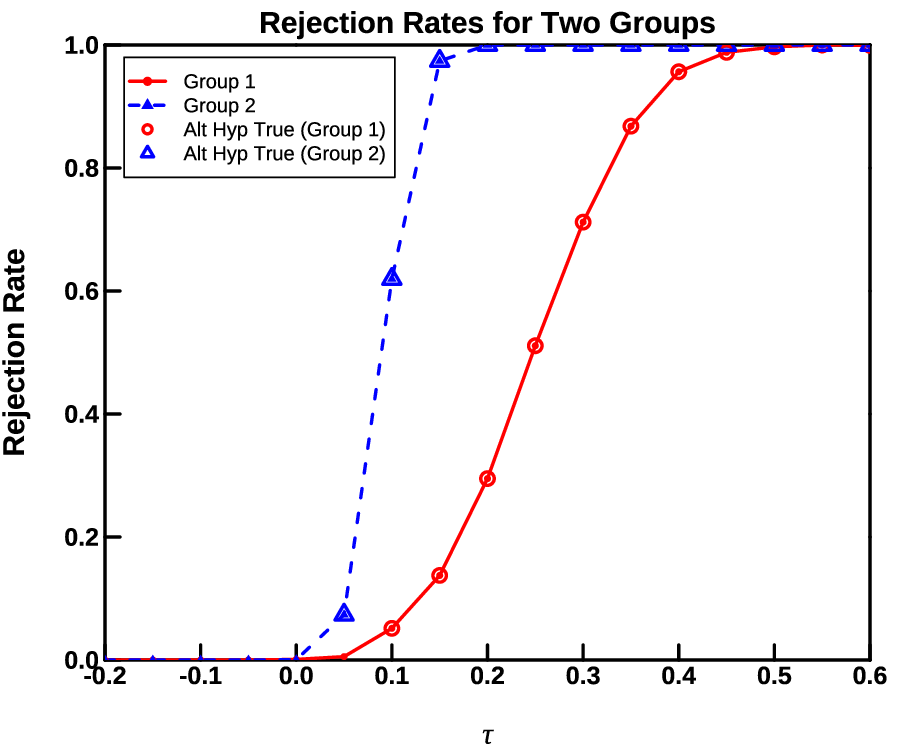 }
\caption{FFR control and statistical power under Model \eqref{model3}: (a) As $\tau$ increases, the test parameter deviates further from the true value. The FFR, representing the proportion of falsely flagged groups, remains below $\alpha$ as shown in the figure, while the power approaches 1. (b) The circle indicates rejections for the first group, and the triangle indicates rejections for the second group. As $\tau$ increases and the test parameter deviates further from the true group values, the tests rapidly converge to rejection.}
\label{fig-FFR}
\end{figure}

\section{Lagrange multiplier method} \label{appsec-lagrange-multiplier}

\begin{lemm} \label{lemmLag}
Let the weights that maximize the empirical log-likelihood ratio function be
\bea 
(\widehat{p}_1, \widehat{p}_2, \dots, \widehat{p}_n) = \arg\sup_{p_i,\,1\le i\le n}  \left\{ \log\Big(\prod^n_{i=1}n p_i\Big) \ \Big| \ p_i\geq 0, \,\, \sum_{i=1}^np_i = 1, \,\, \sum_{i=1}^n p_i  \bg_i(\bepsilon;\theta_P) =\bzero \right\}.  \nn
\eea
We have
\[
\widehat{p}_i = \widehat{p}_i(\bepsilon) = \frac{1}{n\left(1 + \blambda^{\tau} \bg_i(\bepsilon;\theta_P)\right)}, \quad 1 \leq i \leq n, 
\]
and 
$\blambda \in \mathbb{R}^m$ is the solution to the following equation:
\[ 
{ 1 \over n} \sum_{i=1}^n{ \bg_i(\bepsilon;\theta_P) \over 1+\blambda^{\tau}  \bg_i(\bepsilon;\theta_P) }=\bzero.
\]
\end{lemm}

{ \sc Proof.} Using the Lagrange multiplier method, construct the objective function
\[
\varphi = \sum_{i=1}^{n} \log(n p_i) - n \blambda^{\tau} \sum_{i=1}^{n} p_i \bg_i(\bepsilon;\theta_P)  - \rho \left( \sum_{i=1}^{n} p_i - 1 \right).
\]
Taking the partial derivative of \(\varphi\) with respect to \(p_k\) and setting it to zero, we obtain
\[
\frac{\partial \varphi}{\partial p_k} = \frac{1}{p_k} - n \blambda^{\tau} \bg_k(\bepsilon;\theta_P)  - \rho = 0, \quad k = 1, 2, \dots, n.
\]
Multiplying the above equation by \(p_k\), we get
\[
n \blambda^{\tau} p_k \bg_k(\bepsilon;\theta_P)  + \rho p_k = 1.
\]
Summing over \(k\) from 1 to \(n\) and using the constraints \(\sum_{i=1}^{n} p_i = 1\) and \(\sum_{i=1}^{n} p_i \bg_i(\bepsilon;\theta_P)  = 0\), we obtain $\rho = n.$ Substituting into the previous equation, we solve for
\[
p_k = \frac{1}{n} \cdot \frac{1}{1 + \blambda^{\tau} \bg_k(\bepsilon;\theta_P) }.
\]
Substituting this into the constraint \(\sum_{i=1}^{n} p_i \bg_i(\bepsilon;\theta_P)  = 0\), we have
\[
\sum_{i=1}^{n} \frac{\bg_i(\bepsilon;\theta_P) }{1 + \blambda^{\tau} \bg_i(\bepsilon;\theta_P) } = 0.
\]

% -------------------------------------------------------------------------------------------------
\section{Algorithm} \label{appsec-algorithm}

\begin{algorithm}[H]
\caption{The EEL Certification under the null hypothesis $H_0 : \bepsilon = \bzero$ }
\label{algo-EEL_certification}
\begin{algorithmic}[1]
\Require Subpopulation set $\mathcal{G}$, holdout dataset $\mathcal{D}$, %target $\theta_P$, 
confidence level $\alpha$.
\State Compute: $\bg_i(\bzero)$ according to \eqref{eq:3.2};
\State Compute: $\hat{p}_i$ as in \eqref{eel-solution}; 
\State Compute: $\ell_{EEL}(\bzero)$ and $p^{EEL}$ ;
\If{$p^{EEL} \geq \alpha$}
    \State \Return Model fairness certified
\Else
    \State \Return Model unfairness certified
\EndIf
\end{algorithmic}
\end{algorithm}

\begin{algorithm}[H]
\caption{EL $p$-value under the null hypothesis $H_0(G) : \epsilon_G = \epsilon_0$ }
\label{algo-p04}
\begin{algorithmic}[1]
\Require Subpopulation set $\mathcal{G}$, holdout dataset $\mathcal{D}$, target $\theta_P$, confidence level $\alpha$, parameter $\epsilon_0$
\State Compute: $g(\mathcal{D}; \epsilon_0)$ as in \eqref{gi_epsilon_G};
\State Compute: $\lambda$ and $p_i, 1 \le i \le n$, as in \eqref{lambda_epsilon_G} and \eqref{pi_epsilon_G} respectively;
\State Compute: $ \ell_{EL}(\epsilon_0; \theta_P) $ ;
\State Compute: $T(G) = \ell_{EL}(\epsilon_0; \theta_P) $;
\State Compute: $p$ as in \eqref{p04};
\State \Return $p$
\end{algorithmic}
\end{algorithm}

\begin{algorithm}
\caption{EL $p$-value under the null hypothesis $H_0(G) : \epsilon_G \le \epsilon_0$ }
\label{algo-p24}
\begin{algorithmic}[1]
\Require Subpopulation set $\mathcal{G}$, holdout dataset $\mathcal{D}$, target $\theta_P$, confidence level $\alpha$, parameter $\epsilon_0$
\State Compute: $\hepsilon_G$ as in \eqref{hat_epsilon_G};
\If{$\hepsilon_G > \epsilon_0$}
    \State Compute: $T(G)$ as in Steps 1--4 of Algorithm~\ref{algo-p04};
\Else
    \State $T(G) \gets 0$
\EndIf
\If{$T(G) \neq 0$}
    \State Compute:  $p$ as in \eqref{p24}
\Else
    \State $p \gets 1$;
\EndIf
\State \Return $p$
\end{algorithmic}
\end{algorithm}

\begin{algorithm}[H]
\caption{EL $p$-value under the null hypothesis $H_0(G) : \epsilon_G \ge \epsilon_0$ }
\label{algo-p14}
\begin{algorithmic}[1]
\Require Subpopulation set $\mathcal{G}$, holdout dataset $\mathcal{D}$, target $\theta_P$, confidence level $\alpha$, parameter $\epsilon_0$
\State Compute: $\hepsilon_G$ as in \eqref{hat_epsilon_G};
\If{$\hepsilon_G < \epsilon_0$}
    \State Compute: $T(G)$ as in Steps 1--4 of Algorithm~\ref{algo-p04};
\Else
    \State $T(G) \gets 0$
\EndIf
\State Compute: $p$ as in Steps 7--11 in Algorithm~\ref{algo-p24};
\State \Return $p$
\end{algorithmic}
\end{algorithm}

\begin{algorithm}[H]
 \caption{EL $p$-value under the null hypothesis $H_0(G) : \epsilon_1 \le \epsilon_G \le \epsilon_2$}
 \label{algo-p34}
 \begin{algorithmic}[1]
 \Require Subpopulation set $\mathcal{G}$, holdout dataset $\mathcal{D}$, target $\theta_P$, confidence level $\alpha$, parameters $\epsilon_1$ and $\epsilon_2$.
 \State Compute: $\hepsilon_G$ as in \eqref{hat_epsilon_G};
 \If{$\hepsilon_G < \epsilon_1$}
     \State Compute: $g(\mathcal{D}; \epsilon_1)$ as in \eqref{gi_epsilon_G};
     \State Compute: $T(G)$ as in Steps 2--4 of Algorithm~\ref{algo-p04};
 \ElsIf{$\hepsilon_G > \epsilon_2$}
     \State Compute: $g(\mathcal{D}; \epsilon_2)$ as in \eqref{gi_epsilon_G};
     \State Compute: $T(G)$ as in Steps 2--4 of Algorithm~\ref{algo-p04};
 \Else
     \State $T(G) \gets 0$
 \EndIf
 \State Compute: $p$ as in Steps 7--11 in Algorithm~\ref{algo-p24};
 \State \Return $p$
 \end{algorithmic}
 \end{algorithm}

\section{Discussion} \label{appsec-discussion}

While the ELFA framework provides a principled foundation for auditing, we recognize several avenues for further development.

First is about scope of Fairness. Currently, our framework focuses exclusively on group fairness. Future research could extend these empirical likelihood methods to individual fairness, ensuring that similar individuals receive similar treatment regardless of their subpopulation membership.
Second is about small Sample Constraints.  Our asymptotic results rely on sufficiently large sample sizes. In intersectional auditing, where specific subgroups (e.g., ``elderly minority females in a specific zip code") may have very few observations, these assumptions can be strained. Exploring finite-sample corrections, such as Bartlett-correctable empirical likelihood, would enhance the framework's reliability in data-sparse environments.
Third, we also care about automation and Real-time Monitoring.  Integrating automated subgroup discovery would allow auditors to find disparities in unanticipated intersections of features. Similar to flagging which subgroups exhibit significant disparities, the flagging method can be extended to identify whether the model differs significantly across different group fairness definitions and to flag group fairness types with significant differences.
Additionally, extending ELFA to online settings would enable continuous, real-time fairness monitoring as model behavior shifts over time.

Despite these limitations, the ELFA framework balances theoretical rigor with practical efficiency. By providing a reliable method to certify and flag disparities without requiring model internals, ELFA offers a vital mechanism for ensuring transparency and accountability in the increasingly automated landscape of modern decision-making.

% The proposed framework has several limitations. 
% First, we focus exclusively on group fairness—assessing disparities across predefined subpopulations—rather than individual fairness, which would require ensuring similar treatment for individuals with similar characteristics regardless of group membership.
% Second, our asymptotic results require sufficiently large sample sizes within each subpopulation. This assumption may be violated when examining intersectional groups, where sample sizes can become prohibitively small.
% These limitations point to several directions for future work.
% Methodologically, extending the framework to individual fairness and developing finite-sample corrections (e.g., Bartlett-correctable empirical likelihood) would enhance both scope and practical applicability.
% Practically, integrating automated subgroup discovery could identify disparities in unanticipated subpopulations, while extensions to online settings would enable real-time fairness monitoring.
%
% Despite these limitations, the ELFA framework provides a principled and practical approach to fairness auditing that balances theoretical rigor with computational efficiency. 
% The framework's ability to operate without model internals makes it particularly valuable for third-party auditing scenarios, where transparency and accountability are paramount.

\end{document}